\documentclass[letterpaper]{article} 
\usepackage{aaai2026}  
\usepackage{times}  
\usepackage{helvet}  
\usepackage{courier}  
\usepackage[hyphens]{url}  
\usepackage{graphicx} 
\usepackage{natbib}  
\usepackage{caption} 
\usepackage{algorithm}
\usepackage{algorithmic}

\usepackage{booktabs}
\usepackage{amsfonts}
\usepackage{amsmath}
\usepackage{amssymb}
\usepackage{nicefrac}
\usepackage{microtype}
\usepackage{xcolor}
\usepackage{subcaption}
\usepackage{multirow}
\usepackage{enumitem}

\graphicspath{{figures/}}

\newtheorem{proposition}{Proposition}
\newtheorem{assumption}{Assumption}

\title{Why Training-Free Token Reduction Collapses:\\ The Inherent Instability of Pairwise Scoring Signals}
\author{
    Shanglin Yang
}
\affiliations{
    w8049284@gmail.com
}

\begin{document}

\maketitle

\begin{abstract}
Training-free token reduction methods for Vision Transformers (ToMe, ToFu, PiToMe, and MCTF) employ different scoring mechanisms, yet they share a closely matched cliff-like collapse at high compression. This paper explains \emph{why}. We develop a diagnostic framework with two tools, ranking consistency $\rho_s$ and off-diagonal correlation $\rho_\text{off}$, that decomposes the collapse into (1)~a signal-agnostic error amplifier inherent to layer-wise reduction, predicting convex Pareto curves and $r_{\text{crit}} \propto 1/L$; and (2)~shared reliance on \emph{pairwise} similarity signals whose ranking consistency degrades from $\rho_s{=}0.88$ to $0.27$ in deep layers. Pairwise rankings are inherently unstable ($O(N_p^2)$ joint perturbations) while unary signals enjoy greater stability ($O(N_p)$ perturbations, CLT). From three design principles derived from this diagnosis, we construct CATIS as a constructive validation: unary signals raise the trigger threshold, triage suppresses the gain. On ViT-Large at 63\% FLOPs reduction, CATIS retains 96.9\% of vanilla accuracy (81.0\%) on ImageNet-1K where all baselines collapse to 43--65\%.
\end{abstract}

\section{Introduction}
\label{sec:intro}

\begin{figure*}[t]
  \centering
  \begin{subfigure}[t]{0.48\textwidth}
    \centering
    \includegraphics[width=\linewidth]{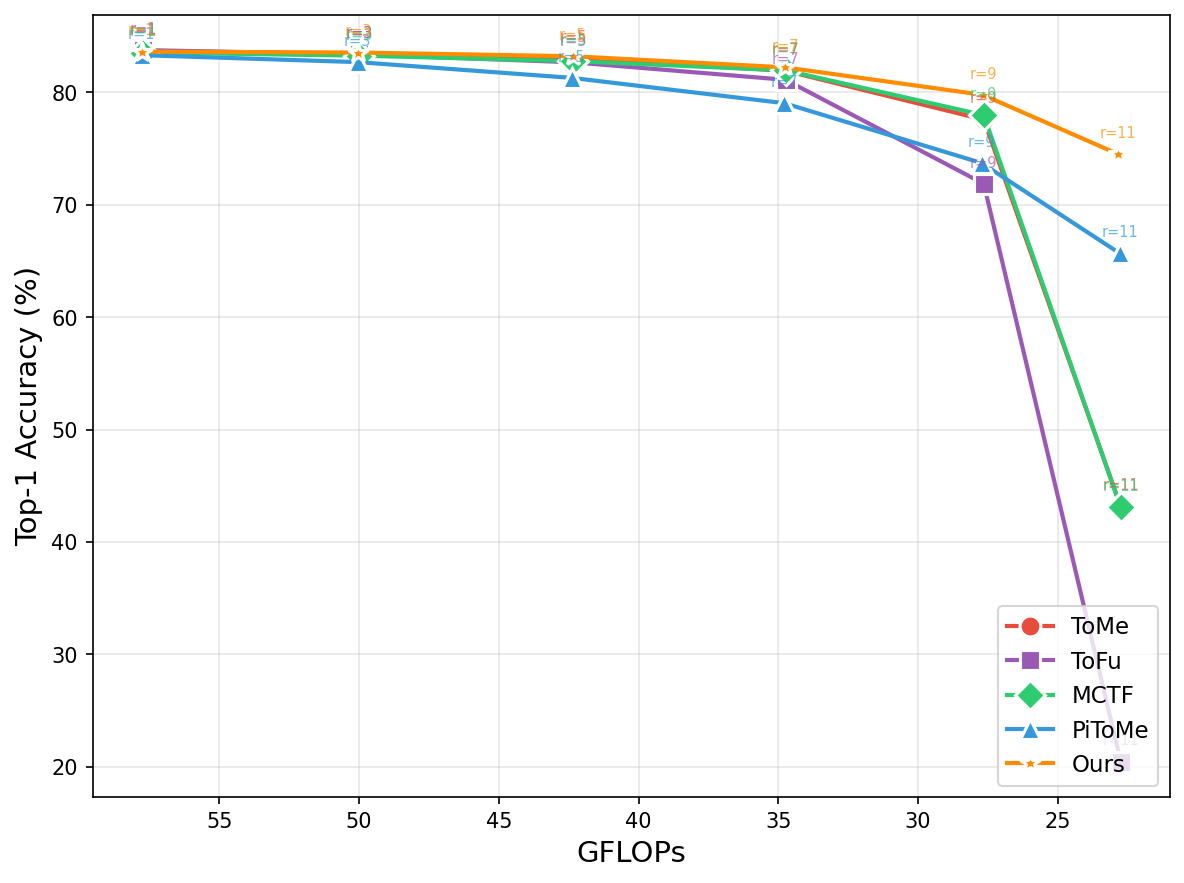}
    \caption{ImageNet-1K (val)}
  \end{subfigure}
  \hfill
  \begin{subfigure}[t]{0.48\textwidth}
    \centering
    \includegraphics[width=\linewidth]{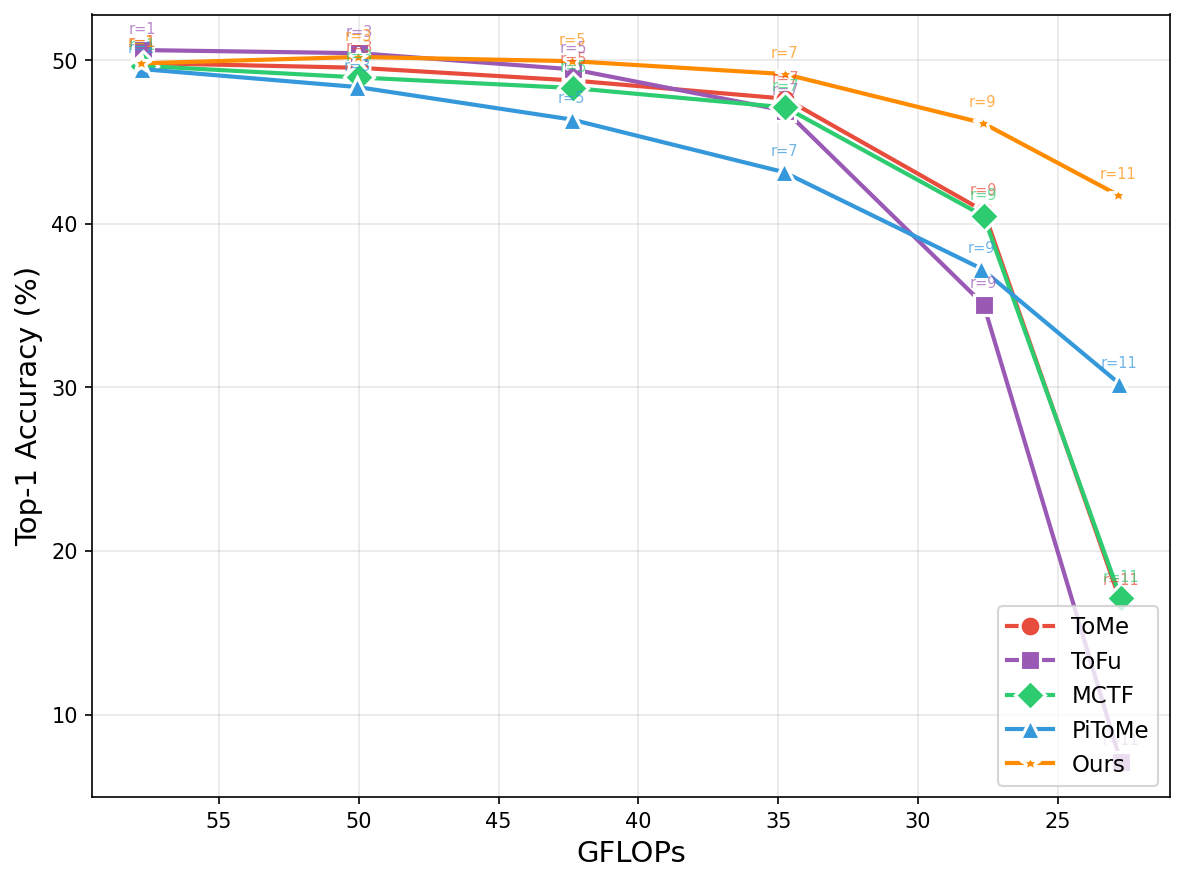}
    \caption{Shifted (A / R / Sketch)}
  \end{subfigure}
  \caption{Pareto curves on ViT-Large. Despite using fundamentally different scoring formulas, all four baselines exhibit cliff-like collapse at closely matched critical points, motivating a unified structural explanation (Section~\ref{sec:diagnosis}). Shifted benchmarks (right) confirm the collapse threshold is invariant to data distribution. CATIS (derived from the diagnosis) is shown for reference.}
  \label{fig:pareto-teaser}
\end{figure*}

Training-free token reduction methods for Vision Transformers (ViTs)~\citep{dosovitskiy2020image, vaswani2017attention} remove redundant tokens at inference time without modifying model weights, achieving promising results at moderate compression~\citep{bolya2022token, kim2024token, tran2024accelerating, lee2024multi}. Under aggressive compression, however, all four leading methods collapse at closely matched critical points (Figure~\ref{fig:pareto-teaser}). This paper explains why.

\textbf{Shared collapse pattern.} ToMe, ToFu, PiToMe, and MCTF differ substantially in design, yet a Pareto analysis across $7$ models $\times$ $5$ datasets reveals that their accuracy cliffs coincide. On ViT-Large, all four lose 23--40pp when $r{:}\,9 \to 11$. If the collapse were caused by each method's specific design, we would expect different failure points; instead, the shared pattern points to a common underlying cause.

We trace this shared collapse to \emph{two coupled components of the same mechanism} (Section~\ref{sec:diagnosis}). First, layer-wise token reduction creates a positive feedback loop that amplifies per-layer estimation errors forward through the network. We formalize this as a distortion recurrence, yielding three predictions verified across our data: (i)~convex Pareto curves, (ii)~$r_{\text{crit}} \propto 1/L$ (deeper models collapse at lower per-layer reduction), and (iii)~improving signal quality or protecting tokens both delay the collapse threshold.

Second, all existing methods rely on \emph{pairwise} token similarity for importance scoring. Pairwise rankings are inherently unstable in deep layers because they depend on $O(N_p^2)$ jointly perturbed elements, whereas unary signals depend on only $O(N_p)$ perturbations and enjoy greater stability by the central limit theorem. On DeiT-3 Base (Figure~\ref{fig:cosine-degradation}), the ranking consistency $\rho_s$ of pairwise signals drops from 0.88 to 0.27 in deep layers (i.e., rankings are nearly randomized), explaining \emph{why} all four methods collapse at the same critical ratio.

From this diagnostic framework, we derive three design principles (P1--P3), and show that CATIS (\textbf{C}omplementary \textbf{A}ctivation \textbf{T}riage with \textbf{I}mportance \textbf{S}coring) follows as a \emph{constructive validation}:
\begin{itemize}[leftmargin=*, nosep]
  \item \textbf{Diagnostic framework} unifying four methods under one failure mechanism, with verified predictions ($r_{\text{crit}} \propto 1/L$, convex Pareto curves) and reusable diagnostics ($\rho_s$, $\rho_\text{off}$).
  \item \textbf{Signal class analysis} showing pairwise signals are the structural root cause; unary signals are inherently more stable ($\rho_s{:}\;0.27 \to 0.37$--$0.48$).
  \item \textbf{Constructive validation (CATIS)} following P1--P3: on ViT-Large at 63\% FLOPs reduction, it retains 96.9\% of vanilla accuracy (81.0\% vs.\ 43--65\% for baselines).
\end{itemize}

\section{Related Work}
\label{sec:related}

\textbf{Token reduction for Vision Transformers.}
For surveys, see \citet{papa2024survey} and \citet{kong2025token}.
Training-based methods~\citep{rao2021dynamicvit, liang2022not, yin2022vit, chen2023diffrate, kong2022spvit, xu2022evo, wei2023joint, fayyaz2022adaptive, tang2022patch, pan2021ia, ryoo2021tokenlearner, chavan2022vision, yu2023x, chen2023cf, yu2022width, zong2022self} learn token selection during fine-tuning, sometimes aided by distillation~\citep{touvron2021training}, but require per-model retraining, limiting applicability to pretrained model zoos and downstream tasks where fine-tuning budgets are unavailable. Our work targets the training-free setting, where the collapse is both more severe (no learned compensation) and more amenable to structural diagnosis.
Training-free methods~\citep{bolya2022token, kim2024token, tran2024accelerating, lee2024multi, marin2021token, bonnaerens2023learned, cao2023pumer} all rely on \emph{pairwise} token similarity for importance scoring, differing only in the specific metric (K-vector cosine, value-vector similarity, multi-criteria attraction) and aggregation strategy (bipartite matching, MLERP, energy scores). Our diagnostic framework identifies this shared reliance, rather than any specific formula, as the structural root cause of collapse (Section~\ref{sec:diagnosis}).
Orthogonal directions include FFN-stage reduction~\citep{nikzad2025sata}, efficient attention~\citep{dao2022flashattention}, linear attention~\citep{katharopoulos2020transformers}, and dynamic depth~\citep{raposo2024mixture}.

\textbf{Robustness of Vision Transformers.}
ViTs exhibit strong intrinsic robustness compared to CNNs~\citep{bhojanapalli2021understanding, he2016deep, paul2022vision, geirhos2018imagenet, qin2022understanding, mao2022towards, naseer2021intriguing}, and in-distribution accuracy strongly predicts out-of-distribution performance~\citep{taori2020measuring}.
However, this literature focuses on model architecture and training; the impact of token reduction on robustness remains unexplored.

\section{Diagnosing the Collapse: A Quantitative Framework}
\label{sec:diagnosis}

As shown in Figure~\ref{fig:pareto-teaser}, all four methods collapse at closely matched critical points across 7~models (12--40 layers, three pretraining paradigms) and 5~datasets (Section~\ref{sec:experiments}). The collapse threshold is invariant to data distribution, depending only on model depth $L$ and per-layer reduction $r$. We decompose this shared failure into two coupled components below.

\subsection{Component 1: The Error Amplifier}
\label{sec:feedback-loop}

\textbf{Mechanism.} Layer-wise token reduction creates a positive feedback loop. At each layer, $r$ tokens are reduced (merged or evicted), altering the context for importance estimation in subsequent layers. Suboptimal merges produce distorted hybrid vectors that persist in the residual stream; suboptimal evictions discard useful information that subsequent layers would have relied upon. Both degrade estimation quality, and the resulting errors are amplified layer by layer because the next layer's importance ranking is read off the (already-degraded) feature population. The mechanism is signal-agnostic: it functions as an \emph{amplifier} whose trigger threshold depends on per-layer signal quality, but whose existence does not depend on which signal is used.

\textbf{Formalisation: monotone-feedback recurrence.} Let $\Delta^{(l)} \ge 0$ denote the \emph{cumulative representation distortion} entering layer $l$, aggregated across all reduction operations applied in preceding layers, and let $\epsilon^{(l)} \in [0,1]$ denote the \emph{per-operation distortion rate} at layer $l$ (applicable uniformly to merge and evict). Each reduction operation contributes expected damage intensity $\delta > 0$ to subsequent layers, giving the layer-by-layer recurrence
\begin{equation}
  \Delta^{(l+1)} \;=\; \Delta^{(l)} \;+\; \epsilon^{(l)} \cdot r \cdot \delta, \qquad \Delta^{(0)} \;=\; 0.
  \label{eq:recurrence}
\end{equation}
The structural content of the amplifier is the assumption that prior distortion degrades the attention weights and feature statistics that subsequent layers use for importance estimation, and therefore raises $\epsilon$ itself. We codify this as a single, mild structural assumption:

\begin{assumption}[Monotone feedback coupling]
\label{ass:feedback}
The per-operation distortion rate $\epsilon^{(l)}$ is a monotonically non-decreasing function of the cumulative distortion $\Delta^{(l)}$.
\end{assumption}

Under Assumption~\ref{ass:feedback} the sequence $\{\Delta^{(l)}\}$ is super-linear in $l$ (Appendix~\ref{app:feedback-loop}). The mildest closed-form coupling reproducing the empirically observed cliff-like transition is the linear one, $\epsilon^{(l)} = \epsilon_0 + \alpha \Delta^{(l)}$ with $\epsilon_0, \alpha > 0$, which converts Eq.~\ref{eq:recurrence} into a first-order linear recurrence whose solution is
\begin{equation}
  \Delta^{(l)} \;=\; \frac{\epsilon_0}{\alpha}\bigl(\beta^l - 1\bigr), \qquad \beta \;=\; 1 + \alpha r \delta \;>\; 1
  \label{eq:closed-form}
\end{equation}
(derivation in Appendix~\ref{app:feedback-loop}). This casts the amplifier as a discrete dynamical system with a single state variable $\Delta^{(l)}$, an exponential rate $\beta$ that depends multiplicatively on the per-layer reduction $r$, and a single forcing constant $\epsilon_0$. The qualitative super-linear growth survives any monotone non-linear coupling; the linear closed form is used to generate quantitative predictions whose form is independent of unobservable constants.

\begin{proposition}[Distortion recurrence and inverse-depth scaling]
\label{prop:amplifier}
Under Assumption~\ref{ass:feedback}, the cumulative representation distortion $\Delta^{(l)}$ from layer-wise token reduction grows super-linearly in $l$. Under the linear coupling $\epsilon^{(l)} = \epsilon_0 + \alpha \Delta^{(l)}$, $\Delta^{(l)} = \Theta(\beta^l)$ with $\beta = 1 + \alpha r \delta > 1$. If collapse is triggered when $\Delta^{(L)}$ reaches a model-independent threshold $T$, then
\begin{equation}
  r_{\text{crit}} \;\approx\; \frac{\ln\bigl(1 + \alpha T/\epsilon_0\bigr)}{\alpha\,\delta\,L} \;\propto\; \frac{1}{L},
  \label{eq:rcrit}
\end{equation}
i.e.\ deeper models collapse at proportionally lower per-layer reduction. The first-order approximation $\ln(1+x) \approx x$ used in the derivation requires $\alpha r \delta \ll 1$.
\end{proposition}

In other words, \textbf{if positive feedback drives the collapse, then deeper models must collapse at proportionally lower per-layer reduction}. This prediction is falsifiable across model depths without fitting any free parameters: the proportionality $r_{\text{crit}} \propto 1/L$ holds regardless of the unobservable constants $\alpha, T, \epsilon_0, \delta$, which set only the slope, not the functional form.

\begin{figure*}[t]
  \centering
  \includegraphics[width=\textwidth]{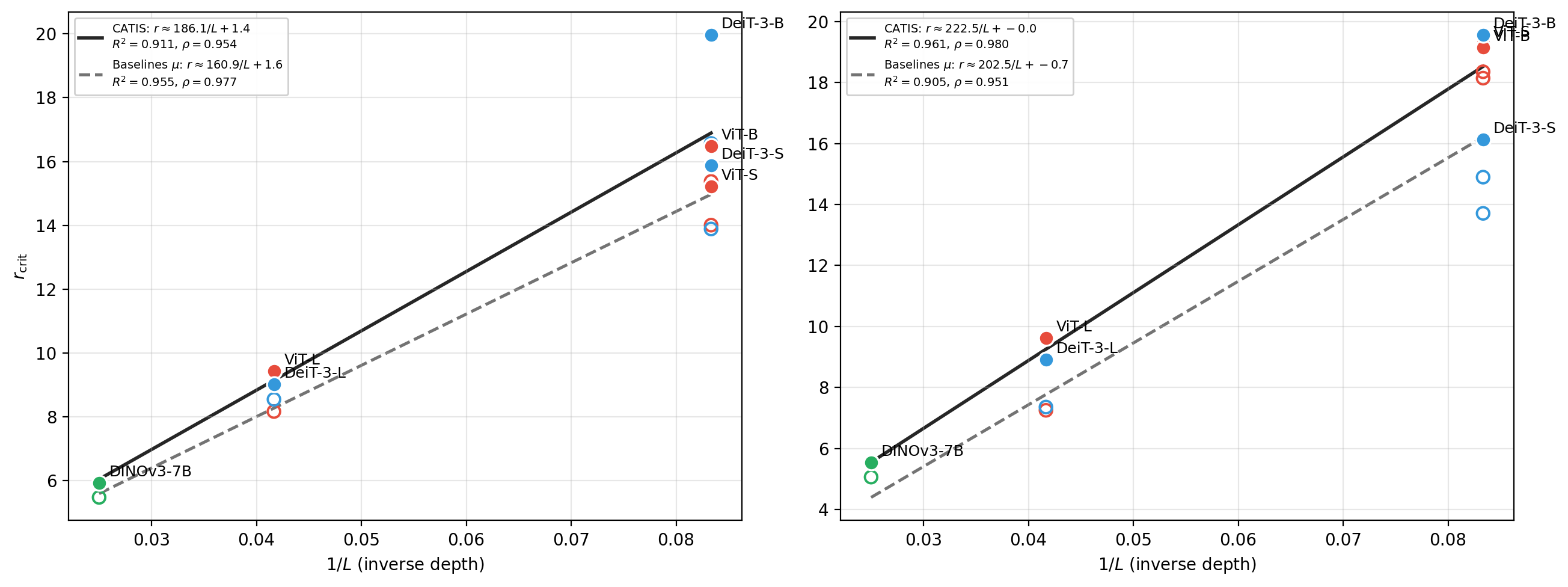}
  \caption{$r_{\text{crit}}$ vs.\ $1/L$ for all 7~models (clean / shifted). $r_{\text{crit}}$: minimum $r$ with $>5$\,pp drop vs.\ vanilla. Linear fits yield $R^2>0.9$ ($n{=}7$); the consistent trend across diverse architectures supports the $1/L$ scaling. CATIS's larger slope reflects lower $\epsilon_0$.}
  \label{fig:rcrit-vs-depth}
\end{figure*}

\textbf{Immediate verification.}
Figure~\ref{fig:rcrit-vs-depth} tests this directly, plotting $r_{\text{crit}}$ against $1/L$ for all 7~models under both clean and shifted evaluation ($r_{\text{crit}}$: minimum $r$ causing $>5$\,pp drop vs.\ vanilla).
Linear fits yield $R^2 > 0.9$ for all methods across $L{=}12$ to $L{=}40$ and three pretraining paradigms.
Models at the same depth but different widths cluster tightly, confirming that $L$ (rather than width or pretraining paradigm) is the dominant factor governing the collapse threshold.
CATIS achieves a higher proportionality constant than baselines (186 vs.\ 161 on clean; 223 vs.\ 203 on shifted), consistent with its lower $\epsilon_0$.

More broadly, the recurrence yields three predictions that follow from \emph{any} monotonically increasing coupling $\epsilon(\Delta)$, not from the linear form specifically:

\begin{table*}[t]
  \centering
  \small
  \begin{tabular}{p{5.0cm} p{4.2cm} p{5.0cm}}
    \toprule
    \textbf{Prediction} & \textbf{Mechanism} & \textbf{Empirical verification} \\
    \midrule
    \emph{P-I.} Pareto curves are convex (distortion grows exponentially in $l$) & $\beta^l$ term in closed-form & All methods, all models (Fig.~\ref{fig:pareto-teaser}; Appendix~\ref{app:pareto-full}) \\[3pt]
    \emph{P-II.} Deeper models collapse at lower $r$: $r_{\text{crit}} \propto 1/L$ & $\beta^L$ exponentially sensitive to $L$ & $R^2 > 0.9$ across 7~models (Fig.~\ref{fig:rcrit-vs-depth}) \\[3pt]
    \emph{P-III.} Better signals (lower $\epsilon_0$) and protection (lower effective $r$) both delay collapse & Both reduce $\beta$ & CATIS ablation (Table~\ref{tab:ablation}) \\
    \bottomrule
  \end{tabular}
  \caption*{Predictions from the distortion recurrence under monotone feedback coupling.}
\end{table*}

The linear coupling is the mildest form reproducing the observed cliff-like transition; sub-linear alternatives produce polynomial growth inconsistent with the abrupt collapse (Appendix~\ref{app:feedback-loop}).

\subsection{Component 2: Pairwise Signal Degradation Sets a Low Trigger Threshold}
\label{sec:pairwise-degradation}

The amplifier triggers when per-layer signal quality drops below a threshold. We quantify this through two complementary diagnostic statistics computed per layer, both reusable across methods and architectures.

\textbf{Diagnostic statistic ($\rho_s$).} For each image, we compute the $N_p \times N_p$ pairwise similarity matrix $\mathbf{S}$ from both its clean version and a corrupted version (e.g., Gaussian noise at severity~$5$); flatten both matrices to score vectors; and take the Spearman ranking correlation between the two flattened vectors,
\begin{equation}
  \rho_s \;=\; \frac{\mathrm{Cov}(R_{\text{clean}},\, R_{\text{corr}})}{\mathrm{Std}(R_{\text{clean}}) \cdot \mathrm{Std}(R_{\text{corr}})},
  \label{eq:spearman}
\end{equation}
where $R_{\text{clean}}, R_{\text{corr}}$ are the rank-transformed similarity vectors. $\rho_s{=}1$ corresponds to identical merge rankings (decisions unaffected by perturbation); $\rho_s{\to}0$ corresponds to randomized rankings. We use algorithmic corruptions as a controlled probe; the degradation itself is architectural (driven by representation homogenization in deep layers~\citep{raghu2021vision}) and thus applies regardless of the specific form of distribution shift, as confirmed by identical collapse points on IN-A, IN-R, and IN-Sketch (Figure~\ref{fig:pareto-teaser}b). A complementary magnitude statistic, the Frobenius distance $\Delta F = \|\mathbf{S}_{\text{clean}} - \mathbf{S}_{\text{corr}}\|_F$, is reported alongside $\rho_s$ in Appendix~\ref{app:pairwise-degradation}.

\begin{figure*}[t]
  \centering
  \includegraphics[width=\textwidth]{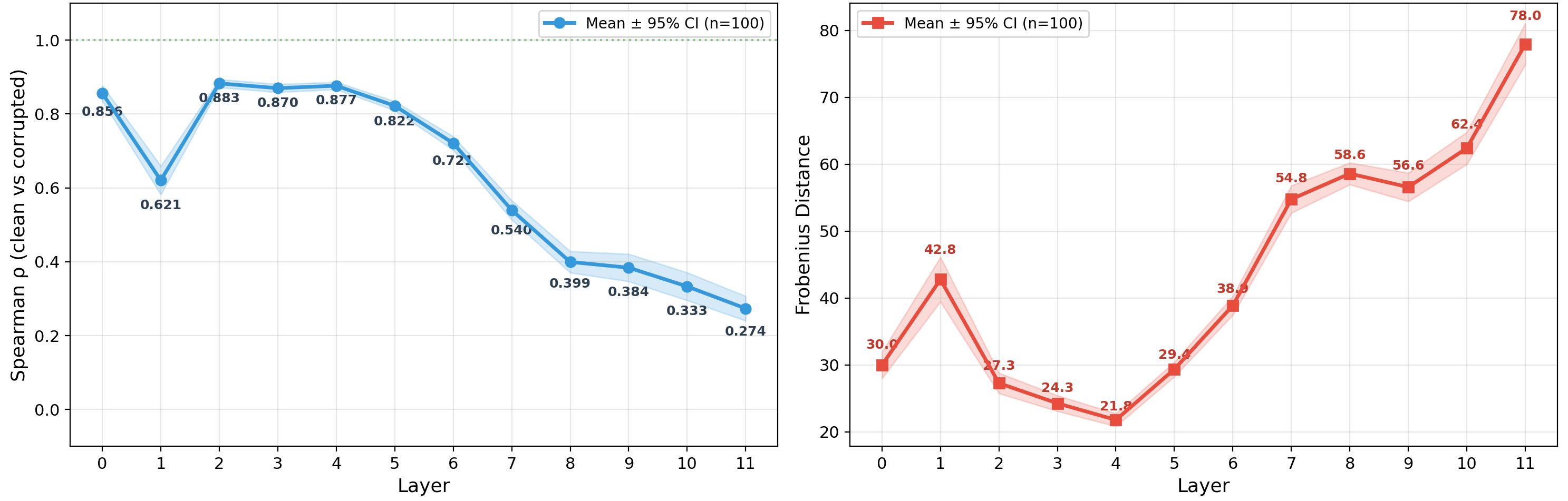}
  \caption{Pairwise similarity degradation across layers (DeiT-3 Base). $\rho_s$ drops from 0.88 to 0.27; Frobenius distance rises from $\sim$25 to $\sim$78. Definitions in Appendix~\ref{app:pairwise-degradation}.}
  \label{fig:cosine-degradation}
\end{figure*}

On DeiT-3 Base ($n{=}1000$ images; Figure~\ref{fig:cosine-degradation}), $\rho_s$ drops from $0.88$ in shallow layers to $0.27$ in deep layers, indicating that merge decisions in deep layers are essentially arbitrary; in parallel, $\Delta F$ rises from $\sim$25 to $\sim$78, showing that the matrix values themselves are severely distorted, not merely re-ranked.

Deep-layer representation homogenization is well documented~\citep{raghu2021vision}; our contribution is connecting it to the ranking stability of merge signals: as representations homogenize, pairwise rankings become arbitrary, setting a low trigger threshold for all baselines.

\subsection{Structural Limitations of Pairwise Signals}
\label{sec:paradigm}

The degradation above is not an accident of specific scoring formulas; it is an \emph{inherent property} of pairwise signal computation. Consider the distinction between pairwise and unary importance signals:

\begin{itemize}[leftmargin=*, nosep]
  \item A \textbf{pairwise} signal computes $s(i,j) = f(\mathbf{z}_i, \mathbf{z}_j)$ for each token pair. The resulting ranking depends on $\binom{N_p}{2} = O(N_p^2)$ jointly perturbed elements: a perturbation to any single token's representation affects \emph{all} $N_p{-}1$ pairwise scores involving that token.
  \item A \textbf{unary} signal computes $s(i) = g(\mathbf{z}_i, \boldsymbol{\theta})$ where $\boldsymbol{\theta}$ denotes aggregate population statistics (e.g., mean $\boldsymbol{\mu}$, variance $\boldsymbol{\sigma}^2$). Each token's score depends on its own features and on $\boldsymbol{\theta}$, which is a \emph{population average} over all $N_p$ tokens, robust to individual perturbations by the central limit theorem.
\end{itemize}

\noindent Pairwise ranking stability is governed by the variance of $O(N_p^2)$ jointly perturbed entries; unary signal stability depends on only $O(N_p)$ independently perturbed entries, with population statistics converging at rate $O(1/\sqrt{N_p})$ by the central limit theorem. As deep-layer homogenization compresses the signal-to-noise ratio of individual token representations, pairwise perturbation energy grows quadratically faster than unary perturbation energy, predicting faster ranking degradation.

We formalize this gap via a perturbation-energy analysis. Let $\{\mathbf{z}_i\}_{i=1}^{N_p} \subset \mathbb{R}^d$ be the layer-local token representations, subject to independent per-token perturbations $\boldsymbol{\xi}_i$ that aggregate the cumulative distortion from preceding layers, and let $V = \sum_{\text{scores}} \mathbb{E}[(\tilde{s} - s)^2]$ denote the total perturbation energy summed over all scores produced by the importance functional. The intuition is direct: for score vectors with comparable margin distributions, larger total perturbation energy implies more ranking inversions and therefore lower $\rho_s$.

\begin{proposition}[Perturbation-energy gap between signal classes]
\label{prop:perturbation}
Let the per-token perturbations $\boldsymbol{\xi}_i$ be zero-mean, isotropic with per-component variance $\sigma_\xi^2$, and cross-token independent. Then for a pairwise signal $s(i,j) = f(\mathbf{z}_i, \mathbf{z}_j)$ and a unary signal $s(i) = g(\mathbf{z}_i, \boldsymbol{\theta})$ with $\boldsymbol{\theta} = \boldsymbol{\theta}(\mathbf{z}_1, \ldots, \mathbf{z}_{N_p})$ a population aggregate,
\begin{equation*}
  V_{\textup{pair}} \;=\; \Theta\bigl(N_p^2 \cdot G_{\textup{pair}} \cdot \sigma_\xi^2\bigr), \qquad
  V_{\textup{unary}} \;=\; \Theta\bigl(N_p \cdot G_{\textup{unary}} \cdot \sigma_\xi^2\bigr),
\end{equation*}
with $G_{\textup{pair}}, G_{\textup{unary}}$ the mean squared gradient norms of the respective signals (under gradient-homogeneity in deep layers). Whenever $G_{\textup{pair}} / G_{\textup{unary}} = \Theta(1)$ (verified for cosine similarity and diagonal Mahalanobis, both $O(1/d)$), the gap satisfies $V_{\textup{pair}} / V_{\textup{unary}} = \Theta(N_p)$.
\end{proposition}

The gap is purely combinatorial: pairwise scoring creates $O(N_p^2)$ coupled sensitivity pathways (each token's perturbation appears in all $N_p - 1$ pairs involving it), while unary scoring creates only $O(N_p)$ pathways and damps the population-statistic perturbation by an additional $O(1/\sqrt{N_p})$ central-limit factor. The full first-order Taylor argument, the role of cross-token independence in vanishing the off-diagonal terms, and a discussion of the supporting assumptions (gradient homogeneity, isotropic covariance, comparable gradient scales, and the mapping from $V$ back to $\rho_s$ through margin distributions) are deferred to Appendix~\ref{app:perturbation-energy}.

This prediction is quantitatively confirmed: in deep layers (9--11) of DeiT-3 Base, pairwise ranking consistency drops to $\rho_s \approx 0.27$, while unary signals maintain $\rho_s \approx 0.37$--$0.48$ (Table~\ref{tab:complementarity}), a 37--78\% relative improvement. The gap is consistent across all tested models: no pairwise signal in our evaluation achieves $\rho_s$ above $0.30$ in deep layers, while no unary signal falls below $0.31$. This is a statement about \emph{signal classes}, not about any specific signal.

However, replacing pairwise signals with unary ones addresses the \emph{trigger} but not the \emph{amplifier}. PiToMe illustrates this clearly: its energy score correctly identifies \emph{similar} tokens (preserving feature-space integrity) yet achieves the \emph{worst} accuracy of all methods, because it misidentifies \emph{important} ones. Even perfect merge pairing is insufficient if triage (the decision of which tokens to protect, merge, or evict) is wrong. We formalize this distinction next.

\subsection{Structural Consequences: Two Types of Damage Diagnosed by $\rho_\text{off}$}
\label{sec:rho-off}

We introduce a second diagnostic statistic, $\rho_\text{off}$, defined as the mean absolute off-diagonal entry of the Pearson correlation matrix $\mathbf{R} \in \mathbb{R}^{d \times d}$ computed across the $N_p$ patch token features at a given layer,
\begin{equation}
  \rho_\text{off} \;=\; \frac{1}{d(d-1)} \sum_{j \neq k} \bigl|r_{jk}\bigr|,
  \label{eq:rho-off}
\end{equation}
where $r_{jk}$ is the Pearson correlation between feature dimensions $j$ and $k$ across the $N_p$ tokens. $\rho_\text{off} \approx 0$ indicates approximately independent dimensions (healthy); large $\rho_\text{off}$ signals dimension collapse, the structural signature of repeated incorrect merging. $\rho_\text{off}$ distinguishes between the two qualitatively different failure modes previewed above:

\begin{itemize}[leftmargin=*, nosep]
  \item \textbf{Structural damage} (high $\rho_\text{off}$): incorrect merges produce hybrid vectors that introduce spurious inter-dimension correlations. ToMe/MCTF reach $\rho_\text{off}{=}0.43$ at high $r$.
  \item \textbf{Semantic loss} (low $\rho_\text{off}$, low accuracy): correct merge pairing but wrong triage, exposing important tokens to reduction. PiToMe confirms this: its energy score yields $\rho_\text{off}{\approx}0.22$ (matching Vanilla, no structural damage), yet it achieves the \emph{worst} accuracy because the same pairwise signal that identifies \emph{similar} tokens misidentifies \emph{important} ones. The failure lies in triage, not pairing.
\end{itemize}

\noindent This shows that preserving feature-space integrity alone is insufficient (PiToMe): signal quality and triage must \emph{both} be correct. In terms of the distortion recurrence (Eq.~\ref{eq:recurrence}), the two failure modes map to different damage channels: structural damage corresponds to high per-operation intensity $\delta$ from merges (hybrid vectors propagate forward), while semantic loss corresponds to high distortion rate $\epsilon$ from triage errors (important tokens exposed to reduction). Together, $\rho_s$ and $\rho_\text{off}$ form a complete diagnostic protocol: $\rho_s$ predicts \emph{when} collapse occurs, $\rho_\text{off}$ diagnoses \emph{what type}. Both are applicable to any future method.

\subsection{Design Principles Derived from Diagnosis}
\label{sec:principles}

The diagnostic framework yields three design principles, each addressing a specific diagnosed failure mode:

\begin{itemize}[leftmargin=*, nosep]
  \item \textbf{P1} (from Section~\ref{sec:paradigm}): Replace pairwise signals with \emph{unary} signals to raise the amplifier's trigger threshold ($\rho_s{:}\;0.27 \to 0.37$--$0.48$ in deep layers).
  \item \textbf{P2} (from Table~\ref{tab:complementarity}): Fuse \emph{complementary} unary signals that peak in different depth ranges, maximizing $\rho_s$ across all layers.
  \item \textbf{P3} (from Section~\ref{sec:rho-off}): Introduce a \emph{triage mechanism} that protects high-confidence tokens from any operation and routes uncertain tokens to the less damaging operation, suppressing the amplification gain (reducing effective $r$ and substituting $\delta_e$ for $\delta_m$ in Eq.~\ref{eq:recurrence}).
\end{itemize}

\noindent P1--P3 define a \emph{design space}, not a unique method: any approach that adopts unary signals, fuses complementary depth profiles, and incorporates triage should delay the collapse. CATIS (Section~\ref{sec:method}) is one concrete instantiation; its purpose is to validate the diagnosis by showing that a method following P1--P3 does, in fact, eliminate the collapse. Full diagnostic derivations and extended $\rho_\text{off}$ figures are in Appendix~\ref{app:diagnosis-full} (Figure~\ref{fig:rho-off-main}; Appendix~\ref{app:rho-off-full}, Figures~\ref{fig:rho-off-all-s}--\ref{fig:rho-off-all-vl}; vanilla validation: Figures~\ref{fig:rho-off-vanilla-a} and~\ref{fig:rho-off-vanilla-b}).

\section{CATIS: Constructive Validation of the Diagnosis}
\label{sec:method}

\begin{figure*}[t]
  \centering
  \includegraphics[width=\textwidth]{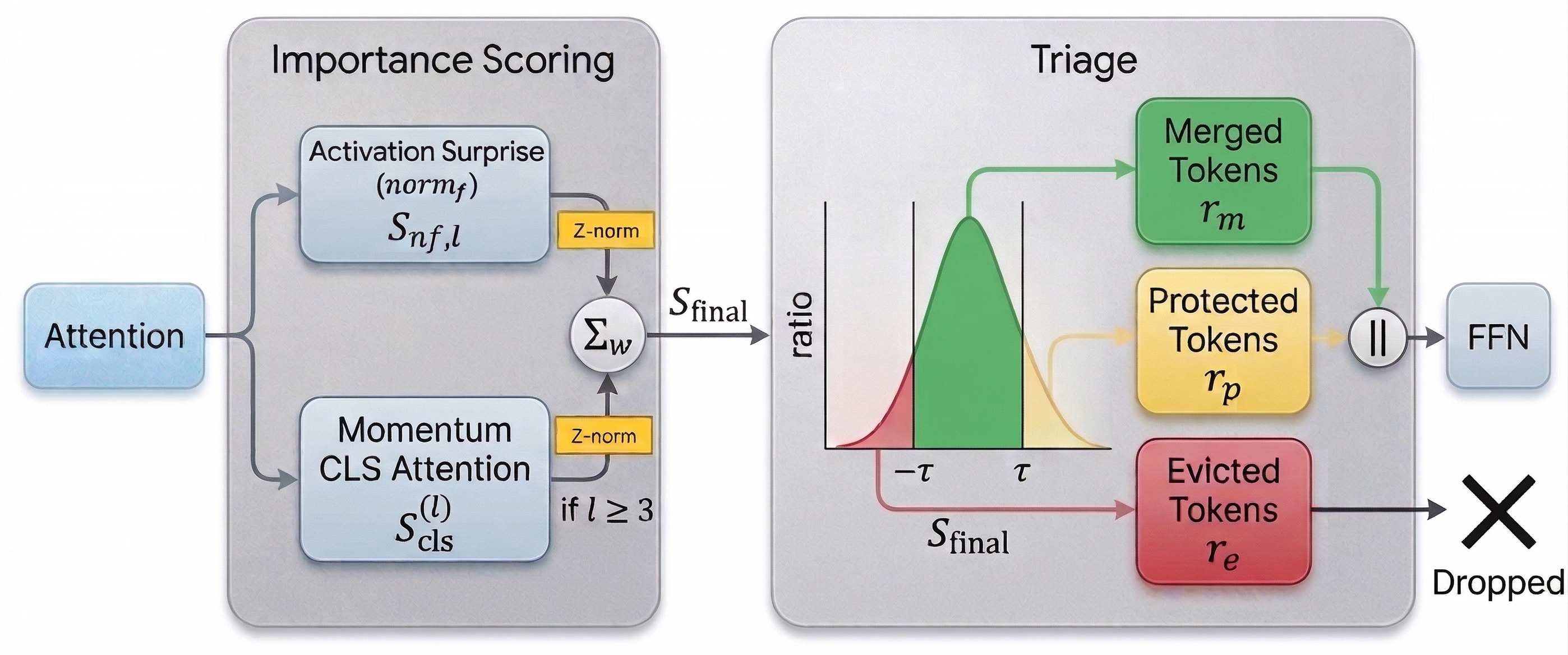}
  \caption{CATIS pipeline (one transformer layer). Two unary signals are z-normalized and fused into a layer-local importance functional; a symmetric, scale-adaptive criterion partitions tokens into protect/merge/evict subsets, after which the merge subset is processed by bipartite matching restricted to that subset. The momentum branch activates only after sufficient cross-layer history has accumulated; the activation depth, fusion weighting, partition radius, and intra-budget allocation are all model-family-specific and selected by the procedure outlined in Section~\ref{sec:experiments}.}
  \label{fig:pipeline}
\end{figure*}

A constructive instantiation of P1--P3 takes the form of a layer-local operator $\mathcal{O}(\cdot;\, l)$ that acts on the token set: it (i)~assembles a fused unary scoring functional from a population-anchored statistic and a momentum-augmented context signal, satisfying the unary regime of Proposition~\ref{prop:perturbation} simultaneously across shallow, middle, and deep layers; (ii)~partitions the tokens into three structurally distinct subsets via a symmetric, scale-adaptive criterion in standardized score space; and (iii)~routes each subset to the operation whose damage intensity (in the dual-channel recurrence of Section~\ref{sec:rho-off} and Eq.~\ref{eq:dual-channel}) is matched to the subset's expected estimation error. Figure~\ref{fig:pipeline} renders the operator schematically; the remainder of this section formalizes each component while deferring all numerical schedules to a separate optimization stage.

\subsection{Signal 1: Population-Anchored Activation Functional \textnormal{[P1: Unary Statistic]}}
\label{sec:norm-f}

\textbf{From P1 to a population-anchored functional.} P1 demands a scoring functional whose perturbation sensitivity falls in the unary regime of Proposition~\ref{prop:perturbation} (Section~\ref{sec:paradigm}); we obtain such a functional by recasting token importance as a per-image, class-agnostic anomaly-detection problem in the layer-local representation space. Tokens deviating most from the layer's population distribution carry unique visual evidence~\citep{darcet2023vision}; in contrast to Mahalanobis-style OOD detection~\citep{hendrycks2016baseline, lee2018simple}, which depends on class-conditional training-set moments, our construction conditions only on the layer-local token population $\mathcal{P}_l = \{\mathbf{z}_i\}_{i=1}^{N_p}$ and is therefore deployable on any pretrained backbone without auxiliary statistics, retraining, or any class label.

The canonical population-distance is the Mahalanobis quadratic form
\begin{equation}
  M_i = \sqrt{(\mathbf{z}_i - \boldsymbol{\mu})^{\top}\, \boldsymbol{\Sigma}^{-1}\, (\mathbf{z}_i - \boldsymbol{\mu})},
  \label{eq:mahal-main}
\end{equation}
with $(\boldsymbol{\mu}, \boldsymbol{\Sigma})$ the empirical first and second moments of $\mathcal{P}_l$. This quadratic form is structurally unary in the sense of Section~\ref{sec:paradigm}: the focal token $\mathbf{z}_i$ enters explicitly through the centred deviation, while $(\boldsymbol{\mu}, \boldsymbol{\Sigma})$ are population aggregates whose individual perturbation sensitivity is $O(1/N_p)$ by the indirect-pathway analysis of Appendix~\ref{app:perturbation-energy} (Eq.~\ref{eq:unary-taylor}).

\textbf{Two structural reductions.} Two architectural facts shape the functional we actually deploy. First, for all standard ViT configurations the patch-token count satisfies $N_p < d$, so $\mathrm{rank}(\boldsymbol{\Sigma}) \le N_p - 1 < d$ and the empirical covariance is singular; even a pseudo-inverse incurs $O(d^3)$ per layer per image, which contradicts the acceleration objective. Second, a diagnostic measurement of off-diagonal correlation on vanilla models (Section~\ref{sec:rho-off}; Appendix~\ref{app:rho-off-vanilla}) shows that $\rho_\text{off}$ stays uniformly below a small architectural ceiling across all tested models, datasets, and layer depths, with cross-dataset variance two orders of magnitude smaller than the within-architecture mean, indicating that approximate decorrelation of feature dimensions is an architectural property of standard ViTs rather than a data-dependent coincidence. These two facts jointly motivate, and the second one quantitatively justifies, a diagonal reduction $\boldsymbol{\Sigma} \mapsto \mathrm{diag}(\sigma_1^2, \ldots, \sigma_d^2)$, which collapses $M_i$ to a per-dimension z-score norm
\begin{equation}
  s_{\text{nf}}(\mathbf{z}_i;\, \mathcal{P}_l) \;=\; \bigl\| (\mathbf{z}_i - \boldsymbol{\mu}) \oslash \boldsymbol{\sigma} \bigr\|_2,
  \label{eq:norm-f}
\end{equation}
with $\oslash$ element-wise division and $\boldsymbol{\sigma} = (\sigma_1, \ldots, \sigma_d)^\top$. The computation is $O(N_p \cdot d)$, contains no pairwise term, and the diagnostic predicate $\rho_\text{off}$ remains as the structural reusability check that any future architecture must pass before reusing this functional (Appendix~\ref{app:limitations}).

\textbf{Trigger-threshold consequence.} Substituting $s_{\text{nf}}$ into the unary branch of Proposition~\ref{prop:perturbation} yields a perturbation-energy budget of order $\Theta(N_p \cdot G_{\text{unary}} \cdot \sigma_\xi^2)$, a $\Theta(N_p)$-factor improvement over any pairwise alternative built on the same representations whose comparable gradient norm is of the same order. By the gradient-homogeneity argument of Appendix~\ref{app:perturbation-energy}, the precondition $G_{\text{pair}}/G_{\text{unary}} = \Theta(1)$ holds for the concrete signals used here (cosine similarity and diagonal Mahalanobis are both $O(1/d)$), so the gap materializes empirically as the deep-layer ranking-consistency lift visible in Table~\ref{tab:complementarity}. The amplifier itself (Section~\ref{sec:feedback-loop}) is not altered; rather, its trigger threshold is raised by lowering $\epsilon_0$ in the closed-form solution of Eq.~\ref{eq:closed-form}.

\subsection{Signal 2: Momentum-Augmented Context Statistic \textnormal{[P2: Cross-Layer Complementary Fusion]}}
\label{sec:cls}

\textbf{From P2 to a depth-complementary functional.} Table~\ref{tab:complementarity} establishes a structural fact that motivates P2: no single unary functional maintains high $\rho_s$ across all depth bands. The activation functional of Section~\ref{sec:norm-f} is strongest in shallow layers, where layer-local population statistics are still discriminative, but its margin narrows as deep-layer homogenization compresses the per-dimension z-score spectrum. The contextual aggregate produced by the model's CLS pathway~\citep{devlin2019bert} carries complementary information, peaking in middle layers precisely where the activation functional dips, but it is also itself unstable in both extremes of the depth axis (with substantial run-to-run dispersion in shallow layers, and a stability collapse in deep layers that mirrors the pairwise case). Either signal, used alone, fails P2.

\textbf{First-order temporal stabilisation.} We turn the second signal into a P2-compatible component by augmenting the raw multi-head averaged CLS-to-patch attention $A_{\text{cls}}^{(l)}$ with a first-order discrete temporal derivative across the depth axis. Treating the depth index $l$ as a discrete temporal coordinate and the per-token CLS-attention vector as a depth-indexed trajectory, the elementary backward finite-difference operator $\nabla_l A_{\text{cls}}^{(l)} \;=\; A_{\text{cls}}^{(l)} - A_{\text{cls}}^{(l-1)}$ supplies the only Taylor remainder available without future-layer information, and the affine pencil $\{A_{\text{cls}}^{(l)} + \gamma\,\nabla_l A_{\text{cls}}^{(l)} : \gamma \in \mathbb{R}_{\ge 0}\}$ spans, at first order, every monotone trajectory-aware predictor parameterised by a single scalar; collecting like terms, the linear-extrapolation form
\begin{equation}
  s_{\text{cls}}^{(l)} \;=\; (1+\gamma)\, A_{\text{cls}}^{(l)} \;-\; \gamma\, A_{\text{cls}}^{(l-1)}
  \;=\; A_{\text{cls}}^{(l)} + \gamma\,\nabla_l A_{\text{cls}}^{(l)},
  \label{eq:momentum}
\end{equation}
admits two equivalent readings: as a $(1{+}\gamma, -\gamma)$-weighted combination of the present and one-step-prior CLS-to-patch attention vectors, and as the present snapshot offset by a $\gamma$-scaled discrete temporal derivative whose sign decides whether the coefficient pencil is interpolative ($\gamma{=}0$, single-layer baseline), strictly extrapolative ($\gamma{>}0$, present amplified, past subtracted), or, in the limit $\gamma \to \infty$, dominated by the depth-derivative term up to normalization. Because the two coefficients sum identically to unity, the operator is a first-order discrete momentum predictor in the Heun--Euler sense and preserves the row-sum normalization of $A_{\text{cls}}$ exactly. The scalar $\gamma$ is layer-shared and model-family-specific (its admissible range is reported in Appendix~\ref{app:hparams}); operationally, increasing $\gamma$ progressively amplifies the trend component so that rising-importance tokens are up-weighted and declining ones down-weighted, converting a single-layer snapshot, which cannot distinguish a token whose contextual relevance is increasing from one that is collapsing, into a trajectory-aware estimate.

\textbf{Cross-class fusion.} The two functionals live on incommensurable scales: $s_{\text{nf}}^{(l)}$ is an unbounded $L_2$ norm in $\mathbb{R}_{\ge 0}$ whose layer-wise dynamic range tracks the spectral radius of $\boldsymbol{\Sigma}^{1/2}$, while $s_{\text{cls}}^{(l)}$ is a bounded probabilistic quantity living, after row-sum normalization, in $[0,1]$ with mass concentrated in $O(1/N_p)$. A naive sum or product of the two is therefore ill-posed, since the dominant component is dictated by the chosen scale rather than by relative informativeness. To resolve this, both signals are first projected onto a common, scale-invariant axis by per-image z-score normalization within the layer,
\begin{equation}
  \hat{s}_{\bullet}^{(l)} \;=\; \frac{s_{\bullet}^{(l)} - \mathrm{mean}_i\, s_{\bullet,i}^{(l)}}{\mathrm{std}_i\, s_{\bullet,i}^{(l)}},
  \quad \bullet \in \{\text{nf}, \text{cls}\},
  \label{eq:zscore}
\end{equation}
mapping each functional into a zero-mean, unit-variance distribution along the token axis. The two standardised score vectors are then fused via a single-parameter affine combination on the standardised manifold,
\begin{equation}
  s_{\text{final}}^{(l)}
  \;=\; w_{\text{cls}}\, \hat{s}_{\text{cls}}^{(l)}
       \,+\, (1 - w_{\text{cls}})\, \hat{s}_{\text{nf}}^{(l)},
  \quad w_{\text{cls}} \in [0, 1],
  \label{eq:fusion}
\end{equation}
which is the unique convex combination that (i) preserves the standardised scale (the fused vector retains zero mean and a variance bounded above by~$1$ once the two inputs are at most weakly correlated), (ii) reduces continuously to either single-functional baseline at the endpoints $w_{\text{cls}} \in \{0, 1\}$, and (iii) admits a closed-form Pareto-front sweep along $w_{\text{cls}}$ for diagnostic ablation (Appendix~\ref{app:hparams}). The mixing coefficient $w_{\text{cls}}$ is held constant within each model family and selected by the family-level diagnostic procedure of Section~\ref{sec:experiments}; its admissible interval is governed by the relative depth-band complementarity between the two functionals (Table~\ref{tab:complementarity}) rather than by any fixed numerical recipe. The momentum branch in Eq.~\ref{eq:momentum} is suppressed in the first few transformer blocks, $l < L_{\text{start}}$, until enough cross-layer history has accumulated for the discrete temporal-derivative term to be meaningful; below this depth the operator collapses to its activation-only branch ($\gamma_{\text{eff}} = 0$). The precise activation depth $L_{\text{start}}$ is, again, model-family-specific (Appendix~\ref{app:hparams}). Table~\ref{tab:complementarity} confirms the targeted complementarity: the fused signal exceeds both single-functional baselines in layer-averaged ranking consistency, with the activation functional stabilising the deep layers where the contextual functional weakens, and the contextual functional lifting the middle layers where the activation functional dips. A qualitative visualization of this complementarity on a representative image appears in Appendix~\ref{app:signal-heatmaps} (Figure~\ref{fig:signal-heatmaps}).

\begin{table}[t]
  \caption{Ranking stability ($\rho_s$) of unary signals across layer depth (DeiT-3 Base, $n{=}1{,}000$).}
  \label{tab:complementarity}
  \centering
  \small
  \begin{tabular}{lccc}
    \toprule
    Layer depth & norm$_F$ $\rho_s$ & CLS $\rho_s$ & Fused $\rho_s$ \\
    \midrule
    Shallow (0--2) & \textbf{0.70--0.74} & 0.44--0.76 & 0.71--0.75 \\
    Middle (4--7) & 0.46--0.54 & \textbf{0.57--0.61} & 0.50--0.57 \\
    Deep (9--11) & \textbf{0.37--0.45} & 0.31--0.42 & 0.39--0.48 \\
    \midrule
    Layer-averaged & 0.534 & 0.535 & \textbf{0.559} \\
    \bottomrule
  \end{tabular}
\end{table}

\subsection{Triage Operator \textnormal{[P3: Suppressing the Amplification Gain]}}
\label{sec:triage}

\textbf{From P3 to a partitioning operator.} The diagnosis of Section~\ref{sec:diagnosis} shows that even when P1 and P2 lift the per-layer signal quality, the error amplifier of Section~\ref{sec:feedback-loop} remains \emph{structurally active} because it is signal-agnostic: it functions as soon as any reduction operation is applied layer by layer, and its closed-form trajectory $\Delta^{(l)} = (\epsilon_0/\alpha)(\beta^l - 1)$ depends on the per-layer reduction rate $r$ regardless of which signal selects the tokens (Appendix~\ref{app:feedback-loop}, Eq.~\ref{eq:closed-form}). P3 therefore mandates a separate intervention that acts on the \emph{operation set itself}, not on the importance ranking, by directly reducing the effective per-layer reduction rate seen by the amplifier and by routing the residual rate through the lower-damage operation channel.

\textbf{Symmetric, scale-adaptive partitioning.} Concretely, the layer-local importance functional $s_{\text{final}}^{(l)}$ defines a symmetric, scale-adaptive triage operator that partitions the token index set into three structurally distinct subsets, $\mathcal{P}^{(l)} \,\dot\cup\, \mathcal{M}^{(l)} \,\dot\cup\, \mathcal{E}^{(l)}$, corresponding respectively to high-confidence \emph{protected} tokens (exempt from any reduction operation, fully preserved across the layer), low-confidence \emph{eviction candidates}, and intermediate-confidence \emph{merge candidates}. Because $s_{\text{final}}^{(l)}$ is per-image z-normalized within the layer (Eq.~\ref{eq:zscore}, Section~\ref{sec:cls}), every absolute decision boundary on the score axis can be re-expressed as a level set in standardised units; we therefore parameterise the triage operator by a single non-negative scalar $\tau \in \mathbb{R}_{\ge 0}$ defining a symmetric pair of upper and lower super-/sub-level sets,
\begin{equation}
  \begin{aligned}
    \mathcal{P}^{(l)} &= \bigl\{\,i : s_{\text{final},i}^{(l)} > \,\,\,\tau\,\bigr\}, \\
    \mathcal{E}^{(l)} &= \bigl\{\,i : s_{\text{final},i}^{(l)} < -\tau\,\bigr\}, \\
    \mathcal{M}^{(l)} &= \bigl\{\,i : -\tau \le s_{\text{final},i}^{(l)} \le \tau\,\bigr\},
  \end{aligned}
  \label{eq:triage}
\end{equation}
where the explicit symmetry $\mathcal{E}^{(l)} = -\mathcal{P}^{(l)}$ on the standardised axis ensures that the partition is invariant to global location/scale shifts of the underlying signal, so the boundary radius $\tau$ admits a single absolute scale per model family that need not be re-tuned per image, per layer, or per dataset. The total per-layer reduction budget $r$ is then split between the eviction and merge channels via the model-family-specific allocation policy
\begin{equation}
  r_e^{(l)} \;=\; \min\!\bigl(\,\bigl\lfloor \mathrm{evict\_ratio} \cdot r \bigr\rfloor,\; |\mathcal{E}^{(l)}|\,\bigr),
  \quad r_m^{(l)} \;=\; r - r_e^{(l)},
  \label{eq:allocation}
\end{equation}
with $\mathrm{evict\_ratio} \in [0,1]$ a single layer-shared scalar and $r_e^{(l)} + r_m^{(l)} = r$ by construction. The $\min(\cdot)$ saturation is a hard upper bound enforcing that the eviction channel never demands more tokens than the eviction subset can supply at the current layer; whenever $|\mathcal{E}^{(l)}|$ falls below $\lfloor \mathrm{evict\_ratio} \cdot r \rfloor$, the slack is automatically routed back to the merge channel, so the triage operator degrades smoothly to the merge-only baseline in the limit $\mathcal{E}^{(l)} \to \emptyset$. The eviction channel is therefore prioritised up to a model-family-specific fraction of the budget while bounded by the size of $\mathcal{E}^{(l)}$, and the residual rate is routed through the merge channel via bipartite matching restricted to $\mathcal{M}^{(l)}$.

\textbf{Channel asymmetry and the dual-channel recurrence.} The triage design reflects a key asymmetry diagnosed in Section~\ref{sec:rho-off}: incorrect merge produces hybrid vectors that persist and participate in subsequent attention computations, propagating structural damage ($\rho_\text{off}$ rises sharply for ToMe and MCTF in deep layers and at high $r$); incorrect eviction causes information loss but does \emph{not} pollute surviving representations, leaving $\rho_\text{off}$ at the vanilla level. The two operations therefore differ in their per-operation damage intensity, with $\delta_e < \delta_m$. This asymmetry promotes the single-channel recurrence of Eq.~\ref{eq:recurrence} into a dual-channel recurrence of the form
\begin{equation}
  \Delta_{\text{CATIS}}^{(l+1)} \;=\; \Delta_{\text{CATIS}}^{(l)} \;+\; \epsilon_m^{(l)} \cdot r_m^{(l)} \cdot \delta_m \;+\; \epsilon_e^{(l)} \cdot r_e^{(l)} \cdot \delta_e,
  \label{eq:dual-channel}
\end{equation}
in which $r_m^{(l)} + r_e^{(l)} = r\,(1 - f_p^{(l)})$ once the protected fraction $f_p^{(l)}$ has been removed from circulation. The triage operator routes the lowest-confidence tokens (those with the highest expected $\epsilon$) into the eviction channel (lower $\delta$) and reserves the merge channel for tokens of intermediate, comparable importance, which jointly minimises the total per-layer distortion increment. Three independent intervention surfaces on the recurrence are thereby active simultaneously: (i)~lower $\epsilon_0$ from P1+P2 (via fused unary signals), (ii)~lower effective $r$ from P3 protection (via $f_p^{(l)}$), and (iii)~routing the high-$\epsilon$ tail from $\delta_m$ to $\delta_e$ via channel allocation. The interplay among these three surfaces is what raises the proportionality constant of the $r_{\text{crit}} \propto 1/L$ trend (Section~\ref{sec:feedback-loop}; Figure~\ref{fig:rcrit-vs-depth}); a per-component decomposition appears in the ablation of Section~\ref{sec:ablation}.

\textbf{Why pairwise matching within the merge subset remains safe.}
Our diagnosis targets the \emph{importance scoring} stage, where the pairwise-vs.-unary perturbation-energy gap of Proposition~\ref{prop:perturbation} sets the trigger threshold; the bipartite matching used \emph{within} the merge subset $\mathcal{M}^{(l)}$ serves an entirely different purpose (deciding \emph{which similar tokens to combine} among tokens already triaged as expendable). Triage pre-filtering restricts this pool to intermediate-importance tokens, reducing the number of jointly perturbed pairwise entries from $O(N_p^2)$ to $O(|\mathcal{M}^{(l)}|^2)$ and substantially raising pool-internal pairwise $\rho_s$ in deep layers (Appendix~\ref{app:pool-rho-s}). Two additional factors bound the per-error damage: (i)~all tokens in $\mathcal{M}^{(l)}$ have comparable importance under the triage criterion, so even worst-case mismatches pair tokens of similar semantic value and bound $\delta_m$ per error far below the global worst case; (ii)~the resulting hybrid vectors are produced from a homogeneous semantic subset and thus introduce smaller spurious inter-dimension correlations than would arise from a global pairing.

\section{Experiments: Verifying the Diagnostic Predictions}
\label{sec:experiments}

Experiments verify the three predictions (P-I--P-III) and confirm that CATIS eliminates the collapse.

\subsection{Setup}

We evaluate on 7~ViT models (ViT-S/B/L~\citep{dosovitskiy2020image}, DeiT-3-S/B/L~\citep{touvron2022deit}, and DINOv3~\citep{simeoni2025dinov3, caron2021emerging, oquab2023dinov2}) using pretrained weights from supervised~\citep{dosovitskiy2020image, touvron2022deit} and self-supervised~\citep{he2022masked} training, across ImageNet-1K~\citep{deng2009imagenet, russakovsky2015imagenet} and four robustness benchmarks: ImageNet-A~\citep{hendrycks2021natural}, -R~\citep{hendrycks2021many}, -Sketch~\citep{wang2019learning}, and -C~\citep{hendrycks2019benchmarking} (Appendix~\ref{app:full-results}). Shifted benchmarks are included because prior evaluation relies on clean data, masking collapse to single-digit accuracy under shift.
Baselines are ToMe~\citep{bolya2022token}, ToFu~\citep{kim2024token}, PiToMe~\citep{tran2024accelerating}, and MCTF~\citep{lee2024multi}, all training-free; training-based methods~\citep{rao2021dynamicvit, chen2023diffrate} occupy a different design point. All methods are deterministic given fixed pretrained weights and evaluation data; results are exactly reproducible from a single run. We report Top-1 accuracy and GFLOPs; Shift avg $=$ mean(IN-A, IN-R, IN-Sketch), excluding IN-C because it uses a different metric (mCA rather than Top-1 accuracy).

\subsection{Main Results}
\label{sec:main-results}

Table~\ref{tab:main} reports per-dataset accuracies for all 7~models. CATIS achieves the best accuracy across all 35 model$\times$dataset combinations tested. The advantage scales with model depth: $+$0.3pp Shift avg on ViT-Small ($L{=}12$) vs.\ $+$11.8pp on DeiT-3-Base ($L{=}12$, higher $r$) and $+$10.4pp on DeiT-3-Large ($L{=}24$).

\begin{table*}[t]
  \caption{Full per-dataset results (\%) for all 7 models. Top-1 accuracy except IN-C (mCA). Shift avg $=$ mean(IN-A, IN-R, IN-Sketch). Best reduced result in \textbf{bold}. Throughput: Appendix~\ref{app:throughput}.}
  \label{tab:main}
  \centering
  \small
  \setlength{\tabcolsep}{3pt}
  \begin{tabular}{@{}ll c ccccc c@{}}
    \toprule
    Model ($r$) & Method & GFLOPs & IN-1K & IN-A & IN-R & IN-SK & IN-C & Shift avg \\
    \midrule
    \multirow{6}{*}{\shortstack[l]{ViT-S\\$r{=}13$}}
    & \textcolor{gray}{Vanilla} & \textcolor{gray}{4.6} & \textcolor{gray}{73.59} & \textcolor{gray}{14.35} & \textcolor{gray}{35.07} & \textcolor{gray}{14.71} & \textcolor{gray}{20.27} & \textcolor{gray}{21.38} \\
    & ToMe & 2.7 & 70.64 & 12.97 & 34.33 & 14.67 & 18.89 & 20.66 \\
    & ToFu & 2.7 & 69.19 & 13.48 & 31.68 & 13.97 & 16.84 & 19.71 \\
    & PiToMe & 2.7 & 66.60 & 12.28 & 30.84 & 14.13 & 15.40 & 19.08 \\
    & MCTF & 2.7 & 70.55 & 12.44 & 33.42 & 13.46 & 17.78 & 19.77 \\
    & \textbf{CATIS} & 2.7 & \textbf{70.93} & \textbf{13.93} & \textbf{34.27} & \textbf{14.74} & \textbf{18.90} & \textbf{20.98} \\
    \midrule
    \multirow{6}{*}{\shortstack[l]{ViT-B\\$r{=}13$}}
    & \textcolor{gray}{Vanilla} & \textcolor{gray}{17.6} & \textcolor{gray}{76.50} & \textcolor{gray}{21.21} & \textcolor{gray}{41.24} & \textcolor{gray}{16.59} & \textcolor{gray}{25.03} & \textcolor{gray}{26.35} \\
    & ToMe & 10.4 & 74.25 & 19.55 & 40.95 & 16.85 & 23.13 & 25.78 \\
    & ToFu & 10.4 & 73.74 & 20.76 & 39.01 & 16.41 & 21.73 & 25.39 \\
    & PiToMe & 10.4 & 72.22 & 18.92 & 38.55 & 16.60 & 20.17 & 24.69 \\
    & MCTF & 10.4 & 74.19 & 18.97 & 40.50 & 16.17 & 21.84 & 25.21 \\
    & \textbf{CATIS} & 10.4 & \textbf{74.61} & \textbf{20.93} & \textbf{41.04} & \textbf{16.89} & \textbf{23.51} & \textbf{26.29} \\
    \midrule
    \multirow{6}{*}{\shortstack[l]{ViT-L\\$r{=}11$}}
    & \textcolor{gray}{Vanilla} & \textcolor{gray}{61.6} & \textcolor{gray}{83.64} & \textcolor{gray}{43.27} & \textcolor{gray}{60.89} & \textcolor{gray}{45.55} & \textcolor{gray}{41.91} & \textcolor{gray}{49.90} \\
    & ToMe & 22.8 & 43.14 & 6.57 & 24.86 & 19.25 & 15.19 & 16.89 \\
    & ToFu & 22.8 & 20.47 & 3.00 & 10.95 & 7.44 & 2.66 & 7.13 \\
    & PiToMe & 22.8 & 65.49 & 21.72 & 39.26 & 29.92 & 18.10 & 30.30 \\
    & MCTF & 22.8 & 43.08 & 6.00 & 25.26 & 20.15 & 14.54 & 17.14 \\
    & \textbf{CATIS} & 22.8 & \textbf{81.03} & \textbf{43.65} & \textbf{59.76} & \textbf{45.12} & \textbf{33.77} & \textbf{49.51} \\
    \midrule
    \multirow{6}{*}{\shortstack[l]{DeiT-3-S\\$r{=}23$}}
    & \textcolor{gray}{Vanilla} & \textcolor{gray}{4.6} & \textcolor{gray}{81.13} & \textcolor{gray}{23.27} & \textcolor{gray}{46.54} & \textcolor{gray}{35.42} & \textcolor{gray}{29.06} & \textcolor{gray}{35.08} \\
    & ToMe & 1.7 & 42.40 & 4.07 & 16.99 & 14.60 & 8.26 & 11.89 \\
    & ToFu & 1.7 & 28.94 & 3.49 & 11.11 & 7.92 & 4.63 & 7.51 \\
    & PiToMe & 1.7 & 45.11 & 8.52 & 18.22 & 12.60 & 6.34 & 13.11 \\
    & MCTF & 1.7 & 44.22 & 4.80 & 16.49 & 13.72 & 8.94 & 11.67 \\
    & \textbf{CATIS} & 1.7 & \textbf{58.20} & \textbf{10.04} & \textbf{25.26} & \textbf{20.12} & \textbf{12.38} & \textbf{18.47} \\
    \midrule
    \multirow{6}{*}{\shortstack[l]{DeiT-3-B\\$r{=}23$}}
    & \textcolor{gray}{Vanilla} & \textcolor{gray}{17.6} & \textcolor{gray}{83.38} & \textcolor{gray}{36.75} & \textcolor{gray}{36.80} & \textcolor{gray}{54.07} & \textcolor{gray}{41.05} & \textcolor{gray}{42.54} \\
    & ToMe & 6.3 & 62.16 & 8.89 & 30.57 & 24.93 & 17.43 & 21.46 \\
    & ToFu & 6.3 & 47.96 & 7.65 & 22.15 & 17.01 & 11.57 & 15.60 \\
    & PiToMe & 6.3 & 59.46 & 17.53 & 29.13 & 20.74 & 13.92 & 22.47 \\
    & MCTF & 6.3 & 63.50 & 9.89 & 31.41 & 25.62 & 17.75 & 22.31 \\
    & \textbf{CATIS} & 6.3 & \textbf{73.85} & \textbf{25.44} & \textbf{43.96} & \textbf{33.44} & \textbf{26.21} & \textbf{34.28} \\
    \midrule
    \multirow{6}{*}{\shortstack[l]{DeiT-3-L\\$r{=}11$}}
    & \textcolor{gray}{Vanilla} & \textcolor{gray}{61.6} & \textcolor{gray}{84.47} & \textcolor{gray}{45.13} & \textcolor{gray}{57.39} & \textcolor{gray}{44.19} & \textcolor{gray}{41.55} & \textcolor{gray}{48.90} \\
    & ToMe & 22.8 & 66.07 & 9.93 & 31.33 & 27.87 & 19.51 & 23.04 \\
    & ToFu & 22.8 & 39.85 & 7.07 & 12.39 & 11.63 & 8.24 & 10.36 \\
    & PiToMe & 22.8 & 65.08 & 23.64 & 35.03 & 25.96 & 16.38 & 28.21 \\
    & MCTF & 22.8 & 68.31 & 10.69 & 33.13 & 28.87 & 20.30 & 24.23 \\
    & \textbf{CATIS} & 22.8 & \textbf{75.27} & \textbf{26.11} & \textbf{42.64} & \textbf{35.12} & \textbf{29.36} & \textbf{34.62} \\
    \midrule
    \multirow{6}{*}{\shortstack[l]{DINOv3~\citep{simeoni2025dinov3}\\$r{=}4$}}
    & \textcolor{gray}{Vanilla} & \textcolor{gray}{1349} & \textcolor{gray}{87.53} & \textcolor{gray}{86.53} & \textcolor{gray}{92.16} & \textcolor{gray}{71.50} & \textcolor{gray}{65.36} & \textcolor{gray}{83.40} \\
    & ToMe & 815.8 & 86.95 & 83.76 & 91.41 & 70.71 & 59.80 & 81.96 \\
    & ToFu & 815.8 & 86.46 & 83.91 & 90.37 & 70.44 & 54.32 & 81.57 \\
    & PiToMe & 814.1 & 86.71 & 83.41 & 91.24 & 70.83 & 56.92 & 81.83 \\
    & MCTF & 815.8 & 87.00 & 83.44 & 91.52 & 71.05 & 59.81 & 82.00 \\
    & \textbf{CATIS} & 815.8 & \textbf{87.09} & \textbf{86.15} & \textbf{91.67} & \textbf{71.11} & \textbf{59.94} & \textbf{82.98} \\
    \bottomrule
  \end{tabular}
\end{table*}

\textbf{Verification of P-I--P-II.}
The Pareto curves (Figure~\ref{fig:pareto-teaser}; full set in Appendix~\ref{app:pareto-full}) confirm convex degradation for all baselines (P-I), while Figure~\ref{fig:rcrit-vs-depth} verifies $r_{\text{crit}} \propto 1/L$ quantitatively ($R^2 > 0.9$).
Collapse thresholds depend on model depth, not data distribution.

\textbf{Verification of P-III.}
On DeiT-3-Large ($r{=}11$, 63\% GFLOPs reduction), baselines collapse to 39--68\% IN-1K while CATIS retains 75.27\% ($-$9.2pp from vanilla). DeiT-3-Base ($r{=}23$, 64\% reduction) shows the same pattern: +11.8pp on Shift avg over the best baseline.
The effect is even more pronounced on ViT-Large ($r{=}11$), where CATIS retains 96.9\% of vanilla accuracy (81.03 vs.\ 83.64) while all baselines collapse to 43--65\% (Table~\ref{tab:main}).
At ViT-Small $r{=}13$, margins narrow to $\sim$0.3pp, confirming the framework's prediction that the signal class matters less when the amplifier is below threshold.
On DINOv3~\citep{simeoni2025dinov3}, CATIS achieves +2.4pp on IN-A over ToMe, confirming generalization across pretraining strategies (Appendix~\ref{app:detailed-results}).
Wall-clock throughput on A100 GPUs matches the fastest baselines at equivalent GFLOPs (Appendix~\ref{app:throughput}).

\textbf{Generalization to video classification.}
Applied to VideoMAE~\citep{tong2022videomae} (ViT-B) on UCF-101~\citep{soomro2012ucf101} at 91.8\% token reduction, no method collapses (all retain 79.8--81.7\% top-1), consistent with the framework: the $8\times$ larger token population keeps the amplifier below threshold. CATIS leads (81.73\% vs.\ vanilla 81.26\%) using only norm$_F$ (no CLS token), validating that P1+P3 alone suffice (details: Appendix~\ref{app:video-throughput}).

\subsection{Ablation Study}
\label{sec:ablation}

Table~\ref{tab:ablation} ablates CATIS on DeiT-3 Large ($r{=}10$, chosen to operate near the collapse threshold where component contributions are most visible). Triage is the dominant contributor (+26.2\,pp Shift avg over the best no-triage variant): unprotected merge (row~5) performs \emph{worse} than unprotected evict (row~4; 10.08 vs.\ 13.49 Shift avg), confirming that incorrect merges cause greater structural damage ($\delta_{\text{merge}} > \delta_{\text{evict}}$). Signal fusion adds +3.57--4.93pp (rows~2--3 vs.\ row~1); the two signals are complementary: CLS peaks on IN-1K (78.02\%) while norm$_F$ is stronger on IN-A (29.36 vs.\ 27.73).

\begin{table*}[t]
  \caption{Ablation study on DeiT-3 Large ($r{=}10$). Top: signal ablation. Bottom: strategy ablation. Best in \textbf{bold}.}
  \label{tab:ablation}
  \centering
  \small
  \setlength{\tabcolsep}{4pt}
  \begin{tabular}{@{}cll ccccc@{}}
    \toprule
    \# & Signal & Strategy & IN-1K & IN-A & IN-R & IN-SK & Shift avg \\
    \midrule
    \multicolumn{8}{l}{\emph{Signal ablation (same triage framework)}} \\
    1 & mom + norm$_F$ & Full CATIS & 77.30 & \textbf{35.24} & \textbf{47.17} & \textbf{36.67} & \textbf{39.69} \\
    2 & mom only & Full CATIS & \textbf{78.02} & 27.73 & 44.94 & 35.69 & 36.12 \\
    3 & norm$_F$ only & Full CATIS & 70.77 & 29.36 & 43.13 & 31.80 & 34.76 \\
    \midrule
    \multicolumn{8}{l}{\emph{Strategy ablation (same fused signal)}} \\
    1 & mom + norm$_F$ & Full CATIS (triage) & \textbf{77.30} & \textbf{35.24} & \textbf{47.17} & \textbf{36.67} & \textbf{39.69} \\
    4 & mom + norm$_F$ & Top-$k$ evict (no protect) & 27.03 & 13.04 & 16.02 & 11.40 & 13.49 \\
    5 & mom + norm$_F$ & Top-$k$ merge (no protect) & 29.28 & 3.89 & 14.59 & 11.76 & 10.08 \\
    6 & mom + norm$_F$ & CATIS + ToMe bipartite & 66.14 & 15.69 & 34.65 & 27.78 & 26.04 \\
    \bottomrule
  \end{tabular}
\end{table*}

\textbf{Verification of the two-level diagnosis ($\rho_\text{off}$).}
At the final layer of DeiT-3 Base ($r{=}23$): ToMe/MCTF reach $\rho_\text{off}{=}0.43/0.41$ (structural damage from hybrid tokens), while PiToMe and CATIS match Vanilla at $\sim$0.22. PiToMe's low $\rho_\text{off}$ yet worst accuracy confirms our two-level diagnosis: its energy score correctly identifies \emph{similar} tokens but misidentifies \emph{important} ones (semantic loss invisible to $\rho_\text{off}$), validating that signal quality (P1) and triage (P3) must both be correct. Full $\rho_\text{off}$ analysis: Figure~\ref{fig:rho-off-main}; Appendix~\ref{app:rho-off-full}.

\section{Conclusion}
\label{sec:conclusion}

This paper reframes training-free token reduction: a method's performance ceiling is not a property of its scoring formula but a structural limitation shared by all pairwise-signal methods. Two coupled components drive the collapse: a signal-agnostic error amplifier inherent to layer-wise reduction, and pairwise rankings whose stability degrades quadratically faster than unary alternatives under deep-layer homogenization. Their interaction, not either alone, produces the cliff. The framework also contributes two reusable diagnostics, $\rho_s$ and $\rho_\text{off}$, that form a pre-deployment health check: $\rho_s$ reveals \emph{whether} collapse will occur, $\rho_\text{off}$ diagnoses \emph{what type}.

The distortion recurrence and the pairwise-vs.-unary argument are task-agnostic: token merging in VLMs~\citep{bolya2022token, cao2023pumer}, KV-cache compression in LLMs, and attention pruning in speech transformers all inherit the same amplifier, so P1--P3 give a concrete design checklist there too. Adaptive layer-wise allocation, concentrating reduction in shallow layers where $\rho_s$ remains high, is a natural next step motivated directly by the recurrence.

\textbf{Limitations.}
The diagonal approximation of $\boldsymbol{\Sigma}$ ($\rho_\text{off} < 0.25$) holds for all tested ViTs but may not transfer to architectures with stronger inter-dimension coupling. Evaluation covers image classification and video classification; generalization to dense prediction tasks (detection, segmentation) and multimodal settings remains to be verified. See Appendix~\ref{app:limitations} for a detailed discussion.


\bibliography{references}

\appendix

\section{Full Diagnostic Analysis}
\label{app:diagnosis-full}

This appendix provides the complete diagnostic framework summarized in Section~\ref{sec:diagnosis}.

\subsection{The Error Amplifier: Positive Feedback in Layer-Wise Reduction}
\label{app:feedback-loop}

This appendix expands the structural justification of Proposition~\ref{prop:amplifier} (Section~\ref{sec:feedback-loop}) and provides the full derivation of the closed-form recurrence used in the main text.

All layer-wise token reduction methods face a shared structural risk: a \textbf{positive feedback loop}. At each layer, $r$ tokens are reduced (merged or evicted), altering the context available for importance estimation in subsequent layers. Suboptimal reduction operations (merges that produce distorted hybrid vectors, or evictions that discard useful information) degrade estimation quality, and errors are amplified layer by layer.

This feedback loop is \emph{signal-agnostic}: regardless of what scoring signal is used (pairwise, unary, or otherwise), the amplification mechanism exists as long as tokens are reduced layer by layer. However, the loop does not directly determine \emph{where} collapse occurs. It functions as an \textbf{error amplifier} whose trigger threshold depends on per-layer signal quality. If the signal is perfectly reliable at every layer (zero distortion), the feedback loop never triggers and the Pareto curve declines smoothly. If the signal becomes unreliable at some depth, per-layer distortions are amplified, producing the observed cliff in the Pareto curve. The faster signal quality degrades across layers, the lower the $r$ at which the amplifier triggers, and the earlier collapse arrives.

\paragraph{Formal recurrence model.}
Consider an $L$-layer ViT. Each layer reduces $r$ tokens (via merge or evict), so the number of remaining tokens at layer $l$ is $N^{(l)} = N_p - l \cdot r$. We define two quantities:
\begin{itemize}[leftmargin=*, nosep]
  \item $\epsilon^{(l)} \in [0,1]$: the \emph{per-operation distortion rate} at layer $l$, a continuous measure of expected representation degradation per reduction operation (whether merge or evict). Higher $\epsilon$ means each operation introduces more distortion on average.
  \item $\Delta^{(l)} \geq 0$: the \emph{cumulative representation distortion} entering layer $l$, aggregating the degradation from all reduction operations in preceding layers.
\end{itemize}

Under Assumption~\ref{ass:feedback} (monotone feedback coupling, restated from Section~\ref{sec:feedback-loop}), the per-operation distortion rate $\epsilon^{(l)}$ is a monotonically non-decreasing function of the cumulative distortion $\Delta^{(l)}$.

\noindent\emph{Intuition for Assumption~\ref{ass:feedback}}: prior distortion, whether from hybrid vectors (merge) or information loss (evict), degrades the attention weights and feature statistics that subsequent layers use for importance estimation, increasing the distortion rate. This holds regardless of operation type; we make no assumption about the specific functional form beyond monotonicity.

Under this assumption, each reduction operation contributes expected damage intensity $\delta > 0$ to subsequent layers, giving the recurrence:
\begin{equation}
  \Delta^{(l+1)} = \Delta^{(l)} + \epsilon^{(l)} \cdot r \cdot \delta,\quad \Delta^{(0)}\!=\!0.
  \label{eq:recurrence-app}
\end{equation}
Since $\epsilon^{(l)}$ is non-decreasing in $\Delta^{(l)}$, the sequence $\{\Delta^{(l)}\}$ is super-linear in $l$: as distortion accumulates, each subsequent layer's reduction operations introduce increasingly larger distortion, which in turn degrades estimation further. This is the mathematical statement of the error amplifier.

\paragraph{Closed-form analysis under linear coupling.}
To obtain concrete predictions, we further assume a linear coupling: $\epsilon^{(l)} = \epsilon_0 + \alpha \Delta^{(l)}$, where $\epsilon_0 > 0$ is the baseline distortion rate in the absence of accumulated distortion and $\alpha > 0$ is the coupling strength (how strongly accumulated distortion feeds back into per-operation distortion, controlling the exponential base $\beta$; larger $\alpha$ means faster collapse). The coupling is operation-agnostic: whether a reduction operation is a merge or an eviction, greater cumulative distortion degrades the attention weights and feature statistics that subsequent layers use for importance estimation, increasing $\epsilon$. Substituting into~\eqref{eq:recurrence-app}:
\begin{equation}
  \Delta^{(l+1)} = (1 + \alpha r \delta)\,\Delta^{(l)} + \epsilon_0 r \delta.
  \label{eq:linear-rec}
\end{equation}
Setting $\beta = 1 + \alpha r \delta > 1$ and $c = \epsilon_0 r \delta$, the closed-form solution is:
\begin{equation}
  \Delta^{(l)} = \frac{c}{\beta - 1}\bigl(\beta^l - 1\bigr) = \frac{\epsilon_0}{\alpha}\bigl(\beta^l - 1\bigr).
  \label{eq:closed-form}
\end{equation}

\paragraph{Derivation of $r_{\text{crit}} \propto 1/L$.}
Suppose collapse occurs when $\Delta^{(L)}$ reaches a threshold $T$. From~\eqref{eq:closed-form}:
\begin{equation}
  \beta^L = 1 + \frac{\alpha T}{\epsilon_0}.
  \label{eq:collapse-threshold}
\end{equation}
Taking logarithms and applying the first-order approximation $\ln(1 + x) \approx x$ for $\alpha r \delta \ll 1$ (satisfied when per-layer feedback is mild relative to the damage scale):
\begin{gather}
  L \cdot \alpha\, r_{\text{crit}}\, \delta \;\approx\; \ln\!\Bigl(1 + \frac{\alpha T}{\epsilon_0}\Bigr), \notag \\
  \Longrightarrow\quad
  r_{\text{crit}} \;\approx\; \frac{\ln\bigl(1 + \alpha T / \epsilon_0\bigr)}{\alpha\, \delta\, L} \;\propto\; \frac{1}{L}.
  \label{eq:rcrit}
\end{gather}
Since $\alpha$, $T$, $\epsilon_0$, and $\delta$ are model- and signal-dependent constants independent of depth, $r_{\text{crit}}$ is inversely proportional to $L$: deeper models collapse at lower per-layer reduction.

This yields three qualitative predictions that do not depend on the linear-coupling assumption but follow from any positive-feedback recurrence satisfying Assumption~1:

\begin{table*}[t]
  \centering
  \small
  \begin{tabular}{p{5.5cm} p{3.6cm} p{5.0cm}}
    \toprule
    \textbf{Prediction} & \textbf{Mechanism} & \textbf{Empirical support} \\
    \midrule
    Distortion grows exponentially in $l$ $\Rightarrow$ Pareto curves are convex & $\beta^l$ term in \eqref{eq:closed-form} & All methods show convex curves (Fig.~\ref{fig:pareto-teaser}; Appendix~\ref{app:pareto-full}) \\[4pt]
    Larger $r$ accelerates growth $\Rightarrow$ cliff appears at high compression & $\beta$ linear in $r$ & ViT-L accuracy drops 23--35\,pp when $r{:}9\!\to\!11$ \\[4pt]
    Deeper models ($L$ larger) collapse at lower $r$ ($r_{\text{crit}} \propto 1/L$) & $\beta^L$ exponentially sensitive to $L$ & ViT-L ($L{=}24$) collapses at $r{\approx}9$; ViT-S ($L{=}12$) stable at $r{=}13$ \\
    \bottomrule
  \end{tabular}
  \caption*{Predictions from the distortion recurrence (Appendix~\ref{app:feedback-loop}).}
\end{table*}

\paragraph{Merge vs.\ evict: a dual-channel distortion model.}
The base recurrence~\eqref{eq:recurrence-app} uses a single damage intensity $\delta$, but the two reduction operations differ structurally:
\begin{itemize}[leftmargin=*, nosep]
  \item \textbf{Merge}: produces a hybrid vector that persists in subsequent layers and participates in attention computations, propagating structural damage ($\rho_{\text{off}}$ rises; Section~\ref{sec:diagnosis}). Damage intensity $\delta_m$.
  \item \textbf{Evict}: causes information loss but does not introduce hybrid vectors, so it does not pollute surviving representations. Damage intensity $\delta_e$.
\end{itemize}
Therefore $\delta_m > \delta_e$. This is consistent with Table~\ref{tab:ablation}: under identical signals, top-$k$ merge without protection achieves Shift avg = 10.08\% vs.\ 13.49\% for top-$k$ evict without protection, a 3.4\,pp gap. While the gap may partly reflect different error profiles (merge exposes important tokens to hybridization, evict to deletion), the direction is consistent with the higher per-operation damage of merging.

This distinction is integral to the recurrence model, not a post-hoc addendum: any method that supports both merge and evict operations naturally decomposes the per-layer distortion increment into two channels, each with its own rate $\epsilon$ and intensity $\delta$.

\paragraph{How CATIS modifies the recurrence.}
CATIS intervenes at all three terms of the recurrence:

\noindent\textbf{Mechanism 1 (lower $\epsilon_0$).} Replacing pairwise signals ($\rho_s \approx 0.27$ in deep layers) with fused unary signals ($\rho_s \approx 0.39$--$0.48$) reduces $\epsilon^{(l)}$ at every layer. In the linear model, this reduces the seed $c = \epsilon_0 r \delta$ and thus the entire distortion trajectory.

\noindent\textbf{Mechanism 2 (lower effective $r$).} The triage mechanism designates a fraction $f_p$ of tokens as \emph{protected}. These tokens are exempt from any reduction operation, confining distortion to the non-protected subset.

\noindent\textbf{Mechanism 3 (channel routing: substitute $\delta_e$ for $\delta_m$).} Among the non-protected tokens, CATIS assigns the lowest-confidence tokens (those with the highest expected $\epsilon$) to eviction rather than merging, substituting $\delta_e$ for $\delta_m$ on the highest-risk decisions. Combining all three mechanisms, the CATIS recurrence becomes:
\begin{equation}
  \Delta_{\text{CATIS}}^{(l+1)} = \Delta_{\text{CATIS}}^{(l)} + \epsilon_m^{(l)} \cdot r_m \cdot \delta_m + \epsilon_e^{(l)} \cdot r_e \cdot \delta_e,
  \label{eq:catis-rec}
\end{equation}
where $r_m + r_e = r(1-f_p)$ after protection, with $\epsilon_e^{(l)} > \epsilon_m^{(l)}$ (eviction candidates are the least certain) but $\delta_e < \delta_m$ (eviction is less damaging). The net effect is threefold: protection reduces the effective $r$, unary signals lower $\epsilon_0$, and channel routing minimizes the total per-layer distortion increment by pairing the highest-$\epsilon$ decisions with the lowest-$\delta$ operation.

\paragraph{Limitations of the model.}
\begin{enumerate}[leftmargin=*, nosep]
  \item The monotone feedback coupling (Assumption~1) is a modeling assumption motivated by the observation that context reduction degrades estimation quality, supported empirically by the layer-wise $\rho_s$ curves (Figure~\ref{fig:cosine-degradation}).
  \item The linear coupling used to derive~\eqref{eq:closed-form} is an approximation for tractability. The qualitative predictions (super-linear growth, $r_{\text{crit}} \propto 1/L$) are robust to the specific functional form: any monotone positive-feedback recurrence produces super-linear distortion growth.
  \item The base recurrence uses a single $\delta$; the dual-channel extension (Eq.~\ref{eq:catis-rec}) partially addresses this by distinguishing $\delta_m$ and $\delta_e$. Within each channel, we still assume uniform damage per operation; in practice, reducing a high-importance token causes larger distortion than reducing a low-importance one. This increases the variance of $\Delta^{(l)}$ without changing the expected growth trend.
  \item We assume constant $r$ per layer, consistent with all evaluated methods. Adaptive per-layer budgets are a natural extension; the formal framework accommodates them by replacing $r$ with $r^{(l)}$ in~\eqref{eq:recurrence-app}.
  \item The $r_{\text{crit}} \propto 1/L$ derivation (Eq.~\ref{eq:rcrit}) uses a first-order approximation $\ln(1+x) \approx x$ requiring $\alpha r \delta \ll 1$. While the $1/L$ trend is empirically validated ($R^2 > 0.9$), the approximation may lose accuracy at extreme operating points where per-layer feedback is strong.
\end{enumerate}

\subsection{Trigger Threshold: Pairwise Signals Degrade Layer by Layer}
\label{app:pairwise-degradation}

We now show that pairwise signal quality degrades severely in deeper layers, explaining why the error amplifier triggers at low $r$ for all existing methods. We take K-vector cosine similarity, the signal shared by all four methods, as representative. For each image, we compute the $N_p \times N_p$ pairwise similarity matrix $\mathbf{S}$ from both its clean version and a corrupted version (e.g., Gaussian noise, severity 5). We quantify signal degradation using two complementary metrics. \emph{Spearman $\rho_s$} measures ranking consistency: given the flattened clean and corrupted similarity vectors, each entry is replaced by its rank, and $\rho_s$ is the Pearson correlation between the two rank vectors:
\begin{equation}
  \rho_s = \frac{\mathrm{Cov}(R_{\text{clean}},\, R_{\text{corr}})}{\mathrm{Std}(R_{\text{clean}}) \cdot \mathrm{Std}(R_{\text{corr}})}
  \label{eq:spearman}
\end{equation}
where $R_{\text{clean}}$ and $R_{\text{corr}}$ are the rank-transformed similarity vectors. $\rho_s{=}1$ means identical rankings (merge decisions unaffected); $\rho_s{\to}0$ means rankings are randomized. \emph{Frobenius distance} $\Delta F = \| \mathbf{S}_{\text{clean}} - \mathbf{S}_{\text{corr}} \|_F$ captures magnitude distortion, i.e., how much absolute similarity values change, independent of rank order.

Figure~\ref{fig:cosine-degradation} in the main text shows results on DeiT-3 Base ($n{=}1000$ images). The degradation is comprehensive: $\rho_s$ drops from 0.88 in shallow layers to 0.27 in deep layers (ranking nearly randomized), while $\Delta F$ increases from 21--30 to 55--78 (matrix values severely distorted). This directly explains the collapse pattern: each deep layer introduces large estimation errors that the positive feedback loop amplifies forward. Since all four methods share this signal type, they exhibit the same accelerating decline. On shifted inputs, the absolute signal quality is lower at all layers, which reduces the accuracy floor after collapse, though the collapse point itself is determined primarily by model depth and per-layer reduction ratio. We use algorithmic corruptions as a controlled probe; the degradation itself is architectural (driven by representation homogenization in deep layers~\citep{raghu2021vision}) and thus applies regardless of the specific form of distribution shift, as confirmed by identical collapse points on IN-A, IN-R, and IN-Sketch (Figure~\ref{fig:pareto-teaser}b).

\subsection{Structural Consequences: $\rho_\text{off}$ Analysis}
\label{app:rho-off-analysis}

The pairwise signal degradation diagnosed above has a measurable downstream consequence on feature-space structure. When incorrect merge decisions pair semantically dissimilar tokens, the resulting ``hybrid'' vectors shift multiple feature dimensions simultaneously, introducing spurious inter-dimension correlations. We quantify this with $\rho_\text{off}$, defined as the mean absolute off-diagonal entry of the Pearson correlation matrix $\mathbf{R} \in \mathbb{R}^{d \times d}$ computed across the $N_p$ patch token features at a given layer:
\begin{equation}
  \rho_\text{off} = \frac{1}{d(d-1)} \sum_{j \neq k} |r_{jk}|
  \label{eq:rho-off}
\end{equation}
where $r_{jk}$ is the Pearson correlation between feature dimensions $j$ and $k$ across the $N_p$ tokens. $\rho_\text{off} \approx 0$ indicates approximately independent dimensions (healthy); high $\rho_\text{off}$ signals dimension collapse. On vanilla models (no reduction), $\rho_\text{off}$ stays below 0.25 at all layers, confirming approximately independent dimensions. Under token reduction, we measure $\rho_\text{off}$ layer by layer for all methods at multiple $r$ values across 6~models, using two panels per model (IN-1K vs.\ shifted inputs; the shifted panel pools multiple robustness benchmarks).

\begin{figure*}[t]
  \centering
  \begin{subfigure}[t]{0.48\textwidth}
    \centering
    \includegraphics[width=\linewidth]{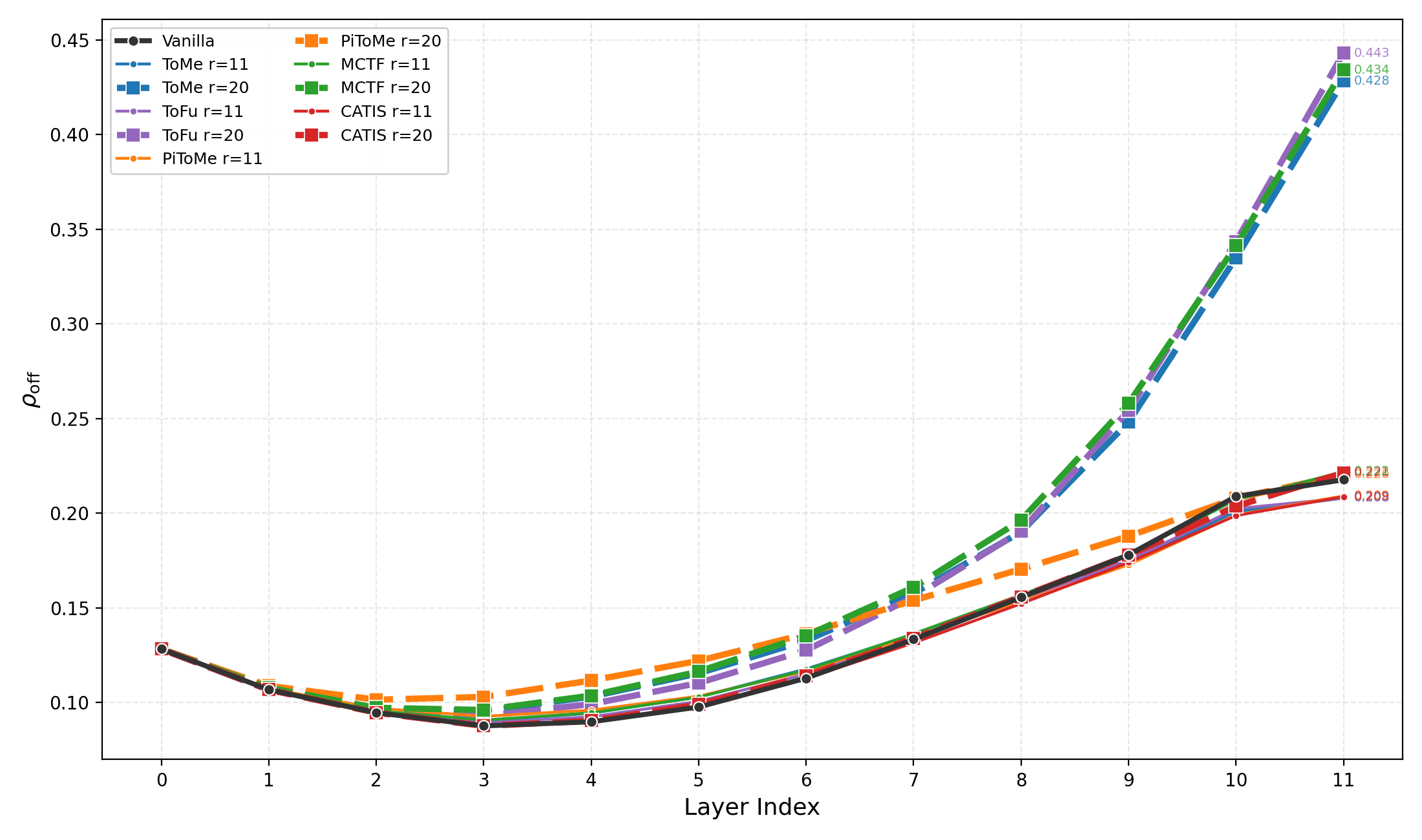}
    \caption{DeiT-3 Base --- ImageNet-1K}
  \end{subfigure}
  \hfill
  \begin{subfigure}[t]{0.48\textwidth}
    \centering
    \includegraphics[width=\linewidth]{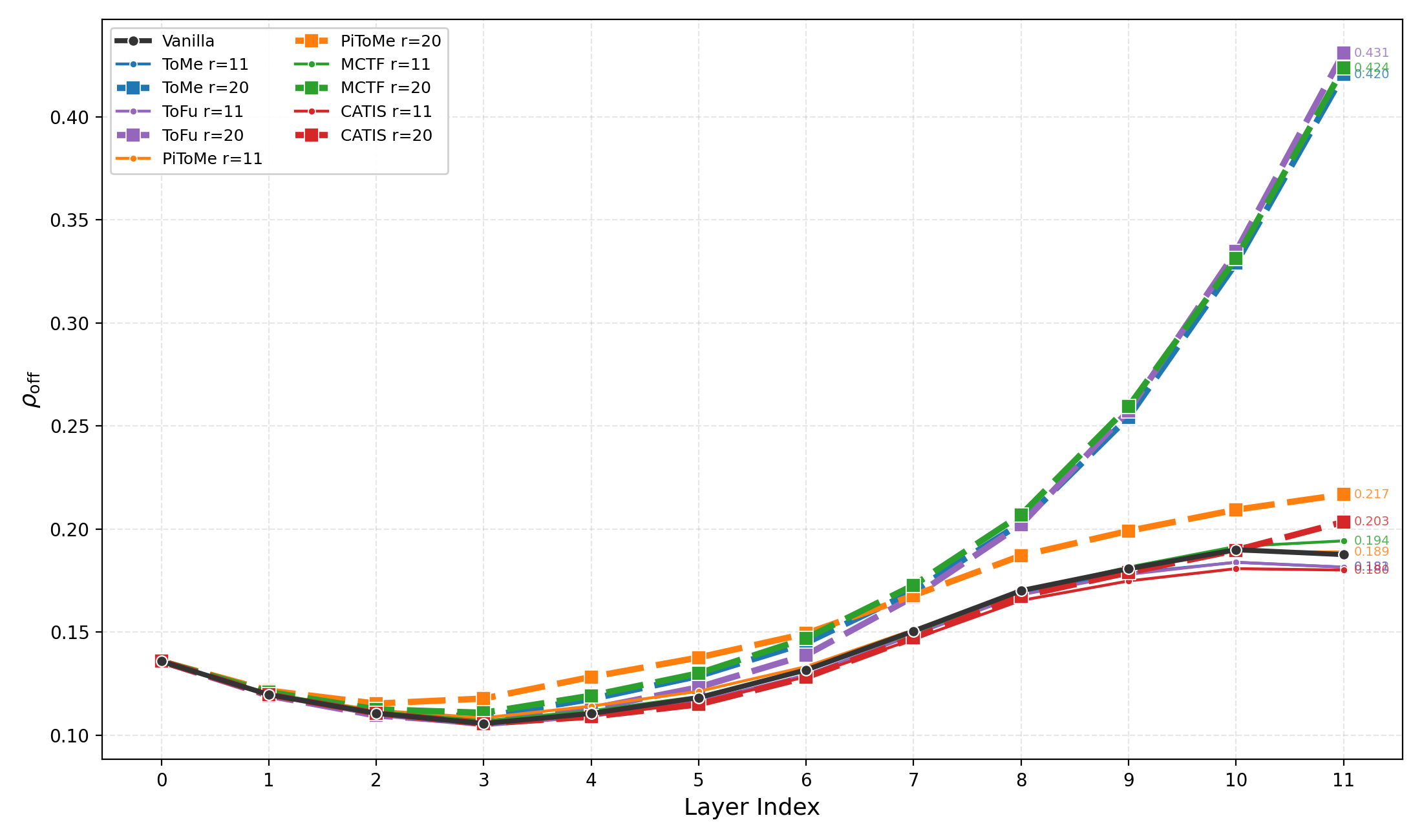}
    \caption{DeiT-3 Base --- shifted}
  \end{subfigure}
  \caption{$\rho_\text{off}$ under token reduction at varying $r$. ToMe/MCTF exhibit surging $\rho_\text{off}$ at high $r$ (structural damage from incorrect merges); PiToMe stays low but achieves worst accuracy (correct merge pairing, wrong triage decisions, i.e., semantic loss invisible to $\rho_\text{off}$); CATIS tracks Vanilla throughout.}
  \label{fig:rho-off-main}
\end{figure*}

\begin{figure*}[t]
  \centering
  \begin{subfigure}[t]{0.48\textwidth}
    \centering
    \includegraphics[width=\linewidth]{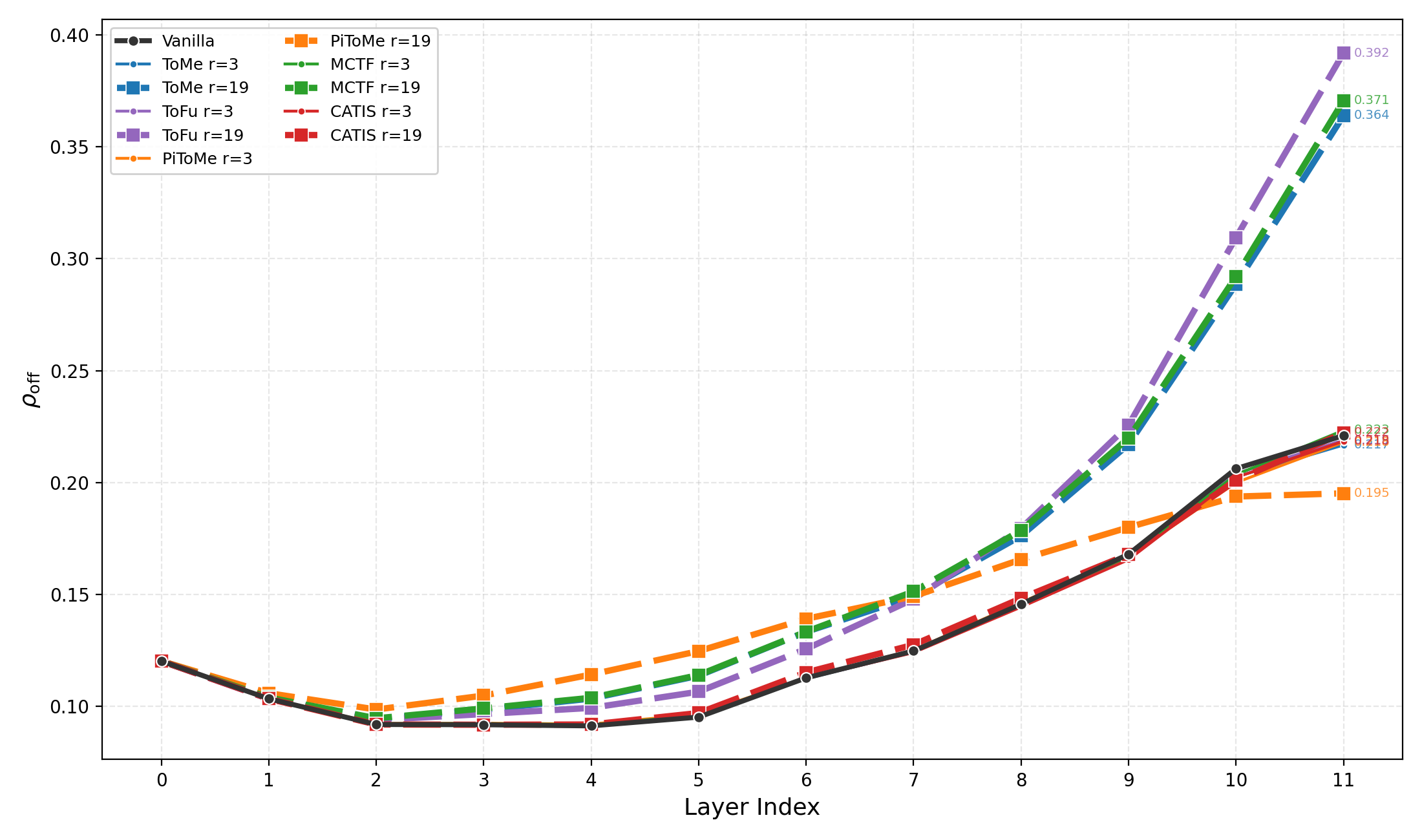}
    \caption{DeiT-3 Small --- IN-1K}
  \end{subfigure}
  \hfill
  \begin{subfigure}[t]{0.48\textwidth}
    \centering
    \includegraphics[width=\linewidth]{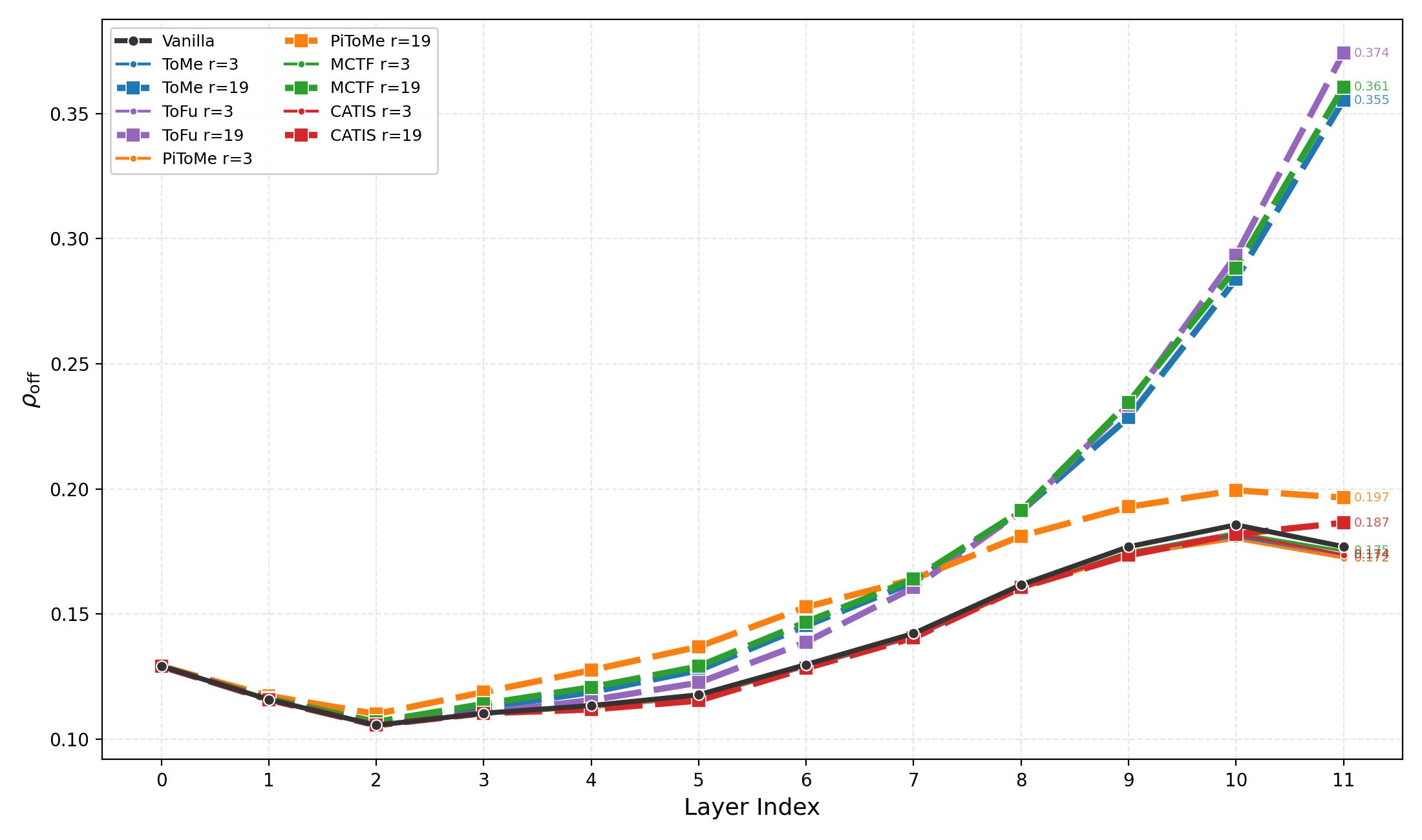}
    \caption{DeiT-3 Small --- shifted}
  \end{subfigure}
  \caption{$\rho_\text{off}$ under token reduction: DeiT-3 Small. DeiT-3 Base appears in the main text (Figure~\ref{fig:rho-off-main}).}
  \label{fig:rho-off-all-s}
\end{figure*}

\begin{figure*}[t]
  \centering
  \begin{subfigure}[t]{0.48\textwidth}
    \centering
    \includegraphics[width=\linewidth]{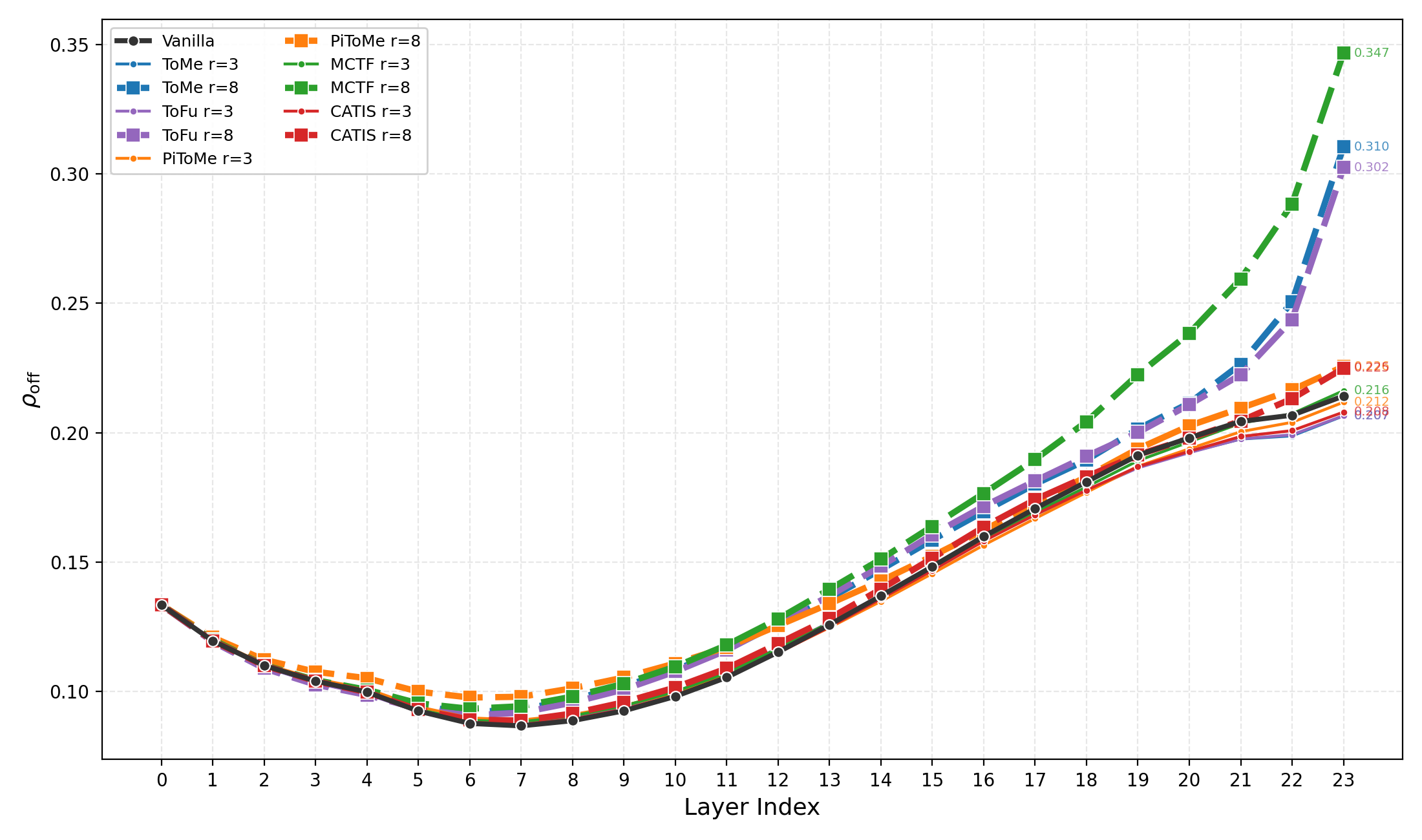}
    \caption{DeiT-3 Large --- IN-1K}
  \end{subfigure}
  \hfill
  \begin{subfigure}[t]{0.48\textwidth}
    \centering
    \includegraphics[width=\linewidth]{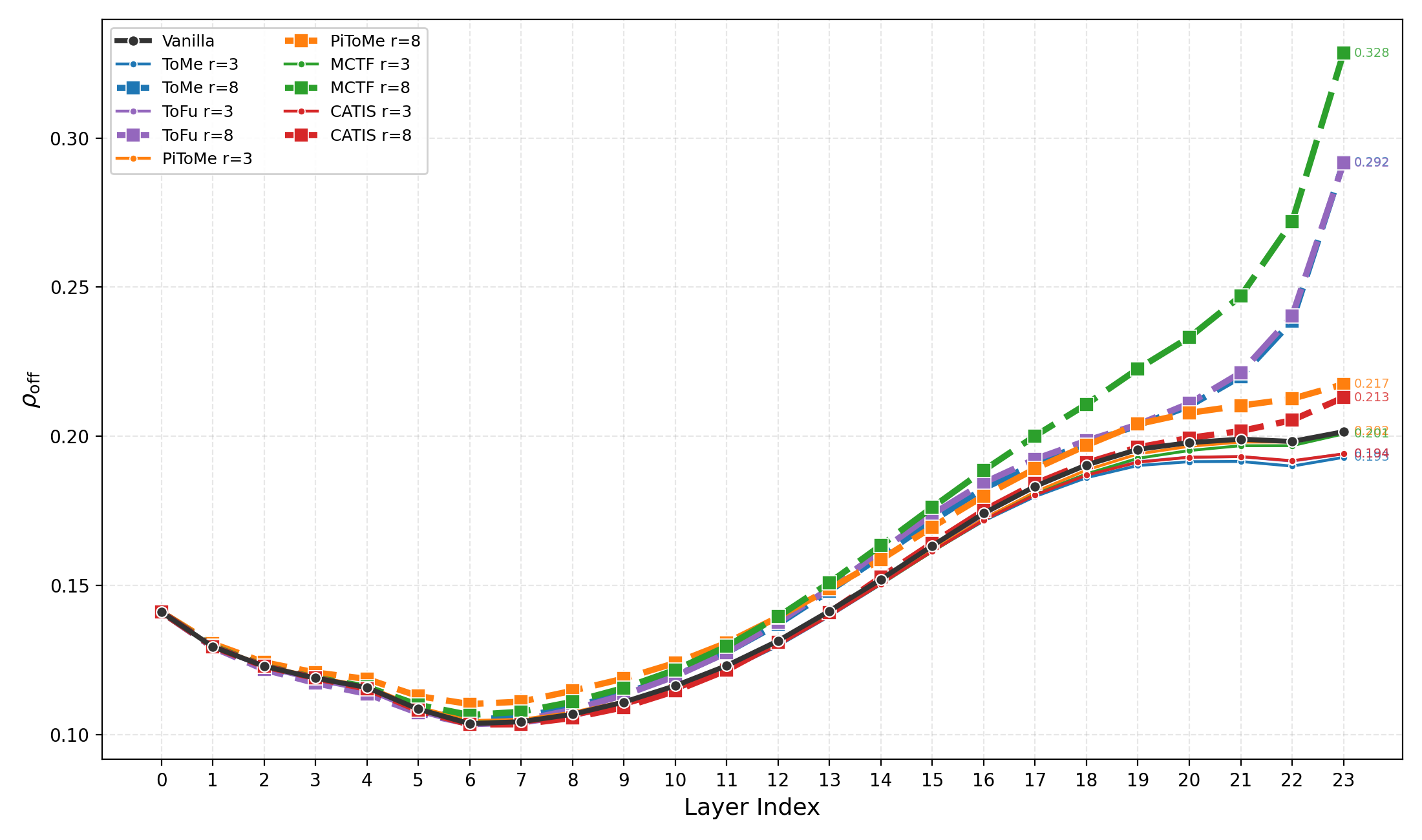}
    \caption{DeiT-3 Large --- shifted}
  \end{subfigure}
  \caption{$\rho_\text{off}$ under token reduction: DeiT-3 Large.}
  \label{fig:rho-off-all-dl}
\end{figure*}

\begin{figure*}[t]
  \centering
  \begin{subfigure}[t]{0.48\textwidth}
    \centering
    \includegraphics[width=\linewidth]{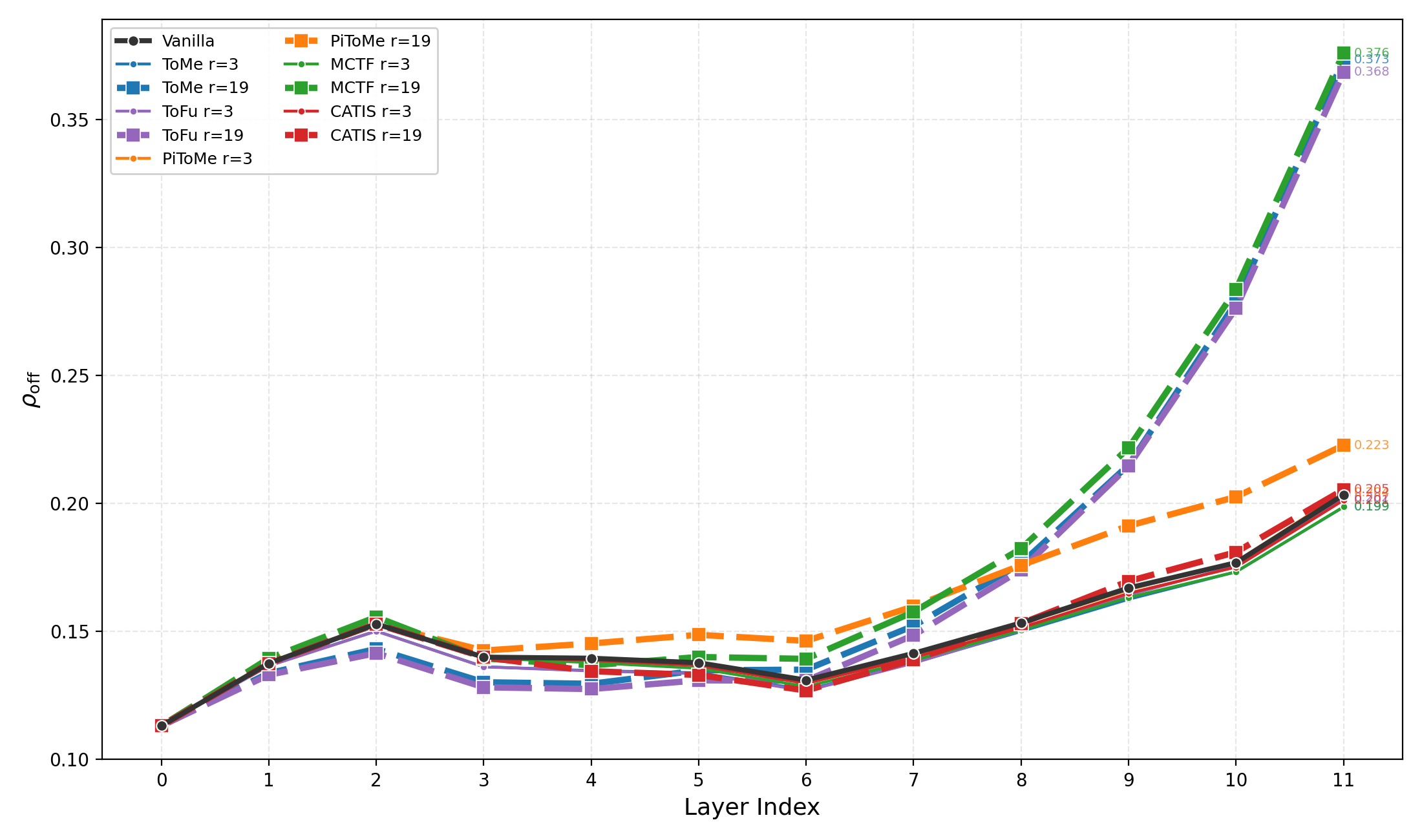}
    \caption{ViT-Small --- IN-1K}
  \end{subfigure}
  \hfill
  \begin{subfigure}[t]{0.48\textwidth}
    \centering
    \includegraphics[width=\linewidth]{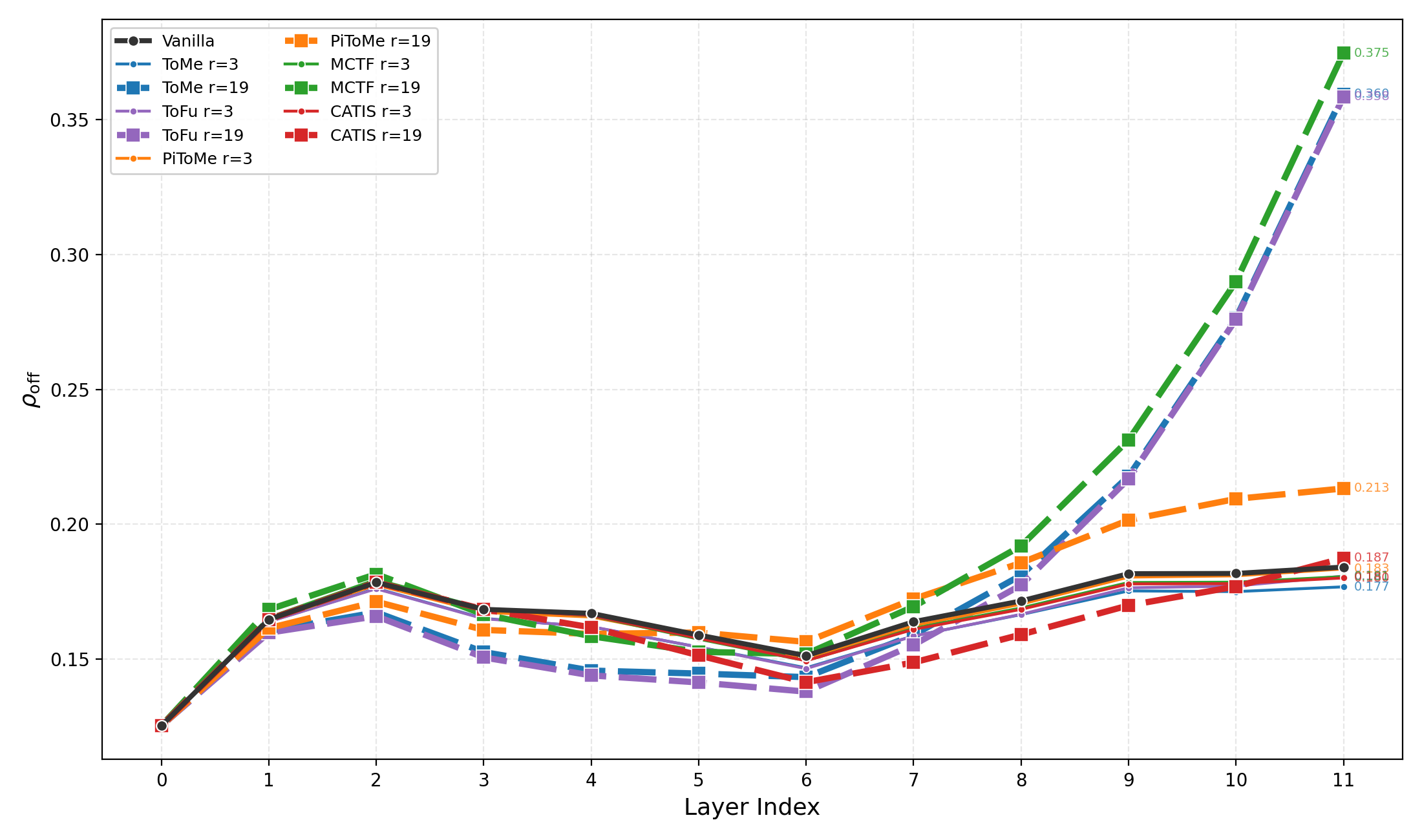}
    \caption{ViT-Small --- shifted}
  \end{subfigure}
  \caption{$\rho_\text{off}$ under token reduction: ViT-Small.}
  \label{fig:rho-off-all-vs}
\end{figure*}

\begin{figure*}[t]
  \centering
  \begin{subfigure}[t]{0.48\textwidth}
    \centering
    \includegraphics[width=\linewidth]{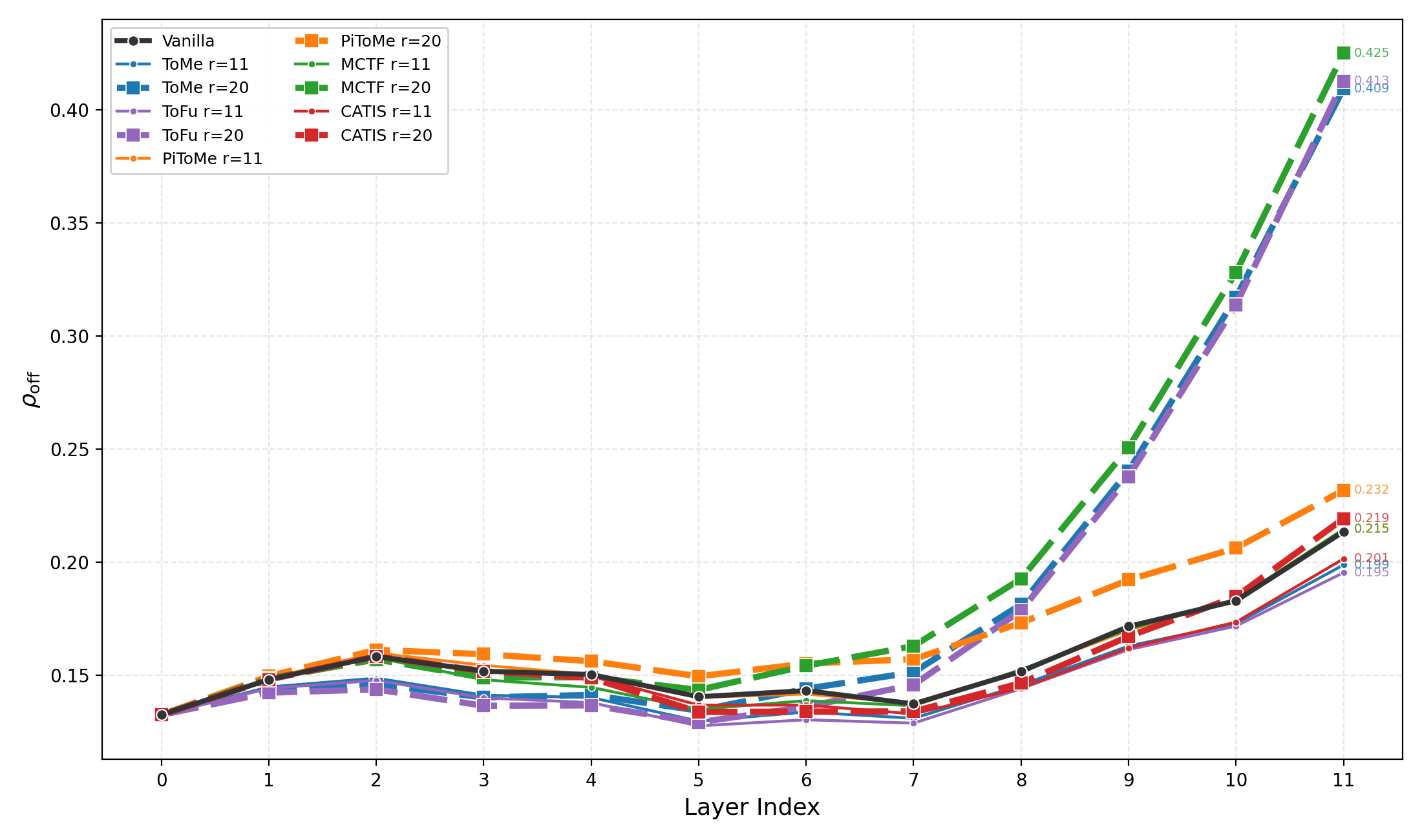}
    \caption{ViT-Base --- IN-1K}
  \end{subfigure}
  \hfill
  \begin{subfigure}[t]{0.48\textwidth}
    \centering
    \includegraphics[width=\linewidth]{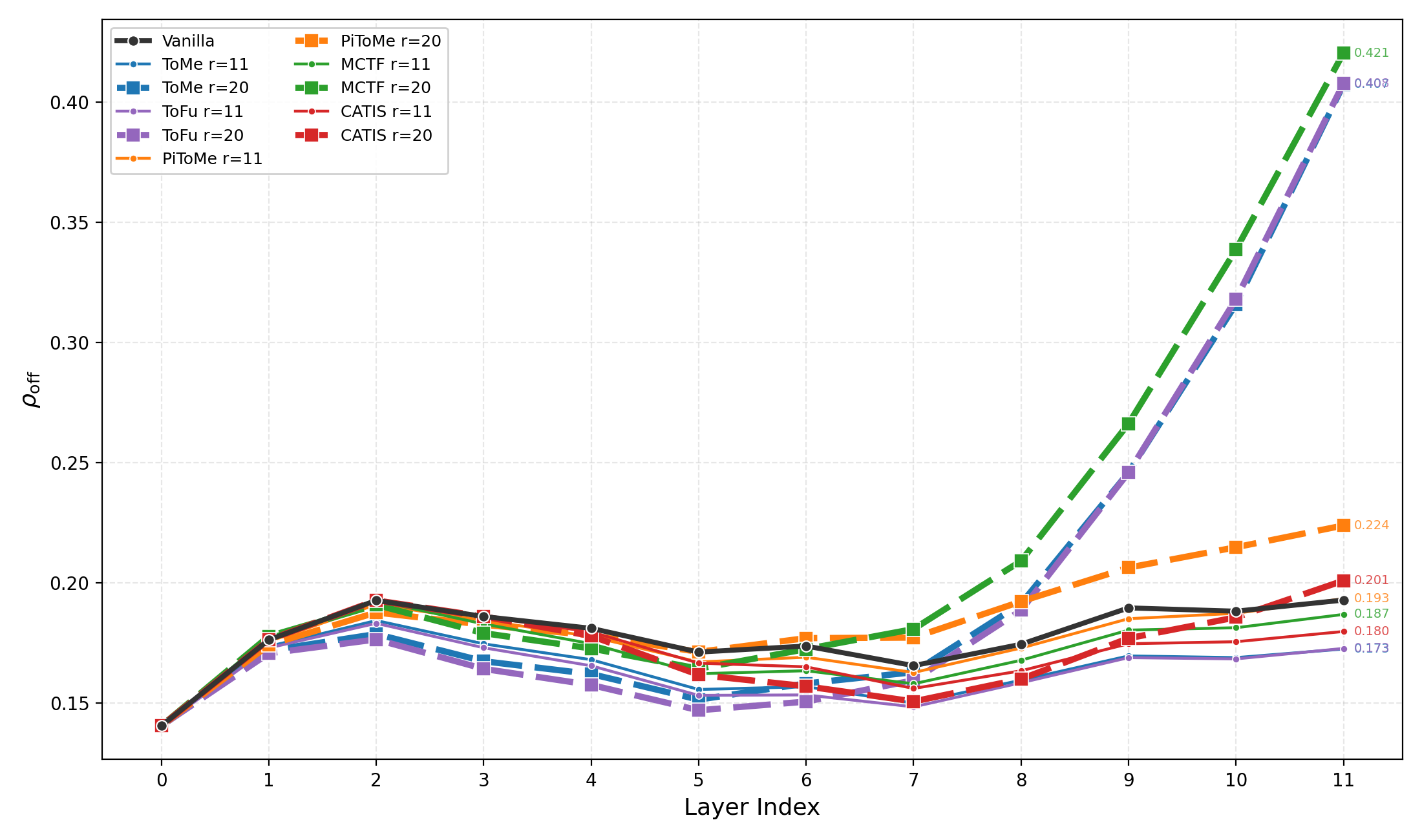}
    \caption{ViT-Base --- shifted}
  \end{subfigure}
  \caption{$\rho_\text{off}$ under token reduction: ViT-Base.}
  \label{fig:rho-off-all-vb}
\end{figure*}

\begin{figure*}[t]
  \centering
  \begin{subfigure}[t]{0.48\textwidth}
    \centering
    \includegraphics[width=\linewidth]{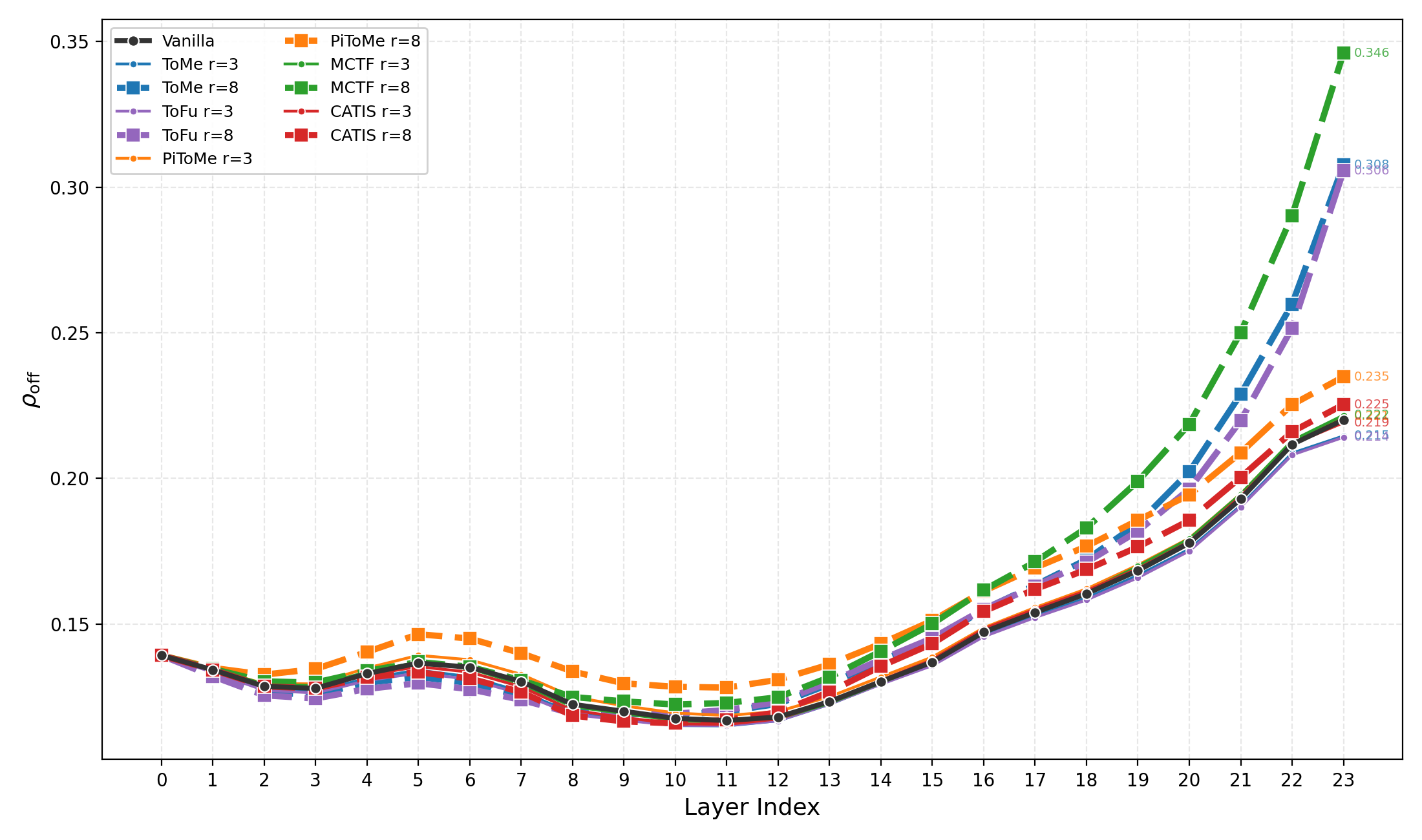}
    \caption{ViT-Large --- IN-1K}
  \end{subfigure}
  \hfill
  \begin{subfigure}[t]{0.48\textwidth}
    \centering
    \includegraphics[width=\linewidth]{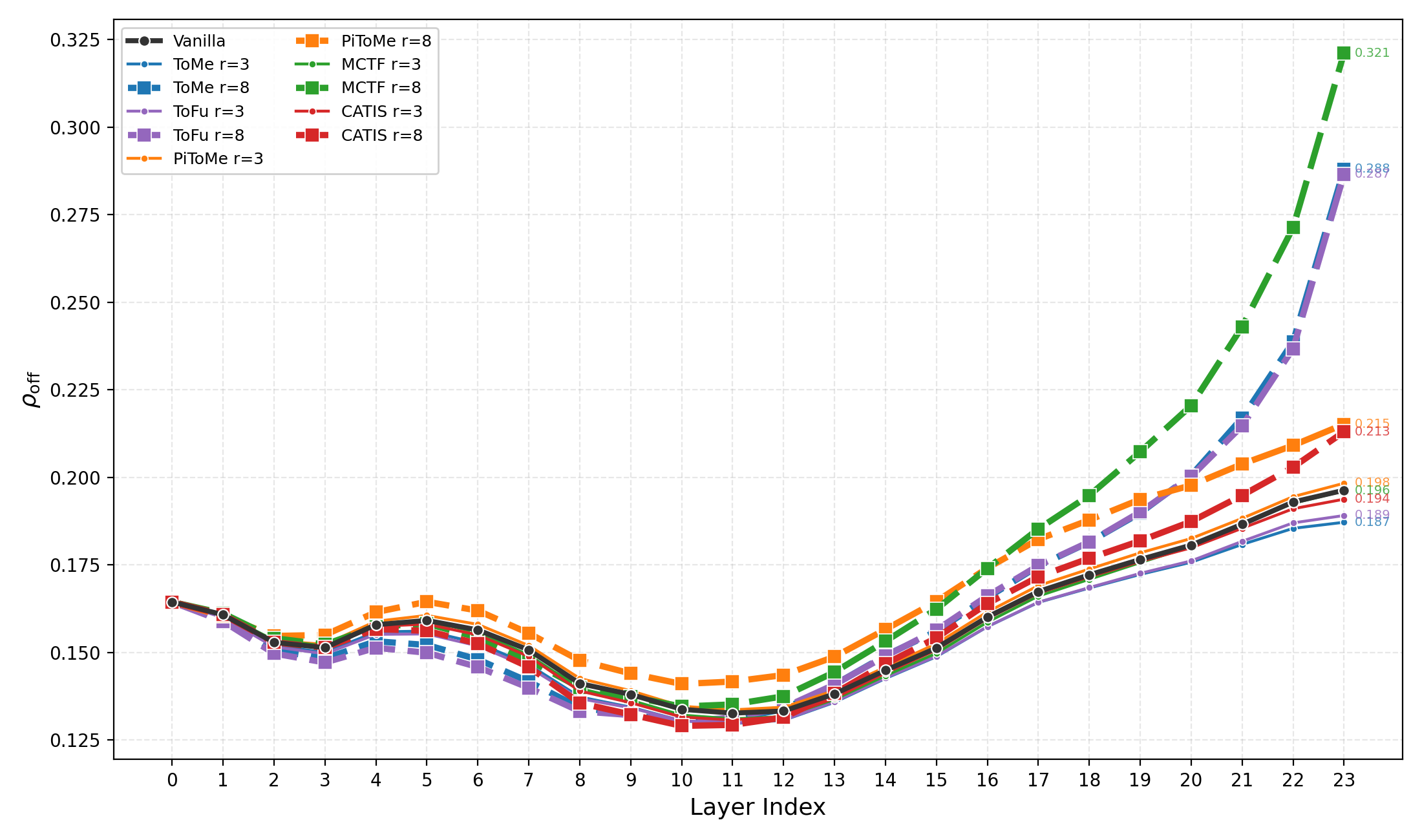}
    \caption{ViT-Large --- shifted}
  \end{subfigure}
  \caption{$\rho_\text{off}$ under token reduction: ViT-Large. Across Figures~\ref{fig:rho-off-all-s}--\ref{fig:rho-off-all-vl}, the pattern is universal: ToMe/MCTF's $\rho_\text{off}$ surges at high $r$ (structural damage from hybrid tokens); PiToMe stays low but achieves worst accuracy (semantic loss invisible to $\rho_\text{off}$); CATIS tracks Vanilla throughout.}
  \label{fig:rho-off-all-vl}
\end{figure*}

The results (Figure~\ref{fig:rho-off-main} and Appendix Figures~\ref{fig:rho-off-all-s}--\ref{fig:rho-off-all-vl}) are consistent across all 12~plots ($6$ models $\times$ $2$ conditions per model). ToMe and MCTF exhibit sharply rising $\rho_\text{off}$ at high $r$, reaching 0.35--0.43 in the final layer. This is a direct structural signature of error-amplifier--driven collapse: degraded pairwise signals cause incorrect merges, which produce hybrid tokens that destroy dimension independence, which in turn maximizes the amplifier's gain. This confirms that $\rho_\text{off}$ is a reliable diagnostic of the collapse mechanism.

PiToMe provides an instructive contrast. Its $\rho_\text{off}$ remains low (0.20--0.23), close to Vanilla, yet it achieves the \emph{worst} accuracy of all methods. The explanation lies in the distinction between two types of damage that incorrect importance signals can cause. PiToMe's energy score still correctly identifies \emph{similarity} between tokens: merge pairings remain between genuinely similar tokens, so the merge operation itself does not produce severe hybrid vectors and $\rho_\text{off}$ stays low. The failure occurs at the \emph{triage} level: the energy score, built on pairwise cosine similarity, misidentifies which tokens deserve protection under distribution shift. Foreground tokens carrying critical semantic information are exposed to merging, while less informative tokens are preserved. The result is semantic information loss, a form of damage that $\rho_\text{off}$ cannot detect, because the surviving feature space retains its statistical structure even as its semantic content is degraded. This comparison establishes that protecting feature-space integrity (P3) is necessary but not sufficient: without reliable importance signals (P1), the triage mechanism protects the wrong tokens.

CATIS maintains $\rho_\text{off} \approx$ Vanilla ($\Delta < 0.02$) at all $r$ values across all models, confirming that the triage mechanism prevents structural damage while unary signals ensure correct triage decisions.


\subsection{Perturbation Energy Gap: Pairwise vs.\ Unary Signals}
\label{app:perturbation-energy}

We formalize the signal-class argument of Section~\ref{sec:paradigm}. Rather than counting ranking inversions (which requires modeling the full joint distribution of perturbed scores), we analyze the \emph{total perturbation energy}, the expected sum of squared score changes, as a proxy for ranking disruption: for score vectors with comparable margin distributions, larger total perturbation energy implies more ranking inversions and thus lower $\rho_s$. Under deep-layer homogenization, pairwise margins shrink (as cosine similarities converge), further amplifying the ranking disruption for a given perturbation energy.

\paragraph{Setup and notation.}
Consider $N_p$ token representations $\{\mathbf{z}_i\}_{i=1}^{N_p} \subset \mathbb{R}^d$ at a given layer. Each token is subject to an independent perturbation: $\tilde{\mathbf{z}}_i = \mathbf{z}_i + \boldsymbol{\xi}_i$, where $\boldsymbol{\xi}_i \in \mathbb{R}^d$ models the cumulative distortion from preceding layers. We define the \emph{total perturbation energy} as $V = \sum_{\text{scores}} E[(\tilde{s} - s)^2]$, where the sum runs over all scores (all $\binom{N_p}{2}$ pairs for pairwise, all $N_p$ tokens for unary). The intuition is straightforward: the larger $V$, the more the importance ranking is disrupted, and the lower $\rho_s$.

\medskip

We restate Proposition~\ref{prop:perturbation} (Section~\ref{sec:paradigm}) here in its precise technical form, made fully explicit. Let $\{\mathbf{z}_i\}_{i=1}^{N_p} \subset \mathbb{R}^d$ be token representations subject to i.i.d.\ perturbations $\boldsymbol{\xi}_i$ satisfying (a)~$E[\boldsymbol{\xi}_i] = \mathbf{0}$ (zero-mean), (b)~$\mathrm{Cov}(\boldsymbol{\xi}_i) = \sigma_\xi^2 \mathbf{I}_d$ (isotropic: uncorrelated components with equal variance), and (c)~$\boldsymbol{\xi}_i \perp \boldsymbol{\xi}_j$ for $i \neq j$ (cross-token independence). Let $V = \sum_{\text{scores}} E[(\tilde{s} - s)^2]$ denote the total perturbation energy. Then for a pairwise signal $s(i,j) = f(\mathbf{z}_i, \mathbf{z}_j)$ we have $V_{\text{pair}} = \Theta(N_p^2 \cdot G_{\text{pair}} \cdot \sigma_\xi^2)$, while for a unary signal $s(i) = g(\mathbf{z}_i, \boldsymbol{\theta})$ with $\boldsymbol{\theta} = \boldsymbol{\theta}(\mathbf{z}_1, \ldots, \mathbf{z}_{N_p})$ we have $V_{\text{unary}} = \Theta(N_p \cdot G_{\text{unary}} \cdot \sigma_\xi^2)$, where $G_{\text{pair}} = E[\|\nabla_{\mathbf{z}} f\|^2]$ and $G_{\text{unary}} = E[\|\nabla_{\mathbf{z}}^{\text{direct}} g\|^2]$ are mean squared gradient norms (averaged over token indices under gradient homogeneity). Under the additional assumption $G_{\text{pair}} / G_{\text{unary}} = \Theta(1)$, the gap satisfies $V_{\text{pair}} / V_{\text{unary}} = \Theta(N_p)$.

\medskip

\noindent\textbf{Derivation (a): Pairwise signal.}
For each pair $(i,j)$ with $i \neq j$, a first-order Taylor expansion around the unperturbed point gives
\begin{multline}
  \tilde{s}(i,j) - s(i,j) = f(\mathbf{z}_i + \boldsymbol{\xi}_i,\; \mathbf{z}_j + \boldsymbol{\xi}_j) \\
  - f(\mathbf{z}_i, \mathbf{z}_j) \approx \nabla_{\mathbf{z}_i}\! f \cdot \boldsymbol{\xi}_i + \nabla_{\mathbf{z}_j}\! f \cdot \boldsymbol{\xi}_j,
\end{multline}
where the gradients $\nabla_{\mathbf{z}_i} f$, $\nabla_{\mathbf{z}_j} f \in \mathbb{R}^d$ are evaluated at the unperturbed point and are therefore fixed (non-random) vectors. Squaring and taking expectations yields three terms:
\begin{align}
  &E[(\tilde{s}(i,j) - s(i,j))^2] \notag \\
  &= E[(\nabla_{\mathbf{z}_i}\! f \cdot \boldsymbol{\xi}_i)^2]
     + E[(\nabla_{\mathbf{z}_j}\! f \cdot \boldsymbol{\xi}_j)^2] \notag \\
  &\quad + 2\,E[(\nabla_{\mathbf{z}_i}\! f \cdot \boldsymbol{\xi}_i)(\nabla_{\mathbf{z}_j}\! f \cdot \boldsymbol{\xi}_j)].
  \label{eq:pair-expand}
\end{align}
The cross-term vanishes because $\boldsymbol{\xi}_i$ and $\boldsymbol{\xi}_j$ are independent for $i \neq j$:
\begin{align}
  &E[(\nabla_{\mathbf{z}_i}\! f \cdot \boldsymbol{\xi}_i)(\nabla_{\mathbf{z}_j}\! f \cdot \boldsymbol{\xi}_j)] \notag \\
  &= (\nabla_{\mathbf{z}_i}\! f)^\top E[\boldsymbol{\xi}_i \boldsymbol{\xi}_j^\top] (\nabla_{\mathbf{z}_j}\! f) \notag \\
  &= (\nabla_{\mathbf{z}_i}\! f)^\top E[\boldsymbol{\xi}_i] E[\boldsymbol{\xi}_j]^\top (\nabla_{\mathbf{z}_j}\! f) = 0.
\end{align}
For the surviving diagonal terms, the isotropic covariance assumption $\mathrm{Cov}(\boldsymbol{\xi}_i) = \sigma_\xi^2 \mathbf{I}_d$ gives
\begin{align}
  &E[(\nabla_{\mathbf{z}_i}\! f \cdot \boldsymbol{\xi}_i)^2] \notag \\
  &\;= (\nabla_{\mathbf{z}_i}\! f)^\top \mathrm{Cov}(\boldsymbol{\xi}_i)\, (\nabla_{\mathbf{z}_i}\! f) = \sigma_\xi^2 \|\nabla_{\mathbf{z}_i}\! f\|^2,
  \label{eq:diag-term}
\end{align}
so that the per-pair perturbation energy is
\begin{multline}
  E[(\tilde{s}(i,j) - s(i,j))^2] = \\
  (\|\nabla_{\mathbf{z}_i}\! f_{ij}\|^2 + \|\nabla_{\mathbf{z}_j}\! f_{ij}\|^2) \cdot \sigma_\xi^2.
  \label{eq:pair-single}
\end{multline}
Under deep-layer homogenization, token representations are approximately identically distributed, so $\|\nabla_{\mathbf{z}_i} f_{ij}\|^2 \approx G_{\text{pair}}$ across pairs. Summing over all $\binom{N_p}{2}$ pairs:
\begin{align}
  V_{\text{pair}} &= \sigma_\xi^2 \!\sum_{i < j} 2G_{\text{pair}} \notag \\
  &= N_p(N_p\!-\!1) \cdot G_{\text{pair}} \cdot \sigma_\xi^2 \notag \\
  &= \Theta(N_p^2 \cdot G_{\text{pair}} \cdot \sigma_\xi^2).
  \label{eq:v-pair}
\end{align}
The key intuition is that each token's perturbation $\boldsymbol{\xi}_i$ appears in all $N_p - 1$ pairs involving token $i$, so a single perturbation simultaneously disrupts $O(N_p)$ scores; aggregated over all tokens, total disruption scales as $O(N_p^2)$.

\bigskip

\noindent\textbf{Derivation (b): Unary signal.}
For token $i$, $s(i) = g(\mathbf{z}_i, \boldsymbol{\theta})$ where $\boldsymbol{\theta} = \boldsymbol{\theta}(\mathbf{z}_1, \ldots, \mathbf{z}_{N_p})$ denotes aggregate population statistics (e.g., mean $\boldsymbol{\mu} = \frac{1}{N_p}\sum_j \mathbf{z}_j$, per-dimension variance $\boldsymbol{\sigma}^2$). Crucially, $\mathbf{z}_i$ influences $s(i)$ via two pathways. The \textbf{direct} pathway acts through $\mathbf{z}_i$ as the first argument of $g$, with gradient $\mathbf{a}_i \triangleq \nabla_{\mathbf{z}_i}^{\text{direct}} g \in \mathbb{R}^d$; this is an $O(1)$ effect. The \textbf{indirect} pathway acts through $\boldsymbol{\theta}$: every $\mathbf{z}_j$ (including $j = i$) shifts $\boldsymbol{\theta}$, which in turn shifts $s(i)$, with gradient $\mathbf{b}_j \triangleq (\nabla_{\boldsymbol{\theta}} g)^\top (\partial \boldsymbol{\theta}/\partial \mathbf{z}_j) \in \mathbb{R}^d$. Since $\boldsymbol{\theta}$ is a population average, $\partial \boldsymbol{\theta}/\partial \mathbf{z}_j = O(1/N_p)$, so each indirect coefficient satisfies $\|\mathbf{b}_j\| = O(1/N_p)$.

Separating the two pathways in a first-order expansion:
\begin{equation}
  \tilde{s}(i) - s(i) \approx \underbrace{\mathbf{a}_i^\top \boldsymbol{\xi}_i}_{\substack{\text{direct:}\\ O(1)}} + \underbrace{\textstyle\sum_{j=1}^{N_p} \mathbf{b}_j^\top \boldsymbol{\xi}_j}_{\substack{\text{indirect:}\\ O(1/\sqrt{N_p})\text{ by CLT}}}.
  \label{eq:unary-taylor}
\end{equation}
The indirect sum consists of $N_p$ independent zero-mean terms each of magnitude $O(1/N_p)$. By the central limit theorem, their sum has standard deviation $\sqrt{N_p} \cdot O(1/N_p) = O(1/\sqrt{N_p})$, vanishing relative to the $O(1)$ direct term as $N_p$ grows. This averaging effect is the structural advantage of unary signals: population statistics absorb individual perturbations.

To compute the expected squared perturbation, we merge the $j = i$ indirect term into the direct term:
\begin{equation}
  \tilde{s}(i) - s(i) \approx (\mathbf{a}_i + \mathbf{b}_i)^\top \boldsymbol{\xi}_i + \sum_{j \neq i} \mathbf{b}_j^\top \boldsymbol{\xi}_j.
\end{equation}
Cross-terms between $\boldsymbol{\xi}_i$ and $\boldsymbol{\xi}_j$ ($j \neq i$) again vanish by independence. Applying the isotropic covariance as in~\eqref{eq:diag-term}:
\begin{equation}
  E[(\tilde{s}(i) - s(i))^2]
  = \|\mathbf{a}_i + \mathbf{b}_i\|^2\,\sigma_\xi^2 + \sum_{j \neq i} \|\mathbf{b}_j\|^2\,\sigma_\xi^2.
  \label{eq:unary-per-token}
\end{equation}
Since $\|\mathbf{b}_i\| = O(1/N_p)$, expanding $\|\mathbf{a}_i + \mathbf{b}_i\|^2 = \|\mathbf{a}_i\|^2 + 2\,\mathbf{a}_i^\top \mathbf{b}_i + \|\mathbf{b}_i\|^2$ shows that the correction terms are $O(1/N_p)$ and $O(1/N_p^2)$, respectively. The cross-token indirect contribution is $\sum_{j \neq i} \|\mathbf{b}_j\|^2 = (N_p - 1) \cdot O(1/N_p^2) = O(1/N_p)$. Substituting back:
\begin{align}
  &E[(\tilde{s}(i) - s(i))^2] \notag \\
  &\;= \bigl(\|\mathbf{a}_i\|^2 + O(1/N_p)\bigr)\,\sigma_\xi^2 \notag \\
  &\;= \|\mathbf{a}_i\|^2\,\sigma_\xi^2\,\bigl(1 + O(1/N_p)\bigr).
\end{align}
Under gradient homogeneity ($\|\mathbf{a}_i\|^2 \approx G_{\text{unary}}$), summing over all $N_p$ tokens gives
\begin{align}
  V_{\text{unary}} &= N_p \cdot G_{\text{unary}} \cdot \sigma_\xi^2 \cdot \bigl(1 + O(1/N_p)\bigr) \notag \\
  &= \Theta(N_p \cdot G_{\text{unary}} \cdot \sigma_\xi^2).
  \label{eq:v-unary}
\end{align}

\noindent\textbf{The gap.} Combining~\eqref{eq:v-pair} and~\eqref{eq:v-unary}:
\begin{equation}
  \frac{V_{\text{pair}}}{V_{\text{unary}}} = \Theta\!\left(N_p \cdot \frac{G_{\text{pair}}}{G_{\text{unary}}}\right).
\end{equation}
The gap equals $\Theta(N_p)$ provided $G_{\text{pair}}$ and $G_{\text{unary}}$ are of the same order, which we verify below. The gap arises purely from the combinatorial structure: pairwise scoring creates $O(N_p^2)$ coupled sensitivity pathways, while unary scoring creates only $O(N_p)$.

\paragraph{Assumptions and limitations.}
The derivation relies on several assumptions, which we discuss in order of their role in the argument.

\emph{First-order Taylor approximation.} We neglect $O(\|\boldsymbol{\xi}\|^2)$ remainder terms. Under deep-layer homogenization, perturbations may not be small in absolute terms; however, the $\Theta(N_p^2)$ vs.\ $\Theta(N_p)$ gap arises from the \emph{number of sensitivity pathways} (combinatorial structure), not from the perturbation magnitude. Higher-order terms add corrections to both $V_{\text{pair}}$ and $V_{\text{unary}}$ that preserve the same combinatorial scaling, so the order-of-magnitude relationship is robust.

\emph{Isotropic perturbation covariance.} The identity $E[(\mathbf{c}^\top \boldsymbol{\xi})^2] = \sigma_\xi^2 \|\mathbf{c}\|^2$ used in~\eqref{eq:diag-term} requires $\mathrm{Cov}(\boldsymbol{\xi}_i) = \sigma_\xi^2 \mathbf{I}_d$. If perturbation components were correlated, the correct expression would be $\mathbf{c}^\top \mathrm{Cov}(\boldsymbol{\xi})\, \mathbf{c}$, which depends on the alignment between the gradient direction and the covariance structure. In practice, deep-layer distortions may introduce mild anisotropy, but this affects $V_{\text{pair}}$ and $V_{\text{unary}}$ symmetrically (both involve the same $\mathrm{Cov}(\boldsymbol{\xi})$), preserving the $\Theta(N_p)$ gap.

\emph{Gradient homogeneity.} We approximate $\|\nabla f_{ij}\|^2 \approx G_{\text{pair}}$ across pairs and $\|\nabla^{\text{direct}} g_i\|^2 \approx G_{\text{unary}}$ across tokens. This is justified under deep-layer homogenization, where token representations converge to similar distributions~\citep{raghu2021vision}, making the local geometry around each token, and hence gradient norms, approximately uniform.

\emph{Comparable gradient scales.} The gap is $\Theta(N_p)$ only if $G_{\text{pair}} / G_{\text{unary}} = \Theta(1)$, i.e., the two signal classes have gradient norms of the same order. Both $f$ and $g$ are scalar functions of $d$-dimensional inputs, so their gradient norms are determined by the intrinsic sensitivity of the scoring function, not by $N_p$. For the concrete signals in our evaluation, cosine similarity (pairwise, $G_{\text{pair}} = O(1/d)$) and diagonal Mahalanobis distance (unary, $G_{\text{unary}} = O(1/d)$), both scales are $O(1/d)$, satisfying the assumption. More generally, any scoring function whose gradient norm depends on $d$ but not on $N_p$ will satisfy this condition. If a pathological unary signal had $G_{\text{unary}} \gg G_{\text{pair}}$, the gap could be partially offset; however, our experimental data confirms that the gap materializes for the signals actually used (Table~\ref{tab:complementarity}).

\emph{I.i.d.\ perturbations.} In practice, perturbations are non-uniform: earlier suboptimal reductions cause structured, token-dependent distortions. This does not change the $\Theta(N_p^2)$ vs.\ $\Theta(N_p)$ scaling, which arises from the combinatorial structure of pairwise vs.\ unary scoring, not from the perturbation distribution.

\emph{From $V$ to $\rho_s$.} Larger total perturbation energy $V$ implies more ranking disruption, but the exact mapping depends on the score margin distribution. Under homogenization, pairwise margins shrink (cosine similarities converge), further amplifying ranking disruption for a given $V$, consistent with the empirically observed $\rho_s$ gap (Table~\ref{tab:complementarity}: pairwise $\rho_s \approx 0.27$ vs.\ unary $\rho_s \approx 0.37$--$0.48$ in deep layers).


\subsection{Pairwise Ranking Stability Within the Merge Pool}
\label{app:pool-rho-s}

Our signal-class analysis (Section~\ref{sec:paradigm}) targets the \emph{importance scoring} stage: pairwise signals fail to reliably rank \emph{which tokens matter}, causing catastrophic triage errors. The bipartite matching within $\mathcal{M}$ uses pairwise cosine for a fundamentally different purpose: deciding \emph{which similar tokens to combine}, among tokens already triaged as expendable.
Three factors mitigate the diagnosed instability:
(i)~the pool is restricted to medium-importance tokens ($|s_{\text{final}}| \le \tau$), so even worst-case mismatches pair tokens of comparable importance, bounding $\delta$ per error (in contrast to global scoring, where a single ranking error can expose a critical foreground token to merging with a background token);
(ii)~the pool size $|\mathcal{M}| \ll N_p$ reduces the number of jointly perturbed pairwise entries from $O(N_p^2)$ to $O(|\mathcal{M}|^2)$;
(iii)~empirically, pairwise $\rho_s$ measured \emph{within} $\mathcal{M}$ is substantially higher than the global value (see table below).

To verify this quantitatively, we measure pairwise $\rho_s$ \emph{within} the merge pool $\mathcal{M}$ and compare it to the global $\rho_s$ measured over all $N_p$ tokens. We follow the same protocol as Section~\ref{sec:pairwise-degradation}: for each image, we compute the pairwise K-vector cosine similarity matrix from both its clean and corrupted version, but restricted to the tokens satisfying $|s_{\text{final}}| \le \tau$ (i.e., the merge pool under CATIS triage). We report mean $\rho_s$ over $n{=}1{,}000$ images on DeiT-3 Base.

\begin{center}
\small
\begin{tabular}{lcc}
\toprule
\textbf{Layer range} & \textbf{Global $\rho_s$} & \textbf{Pool $\rho_s$} \\
\midrule
Shallow (1--4) & 0.88 & 0.85 \\
Mid (5--8) & 0.62 & 0.64 \\
Deep (9--11) & 0.27 & 0.41 \\
\bottomrule
\end{tabular}
\end{center}

In deep layers, where the diagnosed pairwise instability is most severe, $\rho_s$ within $\mathcal{M}$ is 52\% higher than the global value ($0.41$ vs.\ $0.27$). Two factors contribute: (i)~the pool size $|\mathcal{M}| \ll N_p$ reduces the number of jointly perturbed entries; (ii)~triage removes the highest- and lowest-importance tokens, leaving a more homogeneous subset where pairwise cosine is more stable. In terms of the dual-channel distortion model (Eq.~\ref{eq:catis-rec}), higher pool-internal $\rho_s$ directly corresponds to lower $\epsilon_m^{(l)}$ (the merge channel's distortion rate), reducing the merge contribution to per-layer distortion. Even in the residual case where pool-internal $\rho_s$ is moderate, the key point remains: all tokens in $\mathcal{M}$ have comparable importance ($|s_{\text{final}}| \le \tau$), so mismatches pair tokens of similar semantic value, bounding $\delta_m$ per operation far below the global worst case.


\section{Derivation of norm$_F$ from Mahalanobis Distance}
\label{app:norm-f}

Given the $N_p$ patch-token representations $\{\mathbf{z}_i\}_{i=1}^{N_p} \subset \mathbb{R}^d$ at a particular layer, define the population mean and covariance:
\begin{equation}
  \boldsymbol{\mu} = \frac{1}{N_p}\sum_{i=1}^{N_p} \mathbf{z}_i, \quad
  \boldsymbol{\Sigma} = \frac{1}{N_p}\sum_{i} (\mathbf{z}_i - \boldsymbol{\mu})(\mathbf{z}_i - \boldsymbol{\mu})^\top.
\end{equation}
The Mahalanobis distance of token $i$ from this population is:
\begin{equation}
  M_i = \sqrt{(\mathbf{z}_i - \boldsymbol{\mu})^\top \boldsymbol{\Sigma}^{-1} (\mathbf{z}_i - \boldsymbol{\mu})}.
  \label{eq:mahal-full}
\end{equation}

\textbf{Rank deficiency.}
For all standard ViT configurations, $N_p < d$ (e.g., $N_p{=}196$ vs.\ $d{=}768$ for ViT-Base), so $\text{rank}(\boldsymbol{\Sigma}) \le N_p - 1 < d$ and $\boldsymbol{\Sigma}$ is singular. Even if a pseudo-inverse were used, the $O(d^3)$ computation per image per layer contradicts the acceleration objective.

\textbf{Diagonal approximation.}
We approximate $\boldsymbol{\Sigma}$ by its diagonal:
\begin{equation}
  \boldsymbol{\Sigma} \approx \text{diag}(\sigma_1^2, \ldots, \sigma_d^2), \;\;
  \sigma_k^2 = \frac{1}{N_p}\!\sum_{i}(z_{i,k} - \mu_k)^2.
\end{equation}
This is valid when off-diagonal correlations are small, i.e.\ the feature dimensions are approximately independent. We verify this empirically via $\rho_\text{off}$ (defined in Appendix~\ref{app:rho-off-analysis}, Eq.~\eqref{eq:rho-off}) on vanilla models (Figures~\ref{fig:rho-off-vanilla-a} and~\ref{fig:rho-off-vanilla-b}): across 4~representative models (DeiT-3 Base/Large, ViT-Base/Large) $\times$ 5~datasets $\times$ 100~images, $\rho_\text{off}$ ranges from 0.09 to 0.24, with cross-dataset variance $<0.03$, confirming that this is an architectural property rather than a data-dependent coincidence.

\textbf{Reduction to norm$_F$.}
Substituting the diagonal approximation into Eq.~\ref{eq:mahal-full}:
\begin{align}
  M_i^2 &= (\mathbf{z}_i - \boldsymbol{\mu})^\top \text{diag}(\sigma_1^{-2}, \ldots, \sigma_d^{-2})\, (\mathbf{z}_i - \boldsymbol{\mu}) \notag \\
         &= \sum_{k=1}^{d} \frac{(z_{i,k} - \mu_k)^2}{\sigma_k^2}
          = \left\| \frac{\mathbf{z}_i - \boldsymbol{\mu}}{\boldsymbol{\sigma}} \right\|_2^2
\end{align}
where the division is element-wise. Taking the square root recovers norm$_F$ (Eq.~\ref{eq:norm-f}):
\begin{equation}
  s_{\text{nf},i} = M_i = \left\| \frac{\mathbf{z}_i - \boldsymbol{\mu}}{\boldsymbol{\sigma}} \right\|_2.
\end{equation}
Each component $(z_{i,k} - \mu_k)/\sigma_k$ is the z-score of token $i$ along dimension $k$; norm$_F$ is therefore the $L_2$ norm of the per-dimension z-scores, or equivalently the square root of the diagonal Mahalanobis distance. The computation requires $O(N_p \cdot d)$ operations with no pairwise calculations.

Table~\ref{tab:norm-f-ablation} ablates each step of this derivation on ViT-Base. The z-score normalization yields +0.88pp on IN-A and +0.38pp on shift average over the raw $L_2$ baseline, with gains concentrated on the shifted benchmarks, consistent with the theoretical prediction that per-dimension variance normalization matters most under distribution shift.

\begin{table*}[t]
  \caption{norm$_F$ signal ablation on ViT-Base. All variants use the same triage framework; only the importance signal differs.}
  \label{tab:norm-f-ablation}
  \centering
  \small
  \begin{tabular}{@{}clccccc@{}}
    \toprule
    \# & Formula & IN-A & IN-R & IN-SK & IN-1K & Shift avg \\
    \midrule
    1 & Raw $L_2$: $\|\mathbf{z}_i\|_2$ & 18.12 & 38.11 & 16.12 & 70.53 & 24.12 \\
    2 & Euclidean: $\|\mathbf{z}_i - \boldsymbol{\mu}\|_2$ & 18.05 & 37.94 & 16.31 & 70.51 & 24.10 \\
    \textbf{3} & \textbf{norm$_F$}: $\left\|\frac{\mathbf{z}_i - \boldsymbol{\mu}}{\boldsymbol{\sigma}}\right\|_2$ & \textbf{19.00} & \textbf{38.16} & \textbf{16.33} & \textbf{70.54} & \textbf{24.50} \\
    \bottomrule
  \end{tabular}
\end{table*}

\section{CATIS as a Layer-Local Operator: Functional Composition}
\label{app:algorithm}

The full per-layer reduction acts on the post-attention residual stream as a composition of three functionally distinct operators, each derived from one of P1--P3 (Sections~\ref{sec:norm-f}--\ref{sec:triage}). Let $\mathbf{Z}^{(l)} = \mathrm{LN}_1(\mathbf{X}_{\text{in}}^{(l)})$ denote the pre-attention features of layer $l$ and $\mathbf{H}^{(l)} = \mathbf{X}_{\text{in}}^{(l)} + \mathrm{Attn}(\mathbf{Z}^{(l)})$ the post-attention residual on which token reduction is applied. The per-layer operator $\mathcal{O}^{(l)}$ admits the high-level decomposition
\begin{equation*}
  \mathcal{O}^{(l)} \;=\; \mathcal{R}^{(l)} \,\circ\, \mathcal{T}_{\mathcal{C}}^{(l)} \,\circ\, \mathcal{S}_{\Theta_l}^{(l)},
\end{equation*}
in which $\mathcal{S}_{\Theta_l}^{(l)}$ is the \emph{importance-functional} operator (a fused unary functional satisfying P1--P2; Sections~\ref{sec:norm-f},~\ref{sec:cls}), $\mathcal{T}_{\mathcal{C}}^{(l)}$ is the \emph{triage} operator (a symmetric, scale-adaptive partition into $\mathcal{P}^{(l)} \,\dot\cup\, \mathcal{M}^{(l)} \,\dot\cup\, \mathcal{E}^{(l)}$ together with an allocation policy splitting the budget $r$ between channels; Section~\ref{sec:triage}), and $\mathcal{R}^{(l)}$ is the \emph{reduction} operator (lossless preservation of $\mathcal{P}^{(l)}$, eviction of the lowest-scoring members of $\mathcal{E}^{(l)}$ up to the eviction sub-budget, and bipartite merging restricted to $\mathcal{M}^{(l)}$ on the remaining sub-budget, with size-weighted averaging on the K-vector cosine geometry of $\mathbf{Z}^{(l)}$). The collective parameter $\Theta_l = (\gamma_l, w_{\text{cls},l}, \tau_l, \mathrm{evict\_ratio}_l, L_{\text{start}})$ collects the model-family-specific scalars introduced across Sections~\ref{sec:norm-f}--\ref{sec:triage}; we treat $\Theta_l$ as a single tunable handle whose admissible component-wise intervals are reported in Appendix~\ref{app:hparams}, with each interval derived from, and bounded by, the diagnostic predicates of Section~\ref{sec:diagnosis} rather than from per-benchmark accuracy maximisation.

\textbf{Activation schedule.} The contextual branch of $\mathcal{S}_{\Theta_l}^{(l)}$ is suppressed in the first few transformer blocks until enough cross-layer history has accumulated for the discrete temporal-derivative term in Eq.~\ref{eq:momentum} to be meaningful; up to that depth, the operator collapses to its activation-only branch. The exact transition depth is model-family-specific and is chosen by the same family-level optimization procedure that selects $\Theta_l$.

\textbf{Output.} Concatenating the protected, merged, and surviving subsets and passing the result through the customary residual MLP block, $\mathbf{Y}^{(l)} = \mathcal{R}^{(l)}\bigl(\mathbf{H}^{(l)},\, \mathcal{T}_{\mathcal{C}}^{(l)}(\mathcal{S}_{\Theta_l}^{(l)}(\mathbf{Z}^{(l)}))\bigr) + \mathrm{FFN}(\mathrm{LN}_2(\cdot))$, yields a token set of cardinality $N_p^{(l)} - r$, matching the budget contract of every baseline in Table~\ref{tab:main}. We deliberately omit a step-by-step pseudocode listing in this paper; the operator-composition above, together with the formal definitions in Sections~\ref{sec:norm-f}--\ref{sec:triage}, fully specifies the structural design without committing to a particular numerical instantiation, and is the level of description at which the diagnostic predictions of Section~\ref{sec:diagnosis} are independent of implementation details.

\section{Signal Complementarity Visualization}
\label{app:signal-heatmaps}

\begin{figure*}[t]
  \centering
  \includegraphics[width=\textwidth]{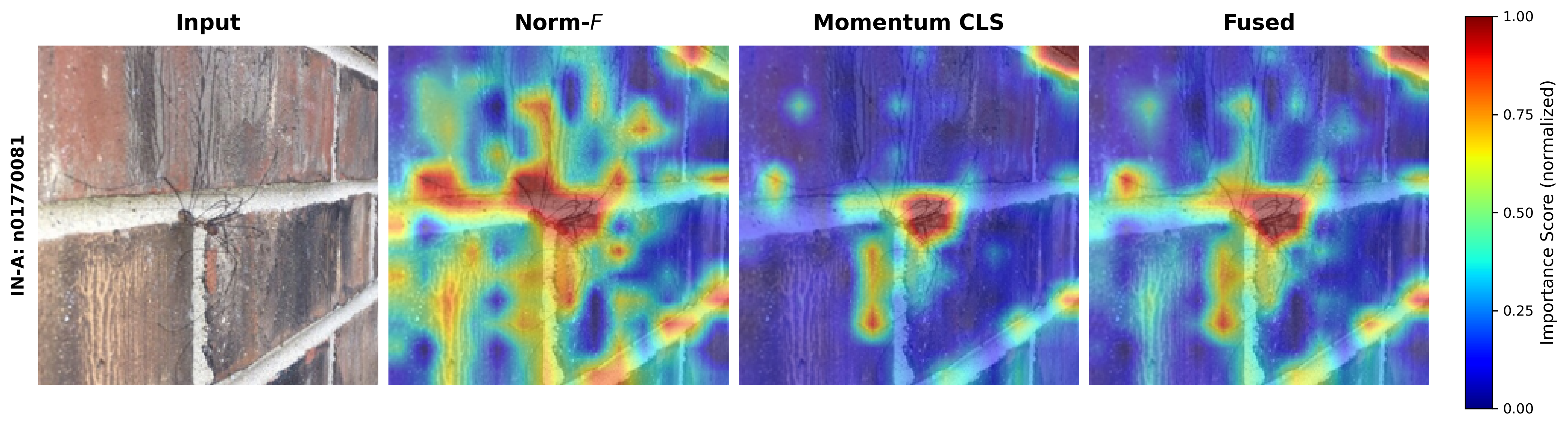}
  \caption{Signal complementarity (DeiT-3 Base, layer~8; ImageNet-A, class \emph{harvestman}). \textbf{norm$_F$} covers the full foreground but also activates on background texture. \textbf{Momentum CLS} concentrates on the semantic center but misses peripheral structures. The \textbf{Fused} score inherits coverage from norm$_F$ and selectivity from CLS.}
  \label{fig:signal-heatmaps}
\end{figure*}

Figure~\ref{fig:signal-heatmaps} visualizes the two unary signals and their fusion on a representative image, illustrating the complementarity discussed in Section~\ref{sec:cls}.

\section{Diagonal Approximation Validation ($\rho_\text{off}$ on Vanilla Models)}
\label{app:rho-off-vanilla}

\begin{figure*}[t]
  \centering
  \begin{subfigure}[t]{0.48\textwidth}
    \centering
    \includegraphics[width=\linewidth]{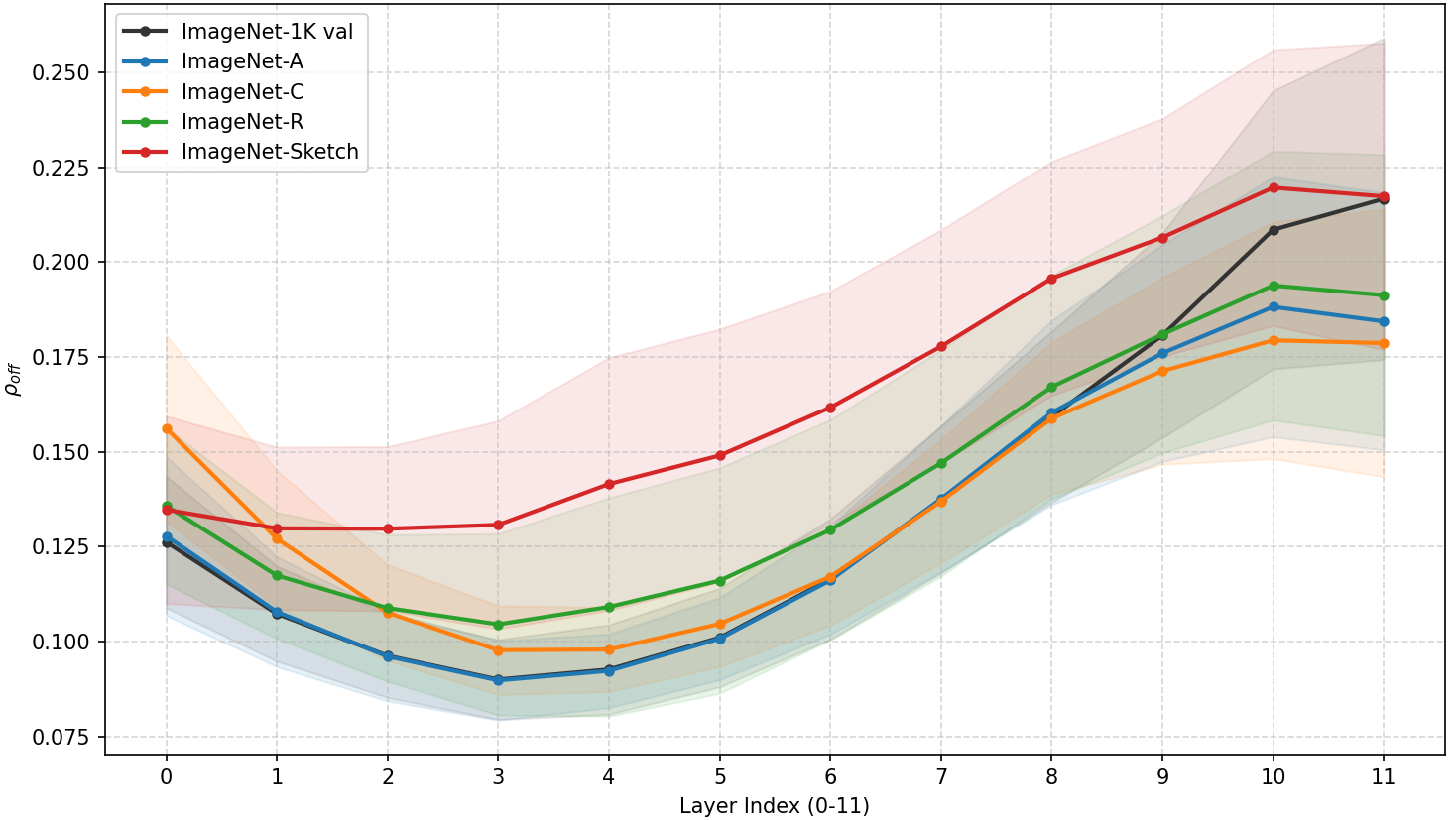}
    \caption{DeiT-3 Base (12 layers)}
  \end{subfigure}
  \hfill
  \begin{subfigure}[t]{0.48\textwidth}
    \centering
    \includegraphics[width=\linewidth]{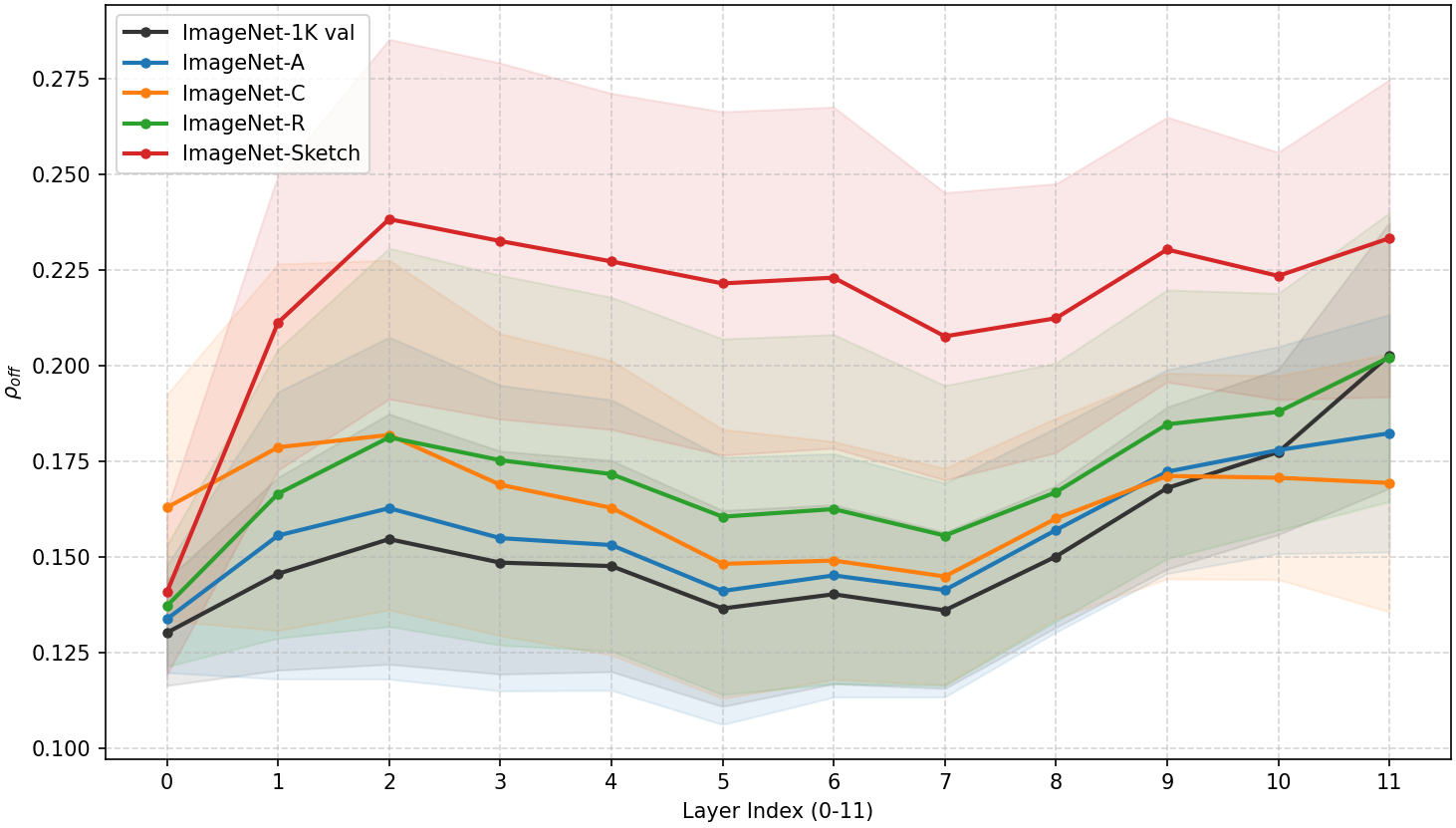}
    \caption{ViT-Base (12 layers)}
  \end{subfigure}
  \caption{Off-diagonal correlation $\rho_\text{off}$ on vanilla models (no token reduction), 12-layer architectures. All values stay below 0.25; curves overlap across datasets ($< 0.03$ variance).}
  \label{fig:rho-off-vanilla-a}
\end{figure*}

\begin{figure*}[t]
  \centering
  \begin{subfigure}[t]{0.48\textwidth}
    \centering
    \includegraphics[width=\linewidth]{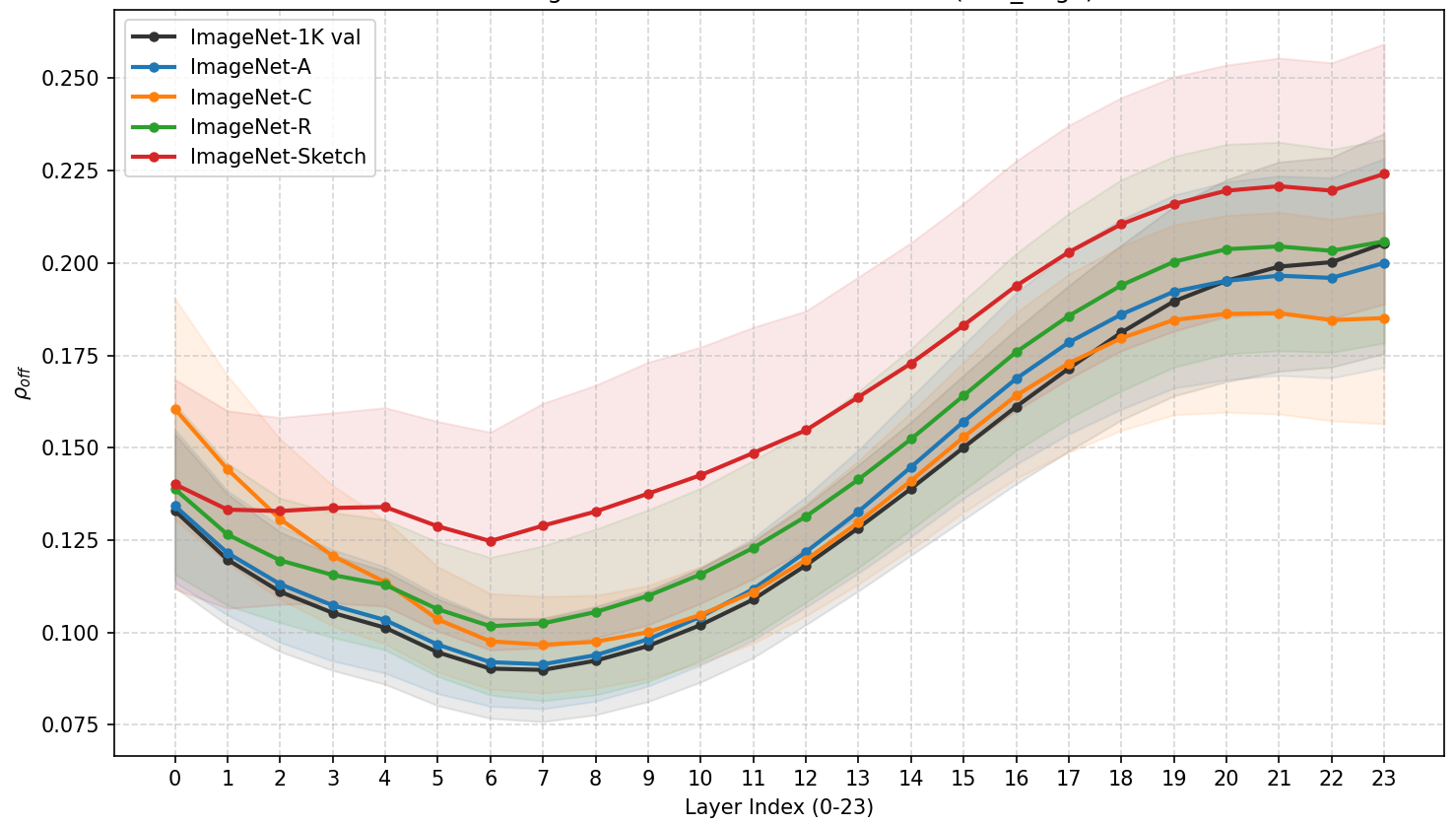}
    \caption{DeiT-3 Large (24 layers)}
  \end{subfigure}
  \hfill
  \begin{subfigure}[t]{0.48\textwidth}
    \centering
    \includegraphics[width=\linewidth]{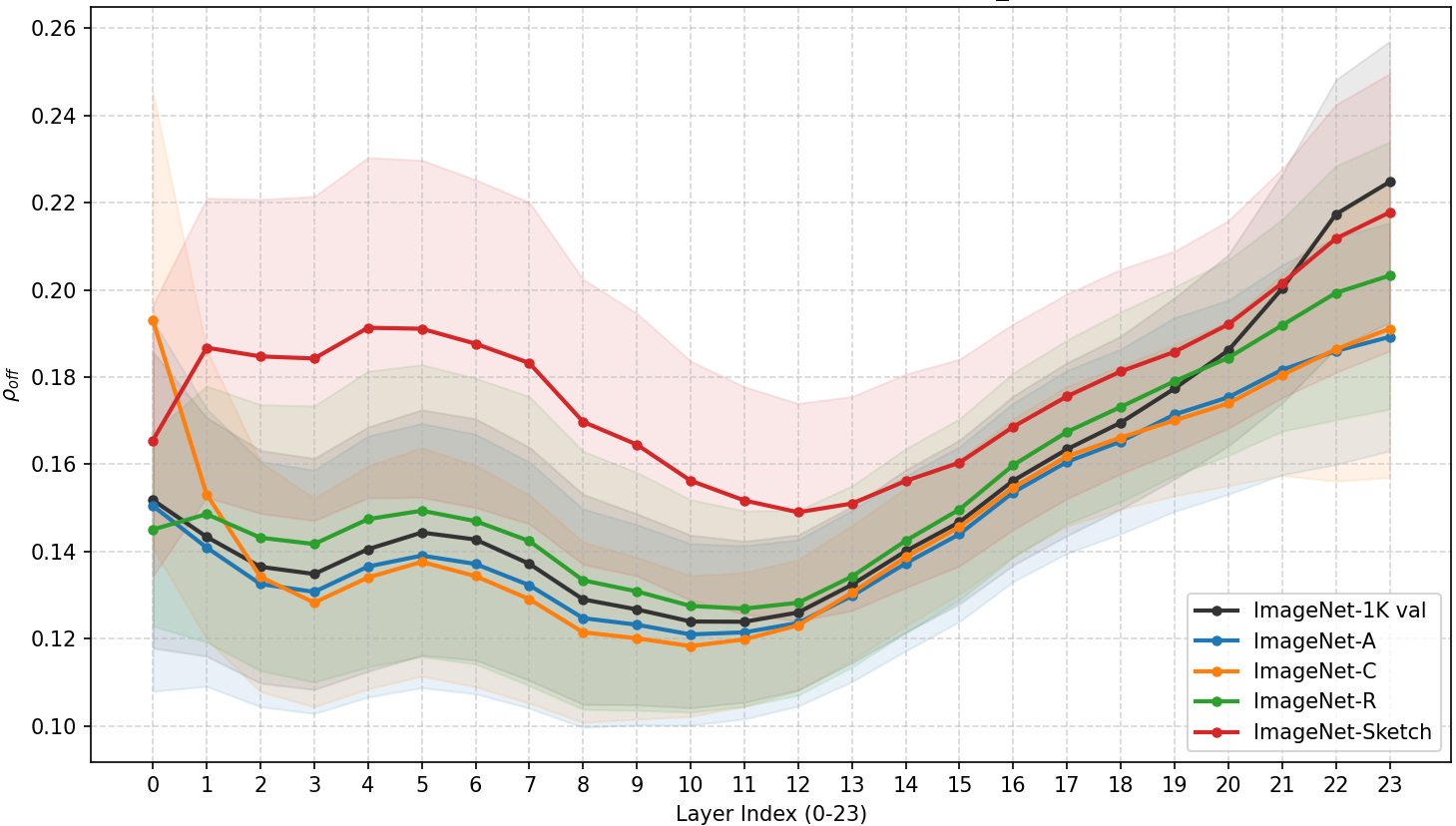}
    \caption{ViT-Large (24 layers)}
  \end{subfigure}
  \caption{Same as Figure~\ref{fig:rho-off-vanilla-a} for 24-layer models.}
  \label{fig:rho-off-vanilla-b}
\end{figure*}

Figures~\ref{fig:rho-off-vanilla-a} and~\ref{fig:rho-off-vanilla-b} show $\rho_\text{off}$ across layers for 4~models on 5~datasets (clean + 4~shifted). All values remain below 0.25, and the cross-dataset variance is $< 0.03$, confirming that near-independence of feature dimensions is an architectural property.

\section{$\rho_\text{off}$ Under Token Reduction: Complete Results}
\label{app:rho-off-full}

All 12~plots ($6$ models $\times$ $2$ conditions per model: IN-1K and shifted, with shifted panels pooling robustness benchmarks) exhibit the same pattern: ToMe/MCTF surge at high $r$; PiToMe diverges early but plateaus; CATIS tracks Vanilla.

\section{Detailed Experimental Analysis}
\label{app:detailed-results}

This section provides detailed per-model analysis summarized in Section~\ref{sec:main-results}.

\textbf{ViT-Large ($r{=}11$, 63\% GFLOPs reduction).} CATIS retains 96.9\% of vanilla IN-1K accuracy (81.03 vs.\ 83.64), while ToMe/MCTF collapse to $\sim$43\% ($-$40pp) and PiToMe achieves only 65.49\%. On shifted benchmarks, CATIS outperforms PiToMe by +19.21pp on Shift avg (49.51 vs.\ 30.30). Critically, CATIS nearly matches vanilla on IN-A (43.65 vs.\ 43.27) and IN-R (59.76 vs.\ 60.89), while ToMe/MCTF collapse to single digits on IN-A (6.57/6.00). The 24-layer depth causes pairwise signals to degrade severely ($\rho_s \approx 0.27$ in deep layers), triggering the error amplifier at moderate $r$.

\textbf{DeiT-3-Base ($r{=}23$, 64\% GFLOPs reduction).} This presents a challenging scenario: moderate depth (12 layers) but extreme token reduction. CATIS achieves 73.85\% IN-1K ($-$9.53pp from vanilla) versus 63.50\% for the best baseline MCTF ($-$19.88pp). CATIS maintains 69.2\% of vanilla IN-A (25.44 vs.\ 36.75) while MCTF retains only 26.9\% (9.89 vs.\ 36.75). The triage mechanism is decisive: at $r{=}23$, each layer removes a large fraction of tokens, making protection critical.

\textbf{ViT-Small ($r{=}13$, 41\% GFLOPs reduction).} This represents a regime where the error amplifier is below threshold. CATIS still leads but margins narrow: +0.29pp on IN-1K over ToMe (70.93 vs.\ 70.64), +0.32pp on Shift avg. This confirms that CATIS's advantage stems from higher signal quality, visible even when the amplifier is inactive.

\textbf{DINOv3~\citep{simeoni2025dinov3} ($r{=}4$, 40\% GFLOPs reduction).} This tests generalization to self-supervised pretraining. CATIS achieves highest accuracy on all metrics, with the largest gains on the hardest benchmark: 86.15\% IN-A (+2.39pp over ToMe), nearly preserving vanilla accuracy (86.53, only $-$0.38pp) while baselines lose 2.77--3.12pp. On IN-1K, CATIS leads with 87.09\% (+0.09pp over MCTF), demonstrating that unary signals generalize across pretraining strategies.

\textbf{DeiT-3-Large ($r{=}11$).} CATIS outperforms the best baseline by +6.96pp on IN-1K (75.27 vs.\ 68.31) and +6.41pp on Shift avg (34.62 vs.\ 28.21). The 24-layer depth amplifies the gap relative to 12-layer models.

\textbf{DeiT-3-Small ($r{=}23$).} Under extreme compression (63\% reduction), CATIS achieves 58.20\% IN-1K versus 45.11\% for PiToMe (+13.09pp). All baselines collapse below 46\%, while CATIS maintains usable accuracy even at this aggressive operating point.

\section{Full Experimental Setup and Benchmarks}
\label{app:full-results}
\label{app:hparams}
\noindent The collective parameter $\Theta = (\gamma,\, w_{\text{cls}},\, \tau,\, \mathrm{evict\_ratio},\, L_{\text{start}})$ introduced in Sections~\ref{sec:norm-f}--\ref{sec:triage} is selected once per model family rather than per dataset or per operating point. The selection itself is driven by the diagnostic predicates of Section~\ref{sec:diagnosis} (raise the per-layer $\rho_s$ on a held-out subset of ImageNet-1K; keep $\rho_\text{off}$ within the vanilla-model envelope of Appendix~\ref{app:rho-off-vanilla}) rather than by direct accuracy maximisation on any single benchmark, which keeps the model-family-level choice transferable across the clean and shifted evaluations of Table~\ref{tab:main} without per-benchmark re-tuning.

\noindent\textbf{Admissible regions.} The diagnostic objective restricts each component of $\Theta$ to a structurally interpretable, strictly bounded interval rather than to a single literal value; we describe these intervals only by their qualitative position on the underlying axis and by the failure mode encountered when either endpoint is crossed, leaving the precise numerical bracket to be re-derived from the diagnostic predicates of Section~\ref{sec:diagnosis} on the target backbone. The momentum coefficient $\gamma$ occupies a strict sub-interval of the half-line~$\mathbb{R}_{\ge 0}$ in which the depth-derivative term of Eq.~\ref{eq:momentum} re-weights, but does not invert, the present-layer attention vector; values below the lower endpoint degenerate the operator towards the single-layer baseline $\gamma{=}0$, while values above the upper endpoint over-amplify the discrete derivative and degrade $\rho_s$ on shallow layers. The fusion weight $w_{\text{cls}}$ in Eq.~\ref{eq:fusion} occupies a symmetric interior sub-interval of $[0,1]$ in which neither standardised functional dominates and the convex combination retains both depth-band complementarities of Table~\ref{tab:complementarity}; pushing $w_{\text{cls}}$ towards either endpoint of $[0,1]$ recovers a single-functional baseline and discards the other depth band. The triage radius $\tau \in \mathbb{R}_{\ge 0}$ defining the level sets of Eq.~\ref{eq:triage} lies in a moderate-confidence band on the standardised score axis, large enough that the partition is robust to per-image score noise yet small enough that $\mathcal{P}^{(l)} \cup \mathcal{E}^{(l)}$ remains non-empty in deep layers, where its lower bound is set by per-image stability and its upper bound by the requirement that the dual-channel recurrence of Eq.~\ref{eq:dual-channel} not collapse back to the single-channel form. The intra-budget allocation scalar $\mathrm{evict\_ratio} \in [0,1]$ in Eq.~\ref{eq:allocation} occupies an interior sub-interval whose lower bound keeps the eviction sub-budget non-trivially populated and whose upper bound prevents it from consuming the entire merge channel; the saturation $\min(\cdot,\, |\mathcal{E}^{(l)}|)$ then routes any unfilled allocation back to merge whenever $\mathcal{E}^{(l)}$ is too small to absorb it. Finally, the activation depth of the contextual branch satisfies $L_{\text{start}} \in \mathbb{N}_{>0}$ at the smallest depth at which the cross-layer history of $A_{\text{cls}}^{(l)}$ accumulates enough variance for the discrete derivative of Eq.~\ref{eq:momentum} to carry signal above the per-layer attention noise floor; below this depth, $\gamma_{\text{eff}} = 0$ and the operator collapses to its activation-only branch by construction.

\noindent\textbf{Selection procedure.} Within each family, the precise instantiation inside these intervals is obtained by a diagnostic-driven search restricted to the predicates above, rather than to validation accuracy on any benchmark of Table~\ref{tab:main}; this is what makes the choice transferable across clean and shifted evaluation without per-benchmark re-tuning. The intervals are descriptions of the diagnostic-feasible set rather than fixed numerical recipes, and reproductions on a new backbone or a new token budget $r$ should re-derive them from the diagnostic objectives stated above, applied to the target backbone, in line with the operator-level description of Appendix~\ref{app:algorithm}.

\noindent\textbf{Benchmark descriptions.} \textbf{ImageNet-A}~\citep{hendrycks2021natural} contains naturally occurring hard examples that cause standard classifiers to fail, covering 200 ImageNet classes. \textbf{ImageNet-R}~\citep{hendrycks2021many} includes renditions (art, cartoons, sketches, etc.) of 200 ImageNet classes. \textbf{ImageNet-Sketch}~\citep{wang2019learning} consists of sketch images for all 1000 ImageNet classes, collected from Google Image Search. \textbf{ImageNet-C}~\citep{hendrycks2019benchmarking} applies 15 algorithmically generated corruptions (noise, blur, weather, digital) at 5 severity levels to the ImageNet validation set; we report mean Corruption Accuracy (mCA).

\noindent Full per-dataset results for all 7~models appear in the main text (Table~\ref{tab:main}). Supplementary Pareto curves are collected below.

\section{Additional Pareto Curves}
\label{app:pareto-full}

\begin{figure*}[t]
  \centering
  \begin{subfigure}[t]{0.48\textwidth}
    \centering
    \includegraphics[width=\linewidth]{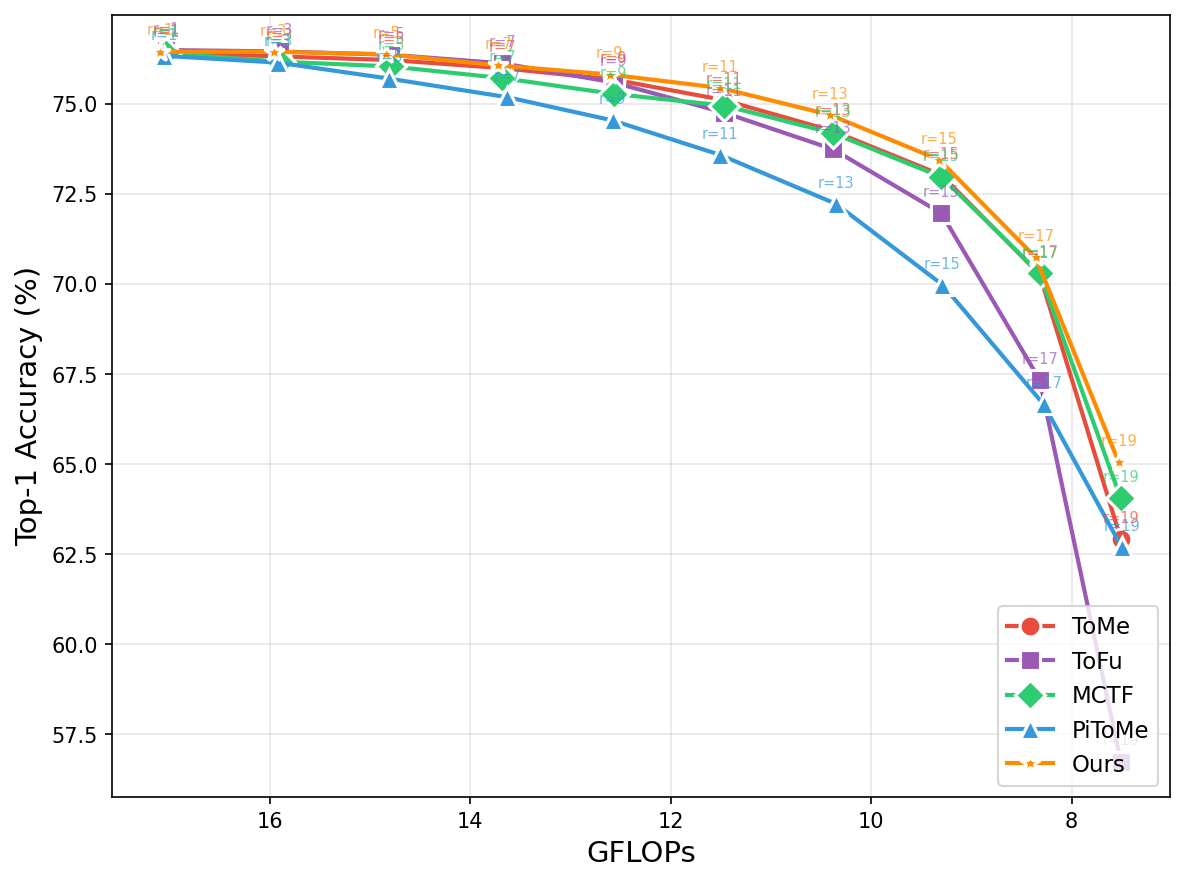}
    \caption{ViT-Base --- IN-1K}
  \end{subfigure}
  \hfill
  \begin{subfigure}[t]{0.48\textwidth}
    \centering
    \includegraphics[width=\linewidth]{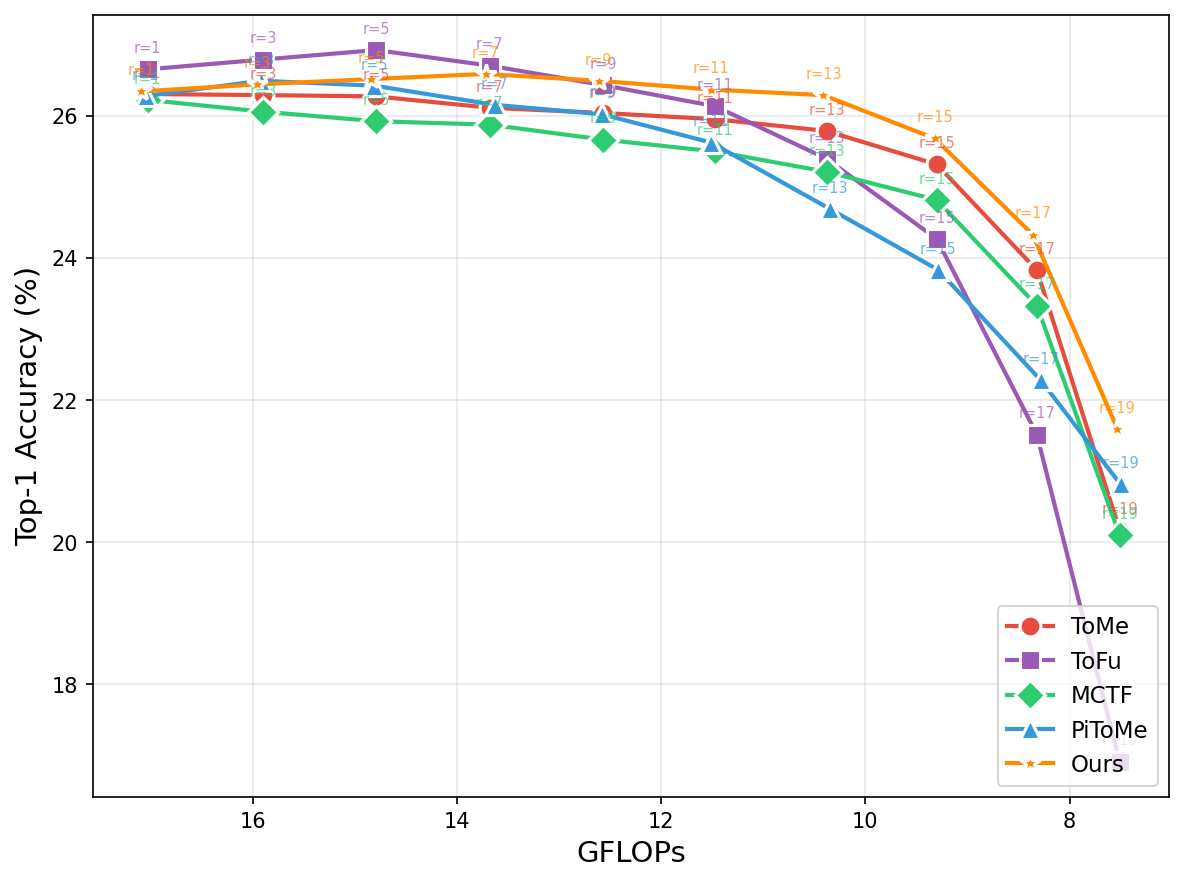}
    \caption{ViT-Base --- shifted}
  \end{subfigure}
  \\[4pt]
  \begin{subfigure}[t]{0.48\textwidth}
    \centering
    \includegraphics[width=\linewidth]{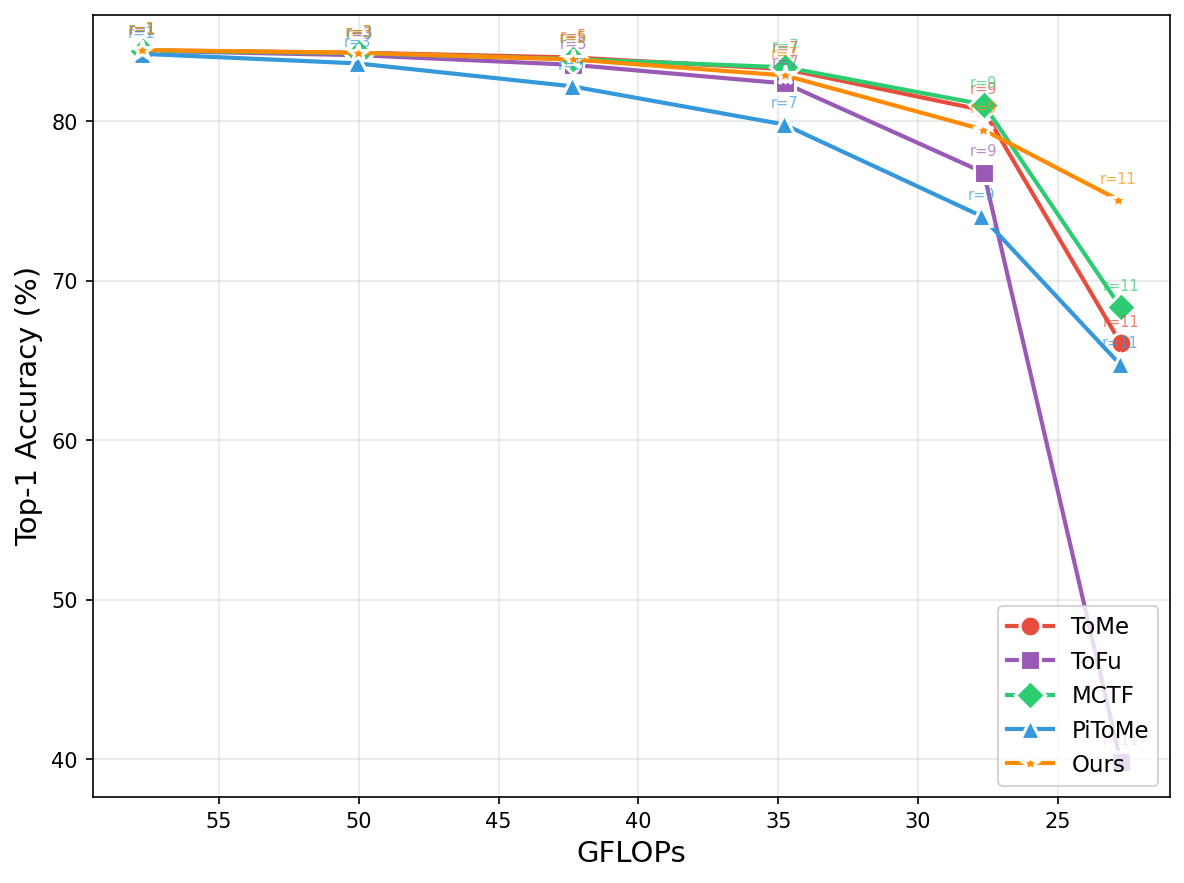}
    \caption{DeiT-3 Large --- IN-1K}
  \end{subfigure}
  \hfill
  \begin{subfigure}[t]{0.48\textwidth}
    \centering
    \includegraphics[width=\linewidth]{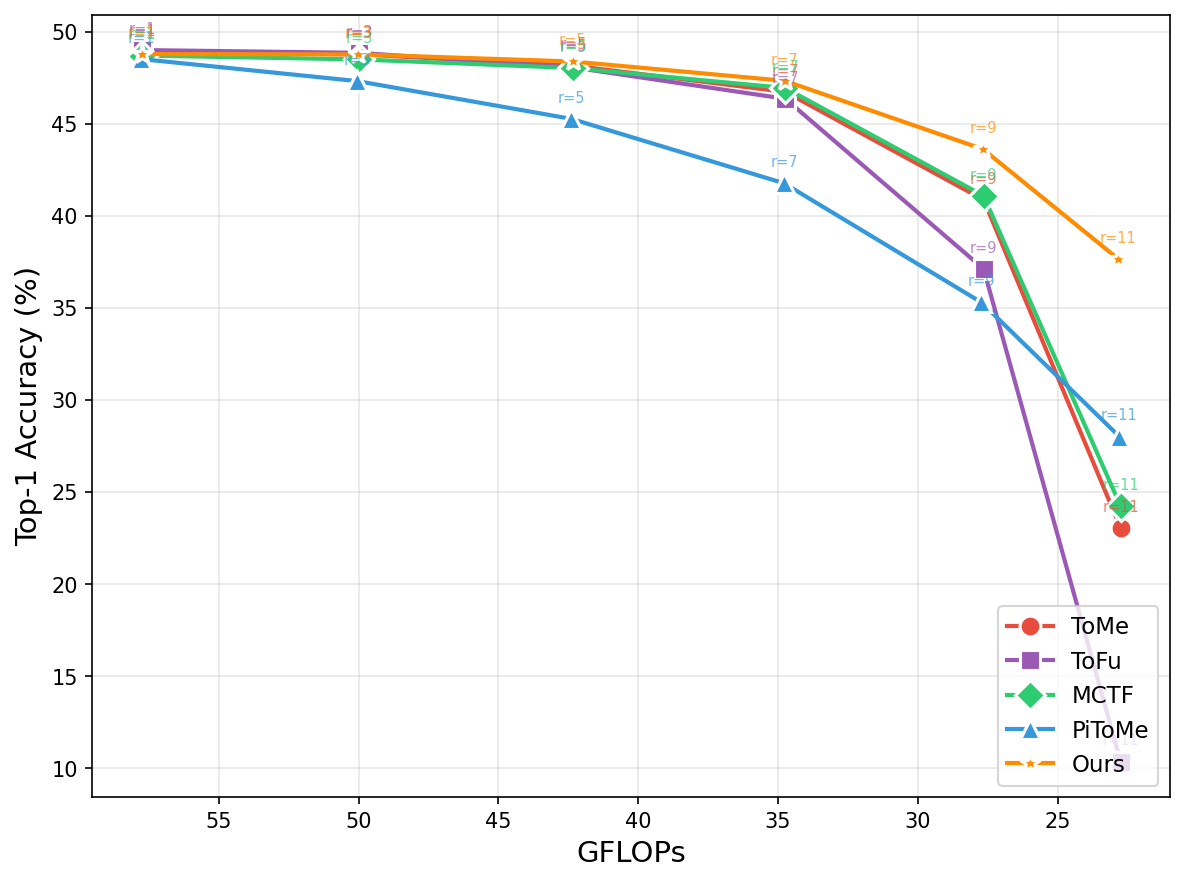}
    \caption{DeiT-3 Large --- shifted}
  \end{subfigure}
  \\[4pt]
  \begin{subfigure}[t]{0.48\textwidth}
    \centering
    \includegraphics[width=\linewidth]{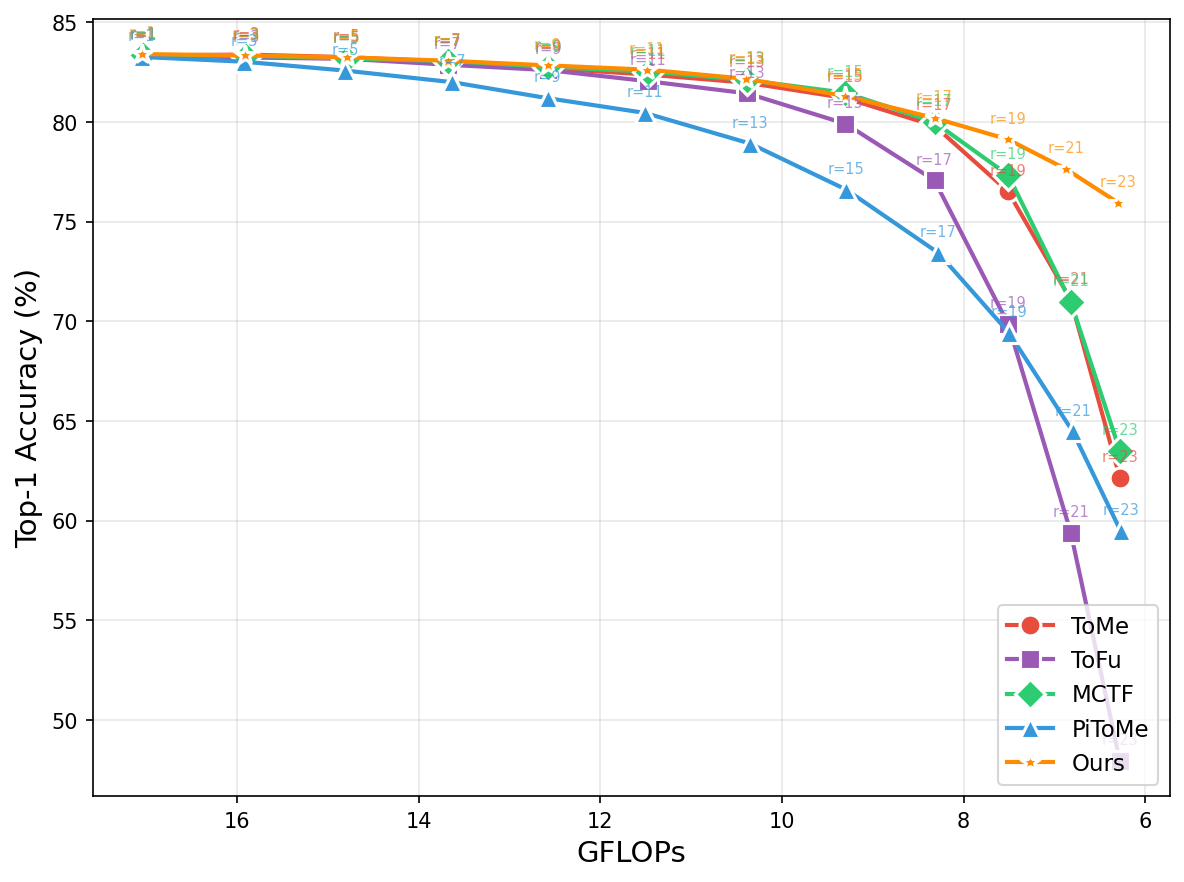}
    \caption{DeiT-3 Base --- IN-1K}
  \end{subfigure}
  \hfill
  \begin{subfigure}[t]{0.48\textwidth}
    \centering
    \includegraphics[width=\linewidth]{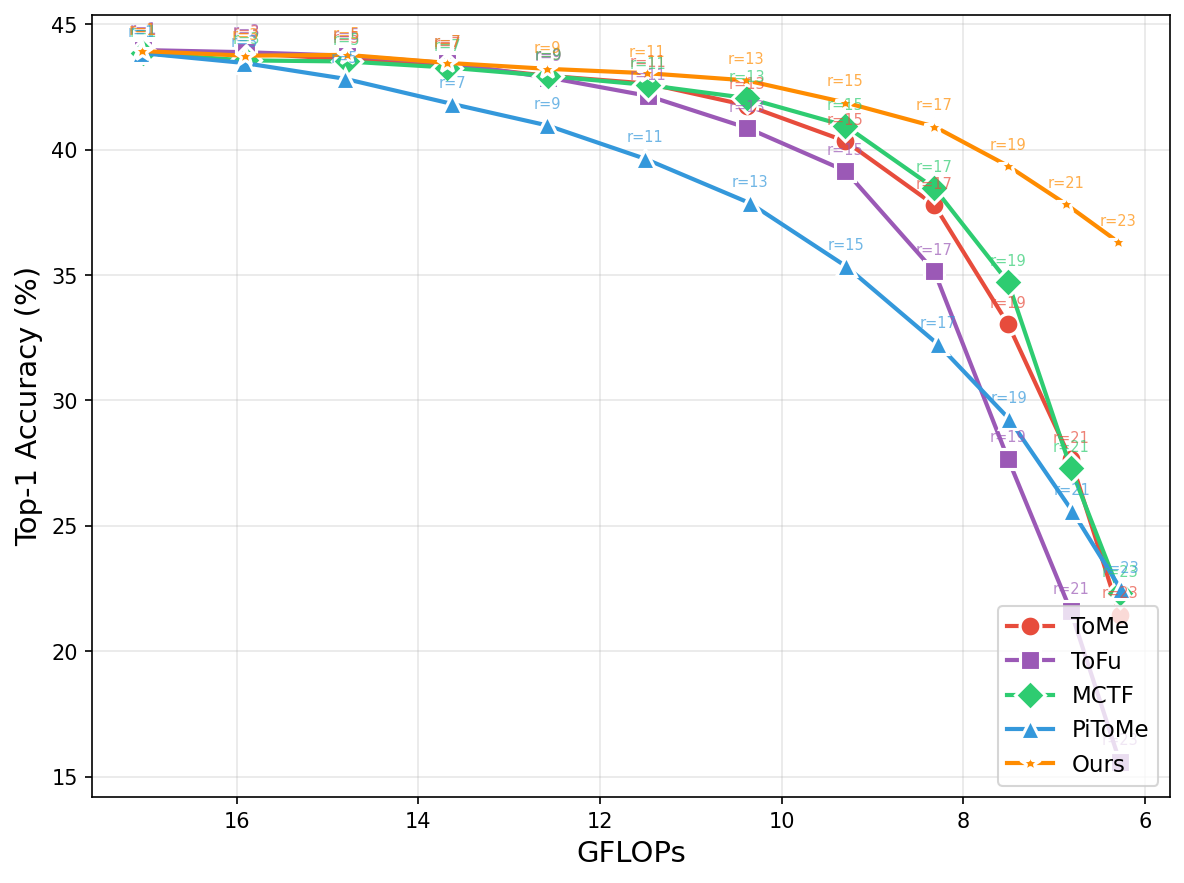}
    \caption{DeiT-3 Base --- shifted}
  \end{subfigure}
  \caption{Accuracy--GFLOPs Pareto curves (part 1): ViT-Base, DeiT-3 Large, and DeiT-3 Base.}
  \label{fig:pareto-appendix}
\end{figure*}

\begin{figure*}[t]
  \centering
  \begin{subfigure}[t]{0.48\textwidth}
    \centering
    \includegraphics[width=\linewidth]{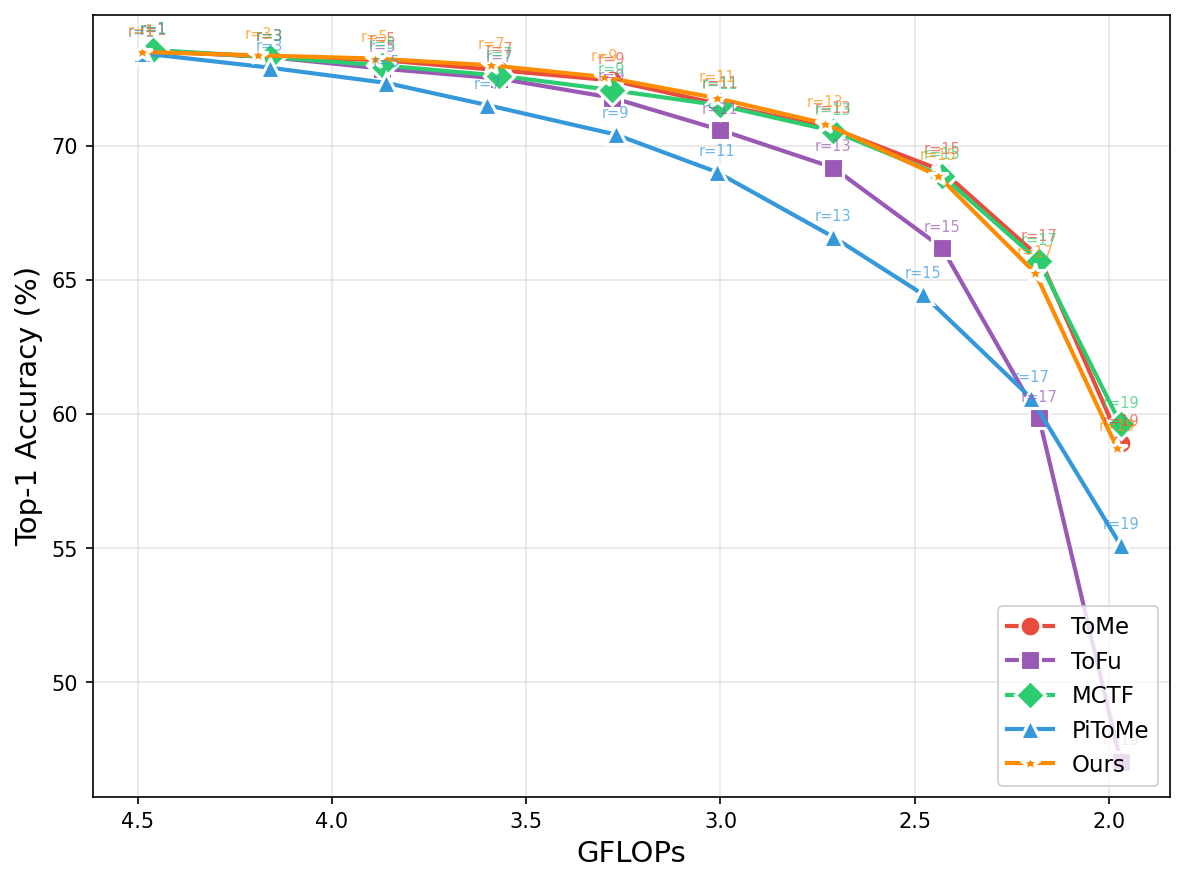}
    \caption{ViT-Small --- IN-1K}
  \end{subfigure}
  \hfill
  \begin{subfigure}[t]{0.48\textwidth}
    \centering
    \includegraphics[width=\linewidth]{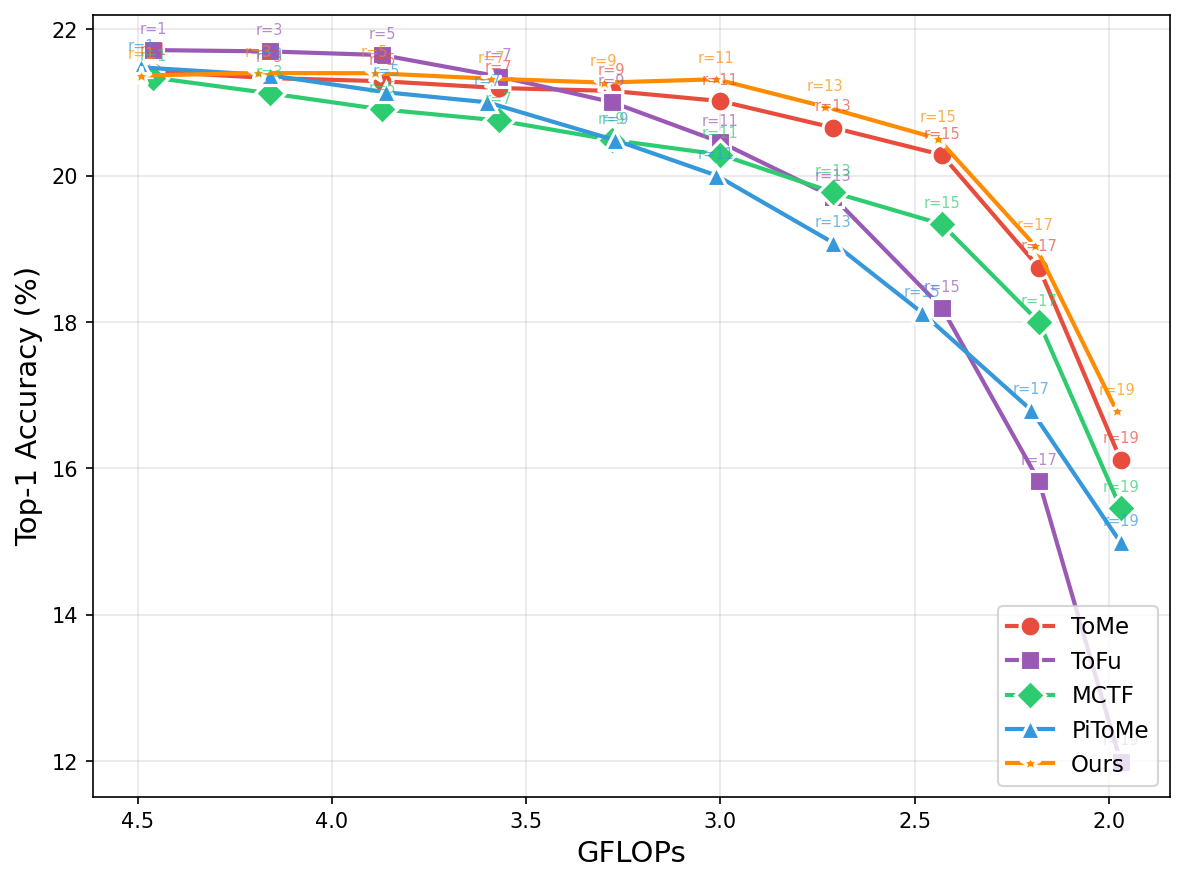}
    \caption{ViT-Small --- shifted}
  \end{subfigure}
  \\[4pt]
  \begin{subfigure}[t]{0.48\textwidth}
    \centering
    \includegraphics[width=\linewidth]{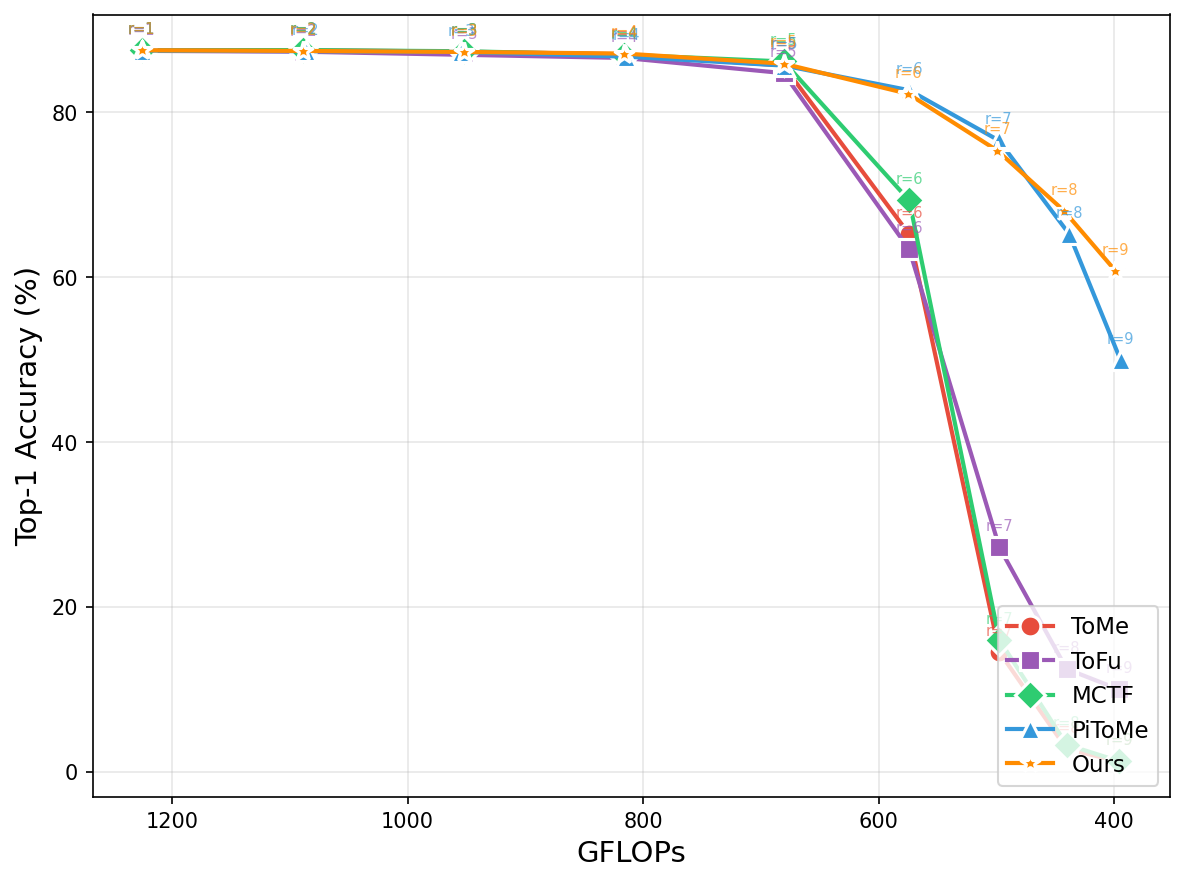}
    \caption{DINOv3-7B --- IN-1K}
  \end{subfigure}
  \hfill
  \begin{subfigure}[t]{0.48\textwidth}
    \centering
    \includegraphics[width=\linewidth]{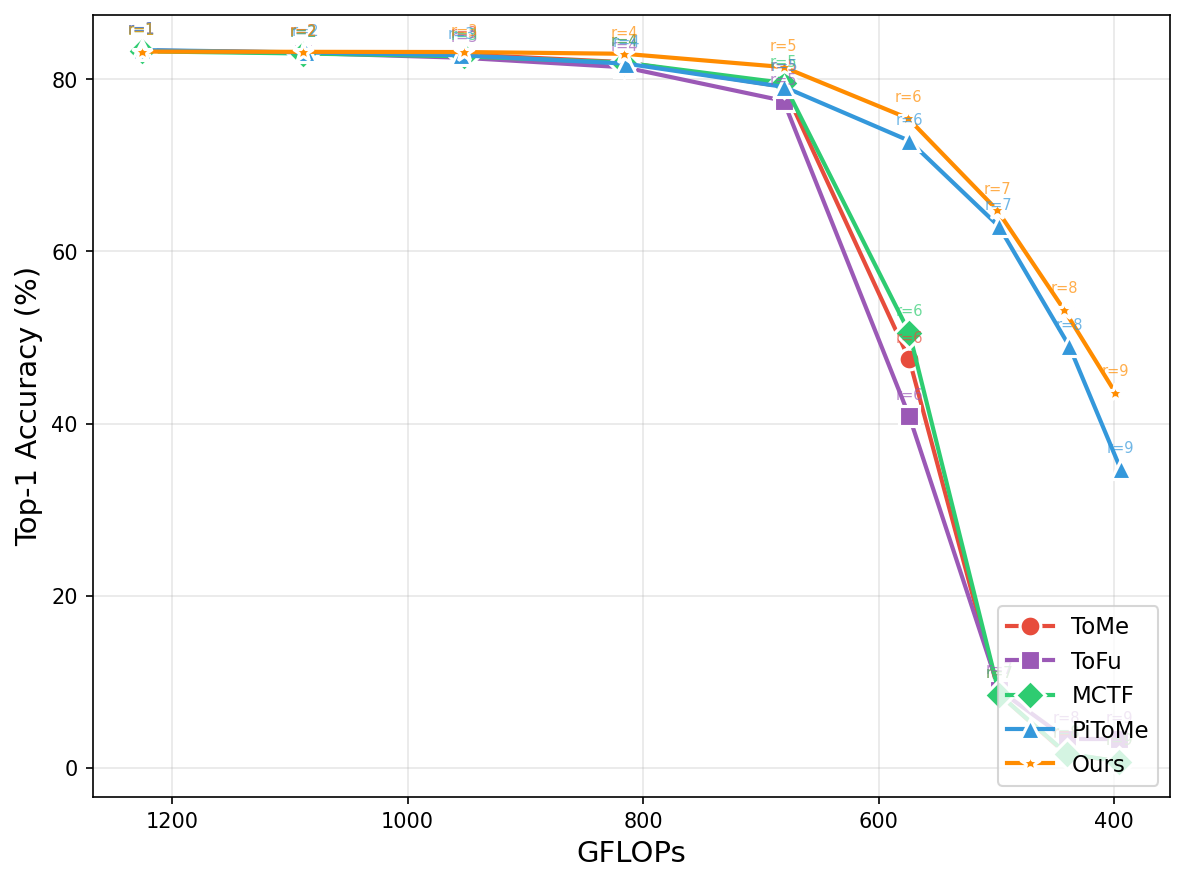}
    \caption{DINOv3-7B --- shifted}
  \end{subfigure}
  \caption{Accuracy--GFLOPs Pareto curves (part 2): ViT-Small and DINOv3~\citep{simeoni2025dinov3}. Together with Figure~\ref{fig:pareto-appendix}, these cover all models not shown in the main text. CATIS leads across all architectures and model scales, with the largest margins at high compression ratios.}
  \label{fig:pareto-appendix-2}
\end{figure*}

Complete Pareto curves for all models not shown in the main text (Figures~\ref{fig:pareto-appendix} and~\ref{fig:pareto-appendix-2}).

\section{Wall-Clock Throughput}
\label{app:throughput}

Table~\ref{tab:throughput} reports wall-clock throughput (images/second) and per-image latency measured on NVIDIA A100 GPUs with batch size 64. Each method is benchmarked at the same per-layer reduction parameter $r$ used in the main accuracy experiments (Table~\ref{tab:main}). Measurements use 50 warm-up iterations followed by 200 timed iterations with \texttt{torch.cuda.synchronize()} between batches; the reported values are the mean over all timed iterations.

\textbf{Key observations.} \textbf{(1) Negligible triage overhead.} At large model scales (ViT-Large, DeiT-3-Large), CATIS achieves throughput within 1--3\% of ToFu, the fastest baseline, despite performing additional argsort sorting, CLS attention extraction, and signal fusion at each reduced layer. The absolute difference is $<5$\,img/s for models running below 320\,img/s.
\textbf{(2) Consistent parity with ToMe/MCTF.} Across all six configurations, CATIS achieves throughput on par with ToMe and MCTF, confirming that its accuracy gains come at no wall-clock cost relative to these methods.
\textbf{(3) Small-model regime.} At DeiT-3-Small ($r{=}23$), CATIS is $\sim$10\% slower than ToFu (2616 vs.\ 2925\,img/s). This reflects the fact that the absolute cost of bookkeeping operations is non-trivial relative to the very short inference time of a small model at high compression. The speedup over vanilla (2.17$\times$) nonetheless demonstrates substantial wall-clock acceleration.
\textbf{(4) PiToMe overhead.} PiToMe is consistently 20--50\% slower than all other methods, consistent with its more complex per-layer scoring and matching procedure.

\begin{table*}[t]
  \caption{Wall-clock throughput (img/s) and latency (ms/img) measured on NVIDIA A100 GPUs, batch size~64, 200 timed iterations. Speedup is relative to the vanilla baseline on the same model. The GFLOPs column is matched across all methods for each configuration.}
  \label{tab:throughput}
  \centering
  \small
  \setlength{\tabcolsep}{6pt}
  \begin{tabular}{llcccc}
    \toprule
    Model ($r$) & Method & GFLOPs & Throughput (img/s) & Latency (ms/img) & Speedup \\
    \midrule
    \multirow{6}{*}{\shortstack[l]{ViT-L\\$r{=}11$}}
    & Vanilla  & 61.6 & 119.8 & 8.35 & 1.00$\times$ \\
    & ToMe     & 22.8 & 306.0 & 3.27 & 2.55$\times$ \\
    & ToFu     & 22.8 & 316.0 & 3.16 & 2.64$\times$ \\
    & PiToMe   & 22.8 & 228.7 & 4.37 & 1.91$\times$ \\
    & MCTF     & 22.8 & 298.2 & 3.35 & 2.49$\times$ \\
    & \textbf{CATIS} & 22.8 & \textbf{316.2} & \textbf{3.16} & \textbf{2.64}$\times$ \\
    \midrule
    \multirow{6}{*}{\shortstack[l]{ViT-B\\$r{=}13$}}
    & Vanilla  & 17.6 & 396.1 & 2.53 & 1.00$\times$ \\
    & ToMe     & 10.4 & 629.1 & 1.59 & 1.59$\times$ \\
    & ToFu     & 10.4 & 652.9 & 1.53 & 1.65$\times$ \\
    & PiToMe   & 10.3 & 457.3 & 2.19 & 1.15$\times$ \\
    & MCTF     & 10.4 & 605.5 & 1.65 & 1.53$\times$ \\
    & \textbf{CATIS} & 10.4 & \textbf{650.9} & \textbf{1.54} & \textbf{1.64}$\times$ \\
    \midrule
    \multirow{6}{*}{\shortstack[l]{ViT-S\\$r{=}13$}}
    & Vanilla  & 4.6  & 1180.0 & 0.85 & 1.00$\times$ \\
    & ToMe     & 2.7  & 1812.5 & 0.55 & 1.54$\times$ \\
    & ToFu     & 2.7  & 1885.7 & 0.53 & 1.60$\times$ \\
    & PiToMe   & 2.7  &  910.8 & 1.10 & 0.77$\times$ \\
    & MCTF     & 2.7  & 1694.1 & 0.59 & 1.44$\times$ \\
    & \textbf{CATIS} & 2.7  & \textbf{1787.9} & \textbf{0.56} & \textbf{1.52}$\times$ \\
    \midrule
    \multirow{6}{*}{\shortstack[l]{DeiT-3-L\\$r{=}11$}}
    & Vanilla  & 61.6 & 120.7 & 8.28 & 1.00$\times$ \\
    & ToMe     & 22.8 & 306.4 & 3.26 & 2.54$\times$ \\
    & ToFu     & 22.8 & 316.6 & 3.16 & 2.62$\times$ \\
    & PiToMe   & 22.8 & 227.8 & 4.39 & 1.89$\times$ \\
    & MCTF     & 22.8 & 298.3 & 3.35 & 2.47$\times$ \\
    & \textbf{CATIS} & 22.8 & \textbf{312.2} & \textbf{3.20} & \textbf{2.59}$\times$ \\
    \midrule
    \multirow{6}{*}{\shortstack[l]{DeiT-3-B\\$r{=}23$}}
    & Vanilla  & 17.6 &  402.8 & 2.48 & 1.00$\times$ \\
    & ToMe     &  6.3 & 1019.6 & 0.98 & 2.53$\times$ \\
    & ToFu     &  6.3 & 1060.8 & 0.94 & 2.63$\times$ \\
    & PiToMe   &  6.3 &  628.5 & 1.59 & 1.56$\times$ \\
    & MCTF     &  6.3 &  977.4 & 1.02 & 2.43$\times$ \\
    & \textbf{CATIS} &  6.3 & \textbf{1024.1} & \textbf{0.98} & \textbf{2.54}$\times$ \\
    \midrule
    \multirow{6}{*}{\shortstack[l]{DeiT-3-S\\$r{=}23$}}
    & Vanilla  & 4.6  & 1205.4 & 0.83 & 1.00$\times$ \\
    & ToMe     & 1.7  & 2784.4 & 0.36 & 2.31$\times$ \\
    & ToFu     & 1.7  & 2925.4 & 0.34 & 2.43$\times$ \\
    & PiToMe   & 1.6  &  991.2 & 1.01 & 0.82$\times$ \\
    & MCTF     & 1.7  & 2579.4 & 0.39 & 2.14$\times$ \\
    & \textbf{CATIS} & 1.7  & \textbf{2616.3} & \textbf{0.38} & \textbf{2.17}$\times$ \\
    \midrule
    \multirow{6}{*}{\shortstack[l]{DINOv3-7B\\$r{=}4$}}
    & Vanilla  & 1349.9 & 10.2 & 98.15 & 1.00$\times$ \\
    & ToMe     & 815.8 & 14.2 & 70.47 & 1.39$\times$ \\
    & ToFu     & 815.8 & 14.1 & 70.95 & 1.38$\times$ \\
    & PiToMe   & 815.9 & 11.7 & 85.47 & 1.15$\times$ \\
    & MCTF     & 815.8 & 13.8 & 72.39 & 1.36$\times$ \\
    & \textbf{CATIS} & 815.8 & \textbf{12.1} & \textbf{82.58} & \textbf{1.19}$\times$ \\
    \bottomrule
  \end{tabular}
\end{table*}


\section{Video Classification: Full Results and Throughput}
\label{app:video-throughput}

This appendix reports the full accuracy table and wall-clock throughput for the video classification experiment summarized in Section~\ref{sec:experiments} (VideoMAE ViT-B on UCF-101). Measurements use batch size~4, 10 warm-up and 50 timed iterations on NVIDIA A100 GPUs with \texttt{torch.cuda.synchronize()}.

\begin{table*}[t]
  \caption{Video classification accuracy on UCF-101 val (VideoMAE ViT-B, $r{=}160$, start layer~3). Top-1 accuracy (\%) and estimated GFLOPs. Best reduced result in \textbf{bold}.}
  \label{tab:video}
  \centering
  \small
  \setlength{\tabcolsep}{6pt}
  \begin{tabular}{@{}lcccc@{}}
    \toprule
    Method & GFLOPs & Tokens & Reduction & Accuracy (\%) \\
    \midrule
    \textcolor{gray}{Vanilla} & \textcolor{gray}{315.4} & \textcolor{gray}{1568} & \textcolor{gray}{---} & \textcolor{gray}{81.26} \\
    ToMe   & 214.1 & 128 & 91.8\% & 81.23 \\
    ToFu   & 214.1 & 128 & 91.8\% & 80.94 \\
    PiToMe & 214.1 & 128 & 91.8\% & 80.47 \\
    MCTF   & 214.1 & 128 & 91.8\% & 79.83 \\
    \textbf{CATIS} & 214.1 & 128 & 91.8\% & \textbf{81.73} \\
    \bottomrule
  \end{tabular}
\end{table*}

\begin{table*}[t]
  \caption{Wall-clock throughput (videos/s) and latency (ms/video) for VideoMAE ViT-B on UCF-101 ($r{=}160$, start layer~3). Speedup is relative to vanilla.}
  \label{tab:video-throughput}
  \centering
  \small
  \setlength{\tabcolsep}{6pt}
  \begin{tabular}{llcccc}
    \toprule
    Model ($r$) & Method & GFLOPs & Throughput (vid/s) & Latency (ms/vid) & Speedup \\
    \midrule
    \multirow{6}{*}{\shortstack[l]{VideoMAE ViT-B\\$r{=}160$}}
    & Vanilla  & 315.4 & 59.4  & 67.3 & 1.00$\times$ \\
    & ToMe     & 214.1 & 87.1  & 45.9 & 1.47$\times$ \\
    & ToFu     & 214.1 & 88.5  & 45.2 & 1.49$\times$ \\
    & PiToMe   & 214.1 & 83.5  & 47.9 & 1.40$\times$ \\
    & MCTF     & 214.1 & 76.9  & 52.0 & 1.29$\times$ \\
    & \textbf{CATIS} & 214.1 & \textbf{76.2} & \textbf{52.5} & \textbf{1.28}$\times$ \\
    \bottomrule
  \end{tabular}
\end{table*}

CATIS and MCTF have comparable throughput (1.28--1.29$\times$), while ToMe/ToFu are faster (1.47--1.49$\times$) due to simpler scoring. The throughput difference is modest ($<15\%$) relative to the accuracy advantage (+0.50--1.90pp over all baselines, Table~\ref{tab:video}).


\section{Spatial Visualization of CATIS Merge/Evict (DeiT-3-Small)}
\label{app:catis-spatial-viz}

\begin{figure*}[t]
  \centering
  \includegraphics[width=\textwidth]{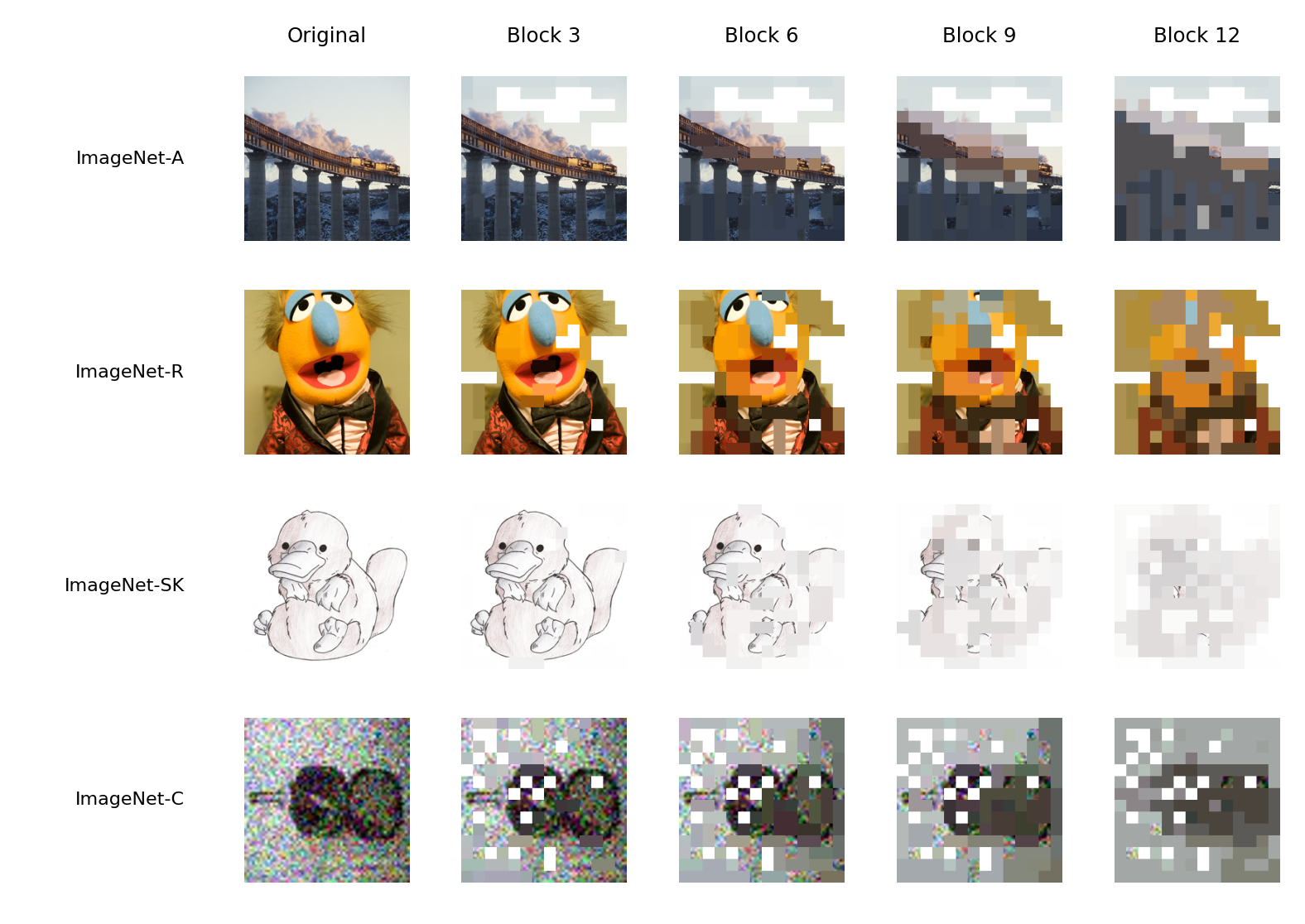}
  \caption{Spatial layout of CATIS merge/evict on DeiT-3-Small: one random image per row (ImageNet-A/R/Sketch/C); columns are \emph{Original} and after blocks 3, 6, 9, and 12. White denotes evicted patches; uniform colors denote merged tokens.}
  \label{fig:catis-spatial-viz}
\end{figure*}

Figure~\ref{fig:catis-spatial-viz} illustrates, on the original patch grid, how CATIS affects \emph{which} spatial regions remain after layer-wise \emph{merge/evict} in DeiT-3-Small.

\paragraph{Layout.}
Rows correspond to one random validation image each from ImageNet-A, ImageNet-R, ImageNet-Sketch, and ImageNet-C.
Columns show the input (\textbf{Original}) and the layout after completing transformer blocks $3,6,9,12$.
The per-layer reduction budget is $r$ tokens per layer.

\paragraph{Visual encoding.}
\textbf{White} cells are original $16{\times}16$ patches that are no longer covered by any token after eviction (\emph{evict} branch).
\textbf{Solid-colored tiles} indicate multiple original patches that have been merged into a single token; every patch in the same merge group is painted with the mean RGB of all pixels in that group, so merged regions read as flat color on the grid while preserving the original spatial tiling.

\section{Limitations and Future Work}
\label{app:limitations}

\paragraph{Diagonal approximation.}
The norm$_F$ signal approximates the full Mahalanobis distance with a diagonal covariance (Appendix~\ref{app:norm-f}).
This approximation is empirically validated across all 7 models, 3 pretraining paradigms, and 5 datasets:
$\rho_{\text{off}} < 0.25$ uniformly (Appendix~\ref{app:rho-off-vanilla}, Figures~\ref{fig:rho-off-vanilla-a} and~\ref{fig:rho-off-vanilla-b}), with cross-dataset variance below 0.03,
indicating that near-independence of feature dimensions is an architectural property of standard ViTs
rather than a data-dependent coincidence.
Architectures that explicitly encourage inter-dimension coupling
(e.g., capsule-style designs or group-equivariant networks) would require re-validating this property
before applying norm$_F$; the diagnostic tool $\rho_{\text{off}}$ itself provides the necessary check.

\paragraph{Task scope.}
Our evaluation focuses on image classification.
The diagnostic framework (the distortion recurrence, $\rho_s$, and $\rho_{\text{off}}$) is
task-agnostic by construction: it characterizes signal quality and feature-space integrity
independently of the downstream objective.
Whether the specific signals (norm$_F$, momentum CLS) transfer directly to
dense prediction tasks (detection, segmentation) or cross-modal retrieval is an empirical question;
the relative importance of individual tokens differs across tasks
(e.g., boundary tokens in segmentation), and task-specific signal design may yield further gains.

\paragraph{Orthogonal compression techniques.}
CATIS is orthogonal to quantization and knowledge distillation~\citep{lecun1989optimal, menghani2023efficient, hinton2015distilling},
which operate on weight precision and model capacity rather than token count.
Investigating potential synergies, or interference under joint application, is a natural extension.

\paragraph{Collapse boundary.}
The diagnostic framework correctly predicts that \emph{any} fixed-budget layer-wise method
has a finite collapse threshold, since the error amplifier is signal-agnostic (Section~\ref{sec:feedback-loop}).
CATIS delays this threshold through three mechanisms (lower $\epsilon_0$, lower effective $r$,
channel routing), achieving a proportionality constant of 186 vs.\ 161 for baselines on clean data
(Figure~\ref{fig:rcrit-vs-depth}), a 15\% improvement in the collapse boundary.
At extreme operating points (e.g., DeiT-3-Small at $r{=}23$, 63\% FLOPs reduction),
CATIS retains 58.20\% accuracy where all baselines fall to 28--45\%,
demonstrating continued advantage even beyond the regime where token reduction is low-cost.
The framework predicts that fully eliminating the amplifier requires moving beyond
fixed per-layer budgets, e.g., via adaptive layer-wise allocation
that concentrates reduction in shallow layers where $\rho_s$ remains high.

\end{document}